\documentclass[a4paper]{article}

\usepackage[pages=all, color=black, position={current page.south}, placement=bottom, scale=1, opacity=1, vshift=5mm]{background}
\SetBgContents{
	\textit{Preprint submitted to Computers \& Geosciences}
}      

\usepackage[margin=1in]{geometry} 

\usepackage{amsmath}
\usepackage{amsthm}
\usepackage{amssymb}

\usepackage{graphicx} 
\usepackage{float}
\usepackage{algorithm}  
\usepackage{algpseudocode}
\usepackage{color}
\usepackage{setspace}

\usepackage{hyperref}
\usepackage{amsmath}
\usepackage{algorithm}
\usepackage{algpseudocode}
\usepackage{emptypage} 
\usepackage{multicol} 
\usepackage{graphicx}
\usepackage{caption} 
\usepackage{float}
\usepackage{nomencl}
\usepackage{amsmath}
\usepackage{amsthm}
\usepackage{amssymb}
\usepackage{amsfonts}
\usepackage{bm}
\usepackage{tabularx}
\usepackage{longtable}
\usepackage{algorithm}
\usepackage{algpseudocode}
\usepackage{subcaption}
\usepackage{multirow} 

\usepackage[utf8]{inputenc}
\usepackage{hyperref}
\hypersetup{
	unicode,
	pdfauthor={Guido Di Federico, Louis J. Durlofsky},
	pdftitle={Latent diffusion models for parameterization and data assimilation of facies-based geomodels},
	pdfsubject={Latent diffusion models for parameterization and data assimilation of facies-based geomodels},
	pdfkeywords={Geological parameterization, diffusion models, latent diffusion, data assimilation, history matching, deep learning},
	pdfproducer={LaTeX},
	pdfcreator={pdflatex}
}

\usepackage[authoryear]{natbib}

\theoremstyle{plain}

\theoremstyle{definition}

\usepackage{graphicx, color}
\graphicspath{{fig/}}

\usepackage{algorithm, algpseudocode} 
\usepackage{mathrsfs} 

\usepackage{lipsum}

\title{Latent diffusion models for parameterization and data assimilation of facies-based geomodels}
\author{Guido Di Federico$^1$ \and Louis J. Durlofsky$^1$}

\date{
	\textit{\small$^1$Department of Energy Science \& Engineering, Stanford University, Stanford, CA, 94305, USA}\\ \small{\texttt{\{gdifede, lou\}@stanford.edu}}\\
}

\begin{document}
	\maketitle
	\hrule
	\begin{abstract}
        Geological parameterization entails the representation of a geomodel using a small set of latent variables and a mapping from these variables to grid-block properties such as porosity and permeability. Parameterization is useful for data assimilation (history matching), as it maintains geological realism while reducing the number of variables to be determined. Diffusion models are a new class of generative deep-learning procedures that have been shown to outperform previous methods, such as generative adversarial networks, for image generation tasks. Diffusion models are trained to ``denoise'', which enables them to generate new geological realizations from input fields characterized by random noise. Latent diffusion models, which are the specific variant considered in this study, provide dimension reduction through use of a low-dimensional latent variable. The model developed in this work includes a variational autoencoder for dimension reduction and a U-net for the denoising process. Our application involves conditional 2D three-facies (channel-levee-mud) systems. The latent diffusion model is shown to provide realizations that are visually consistent with samples from geomodeling software. Quantitative metrics involving spatial and flow-response statistics are evaluated, and general agreement between the diffusion-generated models and reference realizations is observed. Stability tests are performed to assess the smoothness of the parameterization method. The latent diffusion model is then used for ensemble-based data assimilation. Two synthetic ``true'' models are considered. Significant uncertainty reduction, posterior P$_{10}$--P$_{90}$ forecasts that generally bracket observed data, and consistent posterior geomodels, are achieved in both cases. \\

    \noindent\textbf{Keywords:} Geological parameterization, diffusion models, latent diffusion, data assimilation, history matching, deep learning
	\end{abstract}
        \hrule


\section{Introduction}
\label{intro}

\label{sec:intro}

The geological models used in subsurface flow simulation are typically represented on discrete grids, with each cell characterized by a set of properties such as permeability and porosity. Prior geomodels are constructed based on static well data, geological concepts, and geostatistical treatments. During the data assimilation process (referred to in this context as history matching), these models are updated by calibrating to dynamic production data. The large number of grid blocks in practical flow models, combined with the fact that many flow simulations must be performed during history matching, render this workflow computationally expensive. Geological parameterization procedures are useful for history matching as they allow the geomodel to be represented in terms of a relatively small set of independent latent variables. This is beneficial as it reduces the number of parameters that must be determined during data assimilation while acting to preserve geological realism.

A variety of low-dimensional parameterization methods have been proposed over the years. Previous treatments include, e.g., the use of discrete cosine transform \citep{Jafarpour} and principal component analysis (PCA) and its variants, such as K-PCA and O-PCA \citep{Sarma2008,Vo2015}. Although useful in many settings, these methods have limited accuracy in maintaining complex geological features in non-Gaussian systems \citep{Chan_2019}, particularly when there is a limited amount of hard data for conditioning. 

Recently, deep learning models have emerged as powerful alternatives for geological parameterization. Some of the methods relevant to this study include variatonal autoencoders, VAEs \citep{Canchumuni_2019, mo_caae, grana_co2}, in which a neural network learns the mapping from the geomodel data distribution to a latent space with a known distribution (and vice versa), CNN-PCA approaches \citep{LIU2021104676, Tang_2021}, which involve a convolutional neural network post-processing of a PCA representation, and generative adversarial networks, or GANs \citep{Laloy_2018, chan2019parametrization, zhang, Chan_2019, Song_Mukerji_Hou_2021, SONG2023129493, HU2023105290, Feng2024}. GANs have received substantial attention and they perform well in many cases. However, the adversarial training they require is typically time-consuming and may be unstable, which necessitates careful hyperparameter tuning. CNN-PCA methods currently treat only a single geological scenario at a time, and their performance depends on the amount of conditioning data (as is also the case for other PCA-based methods). Finally, while VAEs alone are generally superior to standard PCA for non-Gaussian models, they can produce blurry output of lower quality than GANs, or they may display unrealistic geometries \citep{Canchumuni_2019}.

In the last few years, generative diffusion models, introduced by \citet{ho2020denoising}, have become the state-of-the-art for image generation tasks, outperforming the established GANs in several benchmark studies \citep{dhariwal2021diffusion}. Diffusion methods, which do not require adversarial training, have found successful application in many fields of science and engineering that involve structured 2D or 3D data. This includes image and video generation, molecule and material design, super-resolution, and time-series forecasting \citep{review_dms}. These models are especially suitable for conditioned generative tasks, where conditioning can be in the form of scalars, text or images. These properties make diffusion models ideal candidates for geomodeling and parameterization. Diffusion models entail two discrete-step processes: noising (forward process) and denoising (reverse process). Noising progressively adds noise to the data, and the model is trained to learn the reverse process. In this way, diffusion models can generate new samples from a random noise input. There have been only a few published studies that have used these methods for subsurface flow applications. \citet{mosser}, for example, applied diffusion models for geostatistical simulation of 2D fluvial channel and river datasets. However, the latent space in these treatments had the same dimensionality as the geomodel space, limiting their application to small models. 

To address this issue, \citet{lee2023latent} proposed the use of latent diffusion models, or LDMs, developed by \citet{rombach2022highresolution}. LDMs operate in the same way as standard diffusion models, but they use a low-dimensional latent variable in conjunction with an encoder-decoder network architecture. \citet{lee2023latent} applied LDMs for the conditional generation of multifacies 2D geomodels. This method, however, relies on a random sequence of sampling steps (known as denoising diffusion probabilistic model, DDPM). In addition, a large number of steps is required to achieve high geomodel quality, resulting in long generation times. These features may limit the applicability of this approach for history matching problems. \citet{song2021denoising} introduced denoising diffusion implicit models (DDIMs), which represent a compromise between computational time and accuracy. DDIMs enable faster, deterministic sampling, with added latent space interpolation capabilities and image quality that is comparable to DDPMs \citep{song2021denoising}. To our knowledge, DDIM sampling in combination with LDMs has not been considered for geomodel parameterization or data assimilation.

In this work, we extend and apply LDMs to parameterize 2D multifacies (channel-levee-mud) geomodels. In contrast to previous implementations, our workflow entails fast deterministic sampling and smooth/stable parameterization, both of which are important for data assimilation. Training samples are obtained using object-based modeling in Petrel. Metrics involving spatial statistics and flow quantities are used to assess the quality of the LDM geomodels relative to (reference) object-based geomodels. History matching, specifically ESMDA \citep{EMERICK20133} applied to the latent variables, is performed with the LDM geomodels for two-phase (oil-water) problems. 

This paper proceeds as follows. In Section~\ref{sec:diff_param}, we first review diffusion models as a generative deep learning method, and then describe our LDM procedure for geomodel parameterization. Next, in Section~\ref{sec:results_models}, visual, statistical and flow-based comparisons between diffusion-model realizations and Petrel realizations are presented. An assessment demonstrating the stability of the parameterization is also provided. In Section~\ref{sec:results_hm}, we present results for ESMDA-based history matching using the diffusion model parameterization. Two different synthetic ``true'' models are considered. Finally, in Section~\ref{sec:conclusion} we provide a summary of this work and suggestions for future research. 

An earlier version of this paper will appear as an uncopyrighted contribution to the European Conference on the Mathematics of Geological Reservoirs (ECMOR 2024), to be held in September 2024. The detailed history matching results presented in this paper differ from those in the conference paper.

\section{Diffusion-model-based parameterization}
\label{sec:diff_param}

In this section, we first review the DDPM \citep{ho2020denoising} diffusion generative process. Next, we describe the DDIM \citep{song2021denoising} sampling procedure and LDM parameterization \citep{rombach2022highresolution}. Finally, we present the specific approach and model architecture adopted in this work.

\subsection{Denoising diffusion probabilistic models (DDPMs)}
\label{sec:ddpm}

\begin{figure}
  \centering
  \includegraphics[width=0.7\textwidth]{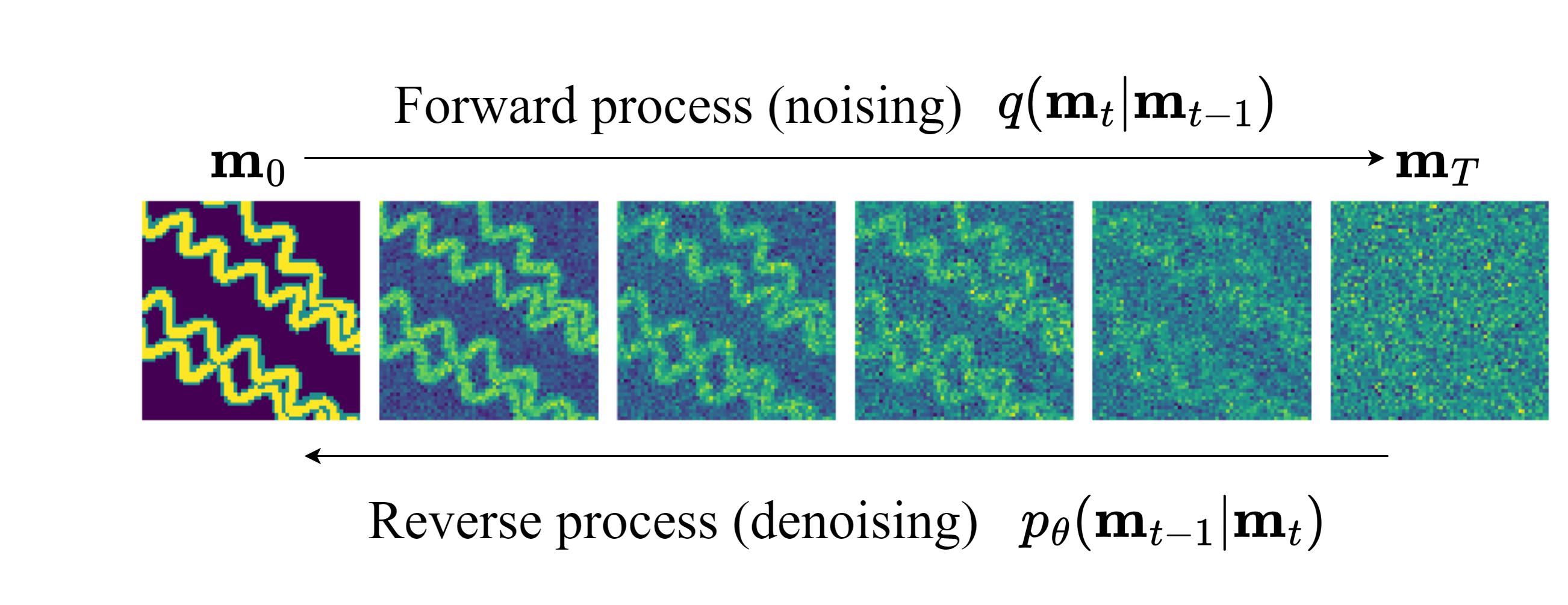}
  \caption{Illustration of the diffusion process for a 2D channelized geomodel.}
  \label{fig:noising_denoising}
\end{figure}

Diffusion models entail two discrete-step processes -- noising and denoising. During noising, random perturbations are gradually added to an image (geomodel) through a specified scheduler function. During denoising, noise is gradually removed. In the noising process, a noisy image is obtained from a clean image, while in the latter process the opposite is achieved. A neural network is trained to predict the noise contained in an image at a given step. Once the model is trained, noise can be successively subtracted such that the noisy image is gradually ``cleaned.'' In this way, starting from Gaussian noise, a new realization can be generated. An illustration of the two processes is shown in Figure~\ref{fig:noising_denoising}.

Consider a set of conditioned geomodels, represented on a discrete grid comprised of $N_x \times N_y = N_c$ cells. We denote a sample (realization) as $\mathbf{m}_0 \in \mathbb{R}^{N_x \times N_y}$, where the $0$ subscript refers to the clean (without noise) geomodel. The distribution characterizing the geomodel, $q(\mathbf{m}_0)$, is unknown. Starting from $\mathbf{m}_0$, the noising (forward) process $q$ over $T$ discrete steps is modeled as a Markov chain of discrete Gaussian transition steps \citep{ho2020denoising}: 
\begin{equation}
    q(\mathbf{m}_{1:T} | \mathbf{m}_0) = \prod_{t=1}^{T} q(\mathbf{m}_t | \mathbf{m}_{t-1}), \ \   \text{where} \ \    q(\mathbf{m}_{t} | \mathbf{m}_{t-1}) = \mathcal{N}(\mathbf{m}_{t}; \sqrt{\alpha_t} \mathbf{m}_{t-1}, (1-\alpha_t)\mathbf{I}).
\end{equation}
Here $\alpha_t$ specifies the scheduler parameter that gradually adds Gaussian noise, with mean $\sqrt{ \alpha_t} \mathbf{m}_{t-1}$ and variance $(1- \alpha_t)\mathbf{I}$, to $\mathbf{m}_t$, through the $\mathcal{N}$ operator, where $\mathbf{I}$ is the identity matrix. It is common to use the unit complement of $\alpha_t$, $\beta_t = 1-\alpha_t$. For a linear scheduler, the value of $\beta_t$ increases linearly from $\beta_1$ to $\beta_T$. 
Overall, noising transforms $\mathbf{m}_0$ into the noisy image $\mathbf{m}_T$. It can be shown that the forward process admits sampling at an arbitrary step $t$ as:
\begin{equation}
    q(\mathbf{m}_t|\mathbf{m}_{0}) =
    \int q(\mathbf{m}_{1:t} | \mathbf{m}_0) d \mathbf{m}_{1:t-1} = \mathcal{N}(\mathbf{m}_{t}; \sqrt{\bar \alpha_t} \mathbf{m}_{0}, (1-\bar \alpha_t)\mathbf{I}),
    \label{eqn:forward}
\end{equation}
where $\bar \alpha_t := \prod_{s=1}^{t} \alpha_s$. As such, the geomodel at noising step $t$ is $\mathbf{m}_t = \sqrt{\bar \alpha_t} \mathbf{m}_0 + \sqrt{1-\bar \alpha_t} \boldsymbol{\epsilon}$,  where $\boldsymbol{\epsilon} = \mathcal{N} (\mathbf{0}, \mathbf{I})$. 

The denoising (inverse) process $p_{\theta}$ is learned by a neural network, where subscript $\theta$ denotes the network parameters. Starting from noise $p_{\theta}(\mathbf{m}_{T})$, this can be expressed as \citep{ho2020denoising}:
\begin{equation}
    p_{\theta}(\mathbf{m}_{0:T}) = p_{\theta}(\mathbf{m}_T) \prod_{t=1}^{T} p_{\theta}(\mathbf{m}_{t-1} | \mathbf{m}_t), \ \  \text{where} \ \      p_{\theta}(\mathbf{m}_{t-1} | \mathbf{m}_t) = \mathcal{N}(\mathbf{m}_{t-1}; \bm{\mu}_{\theta} (\mathbf{m}_t,t), \bm{\Sigma}_{\theta}(\mathbf{m}_t,t)).
    \label{eqn:reverse}
\end{equation}
Here $\bm{\mu}_{\theta}$ and $\bm{\Sigma}_{\theta}$ are the learnable parameters of the inverse Gaussian steps. In \citet{ho2020denoising}, the $\bm{\Sigma}_{\theta}$ parameters are taken as $\sigma_t^2 \bf{I}$, with $\sigma_t^2 = \beta_t$, such that they are constant (not learned) and defined by the scheduler. The means ($\boldsymbol{\mu}_{\theta}$), by contrast, are learned and parameterized as:
\begin{equation}
    \boldsymbol{\mu}_\theta\left(\mathbf{m}_t, t\right)
    =\frac{1}{\sqrt{ \alpha_t}}\left(\mathbf{m}_t-\frac{1 - \alpha_t}{\sqrt{1-\bar{\alpha}_t}} \boldsymbol{\epsilon}_\theta\left(\mathbf{m}_t, t\right)\right),
    \label{eqn:param_means}
\end{equation}
where $\boldsymbol{\epsilon}_{\theta}$ is a function approximator that predicts the noise contained in geomodel $\mathbf{m}_t$ at step $t$. In other words, the network is trained to predict the noise instead of the mean values.

In this way, when the network is provided with a clean model $\mathbf{m}_0$ that has been noised until step $t$ (which gives $\mathbf{m}_t$), the diffusion model is trained to predict the noise $\bm{\epsilon}_{\theta}$. This noise will then be subtracted during denoising. Training is performed by minimizing the loss function between true noise (added by the scheduler and therefore known) and predicted noise \citep{ho2020denoising}:
\begin{equation}
        L(\theta) =   \mathbb{E}_{\mathbf{m}_t,\bm{\epsilon}} \left [ \Big|\Big| \bm{\epsilon} - \bm{\epsilon}_{\theta}(\mathbf{m}_t,t) \Big|\Big|_2 ^2\right ]= \mathbb{E}_{t, \mathbf{m}_0,\bm{\epsilon}} \left [ \Big|\Big| \bm{\epsilon} - \bm{\epsilon}_{\theta}\left( \sqrt{\bar{\alpha}_t} \mathbf{m}_0 + \sqrt{1 - \bar \alpha_t}\bm{\epsilon},t \right) \Big|\Big|_2 ^2\right ].
        \label{eqn:loss_ddpm}
\end{equation}

In practice, the loss is computed over mini-batches and averaged. After training, new models can be generated starting from $\mathbf{m}_T \sim \mathcal{N}(\mathbf{0},\mathbf{I})$ by applying the following expression $T$ times:
\begin{equation}
    \mathbf{m}_{t-1}=\frac{1}{\sqrt{ \alpha_t}}\left(\mathbf{m}_t-\frac{1 - \alpha_t}{\sqrt{1-\bar{\alpha}_t}} \boldsymbol{\epsilon}_\theta\left(\mathbf{m}_t, t\right)\right)+\sigma_t \mathbf{z}, \ \  \text{ with } \ \  \mathbf{z} \sim \mathcal{N} (\bm{0}, \bf{I}).
    \label{eqn:sampling_ddpm}
\end{equation}

\subsection{Denoising diffusion implicit models (DDIMs)}
\label{sec:ddim}

In the DDPM procedure described above, the number of steps $T$ required to achieve high-quality geomodels can be $O(10^3)$. In addition, Eq.~\ref{eqn:sampling_ddpm} contains the random component $\mathbf{z}$, which renders the generation process nondeterministic, even when starting with the same initial noise $\mathbf{m}_T$. This may be a limitation for data assimilation problems, where updates to the input variables ($\mathbf{m}_T$ in this case) should affect the generated output ($\mathbf{m}_0$) in a predictable manner.  To overcome this limitation, \citet{song2021denoising} introduced denoising diffusion implicit models (DDIMs). They suggested reparameterizing the reverse diffusion in Eq.~\ref{eqn:sampling_ddpm} from a Markov chain to a non-Markovian expression for a single step $\mathbf{m}_{t-1}$:
\begin{equation}
    \mathbf{m}_{t-1}=\sqrt{\bar{\alpha}_{t-1}}\left(\frac{\mathbf{m}_t-\sqrt{1-\bar{\alpha}_t} \boldsymbol{\epsilon}_\theta\left(\mathbf{m}_t,t\right)}{\sqrt{\bar{\alpha}_t}}\right)+\sqrt{1-\bar{\alpha}_{t-1}} \boldsymbol{\epsilon}_\theta (\mathbf{m}_t,t),
    \label{eqn:sampling_ddim}
\end{equation}
where $\bar{\alpha}_t = \prod_{s=1}^{t} \alpha_s$, and similarly for $\bar{\alpha}_{t-1}$. DDIMs maintain image quality that is comparable to that of DDPMs \citep{song2021denoising}, with the advantages of faster sampling (with as few as 20-50 steps), higher consistency in the generative process ($\mathbf{m}_0$ depends only on the initial state $\mathbf{m}_T$), and interpolation capabilities. In addition, because the training objective with the DDIM is the same as that for the DDPM, a DDIM can be used for inference on a DDPM-trained diffusion model. More details can be found in \citet{song2021denoising}.

\subsection{Latent diffusion models (LDMs)}
\label{sec:ldm}
\begin{figure}
    \centering
    \begin{subfigure}[b]{0.45\textwidth}
        \centering
        \includegraphics[width=\textwidth, trim=0 0 0cm 0, clip]{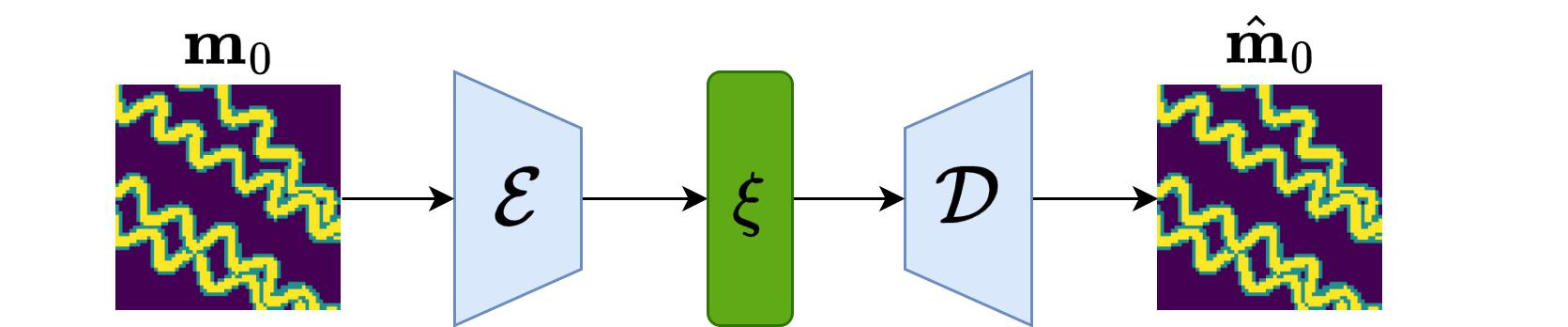}
        \vspace{0.8cm}
        \caption{Encoder-decoder component}
        \label{fig:vae}
    \end{subfigure}
    \begin{subfigure}[b]{0.5\textwidth}
        \centering
        \vspace{0.8cm}
        \includegraphics[width=\textwidth, trim=0.5cm 0 0cm 0, clip]{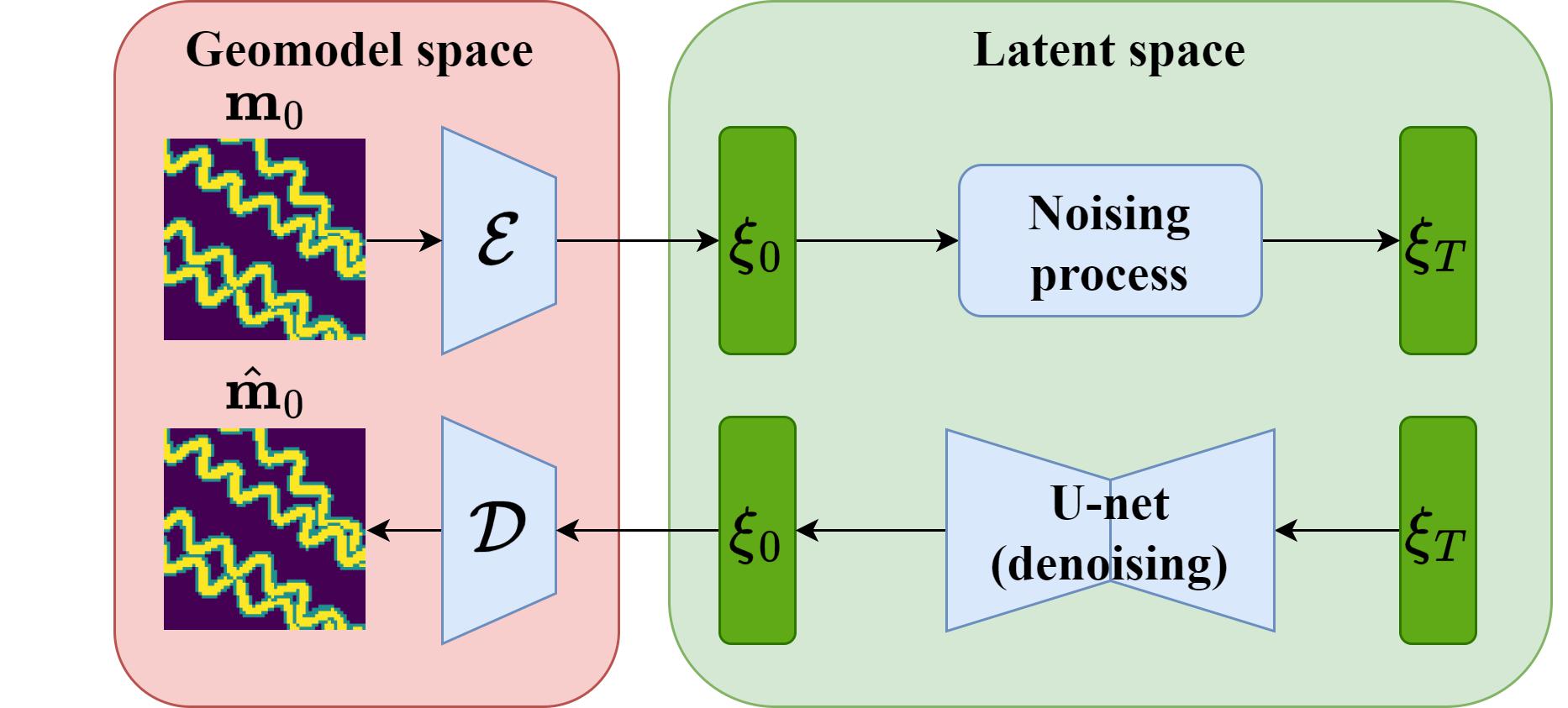}
        \caption{Full LDM with denoising network}
        \label{fig:ldm}
    \end{subfigure}
    \caption{Schematic representation of encoder-decoder and full LDM architecture (inspired by \citet{rombach2022highresolution}).}
    \label{fig:arch_subfigures}
\end{figure}

In the diffusion models described thus far, the latent space has the same dimensionality as the geomodel space, i.e., the parameterization involves a mapping, but no dimension reduction. For data assimilation applications, it is beneficial to reduce the number of geomodel parameters that must be determined, and many previous parameterization methods provide this capability. The LDMs introduced by \citet{rombach2022highresolution} are very useful in this regard. They have the same structure as standard diffusion models, but the noising and denoising operations are performed on a lower-dimension latent variable, $\boldsymbol{\xi} \in \mathbb{R}^{n_x \times n_y}$, where $n_x = N_x / f$ and $n_y = N_y / f$. Here $f$ is referred to as the downsampling ratio. An encoder $\mathcal{E}(\mathbf{m}): \mathbb{R}^{N_x \times N_y} \rightarrow \mathbb{R}^{n_x \times n_y}$ and a decoder $\mathcal{D}(\boldsymbol{\xi}): \mathbb{R}^{n_x \times n_y} \rightarrow \mathbb{R}^{N_x \times N_y}$, as shown in Figure~\ref{fig:arch_subfigures}a, perform these mappings. A schematic of the full LDM is shown in Figure~\ref{fig:arch_subfigures}b.

\begin{figure}
    \centering
    \begin{subfigure}[b]{0.95\textwidth}
        \centering
        \includegraphics[width=\textwidth, trim=0cm 0 0cm 0, clip]{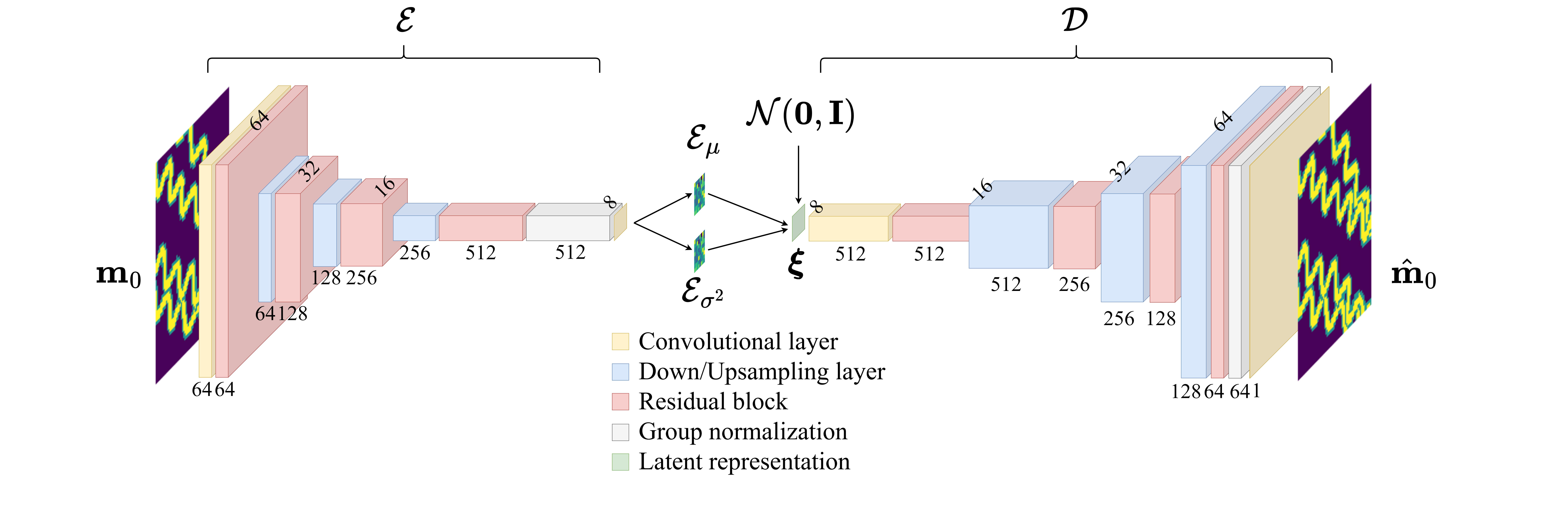}
        \caption{VAE architecture}
        \label{fig:arch_detailed_vae}
    \end{subfigure}
    \vspace{1cm}
        \begin{subfigure}[b]{0.99\textwidth}
        \centering
        \includegraphics[width=\textwidth, trim=0cm 0 0cm 0, clip]{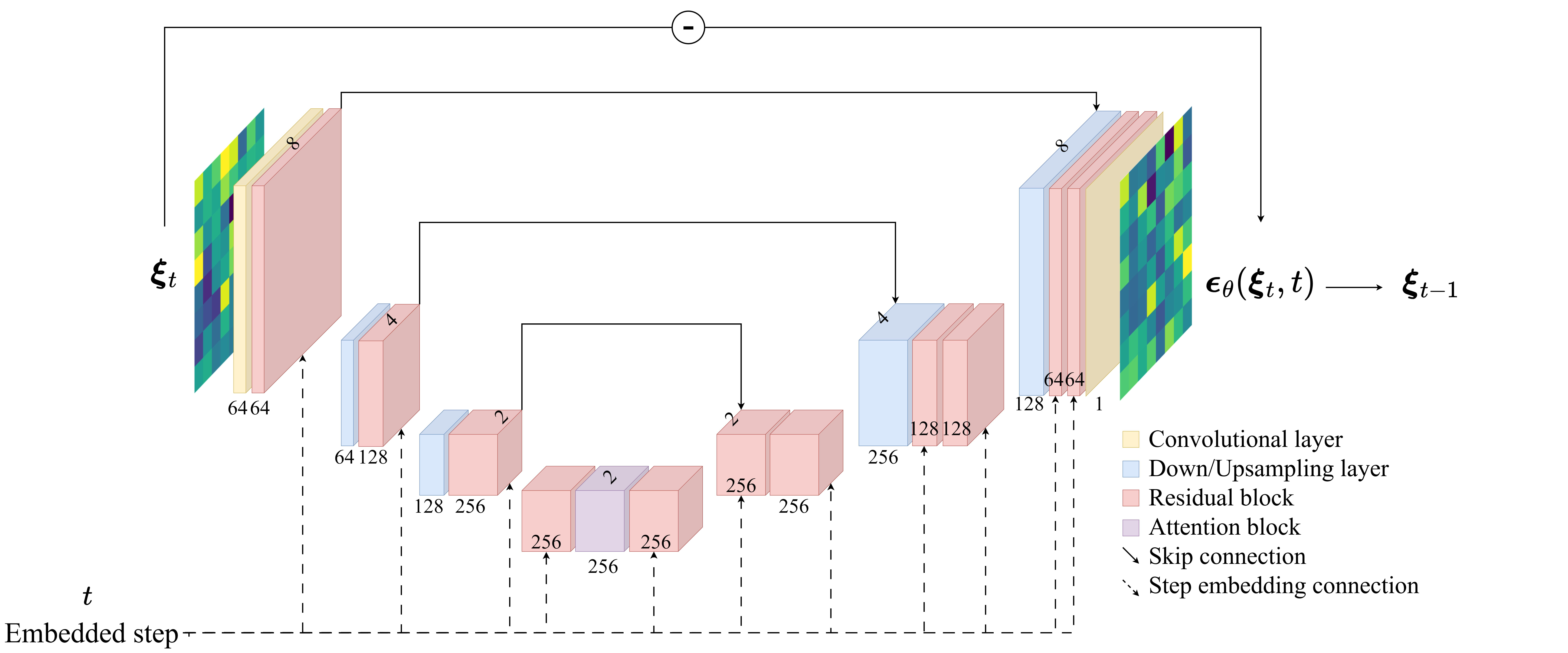}
        \caption{Denoising U-net architecture}
        \label{fig:arch_detailed_unet}
    \end{subfigure}
    \caption{Architecture of the LDM used in this work. The values on the blocks/layers indicate the number of channels and dimensions.}
    \label{fig:arch_detailed}
\end{figure}

The LDM used in this study consists of a variational autoencoder (VAE) and a U-net \citep{unet}. The VAE performs the dimension reduction from the geomodel space to the latent space and vice-versa (Figure~\ref{fig:vae}). The U-net is used for the reverse diffusion process (Figure~\ref{fig:ldm}). The VAE consists of an encoder and a decoder. The encoder is comprised of three downsampling convolutional layers, alternated with four residual blocks (group normalization -- convolution -- group normalization -- convolution), such that the total downsampling ratio $f = 8$ and the latent space $\boldsymbol{\xi}$ is of dimensions $N_x / 8 \times N_y /8 $. Note that, in contrast to most VAE implementations, which rely on a 1D latent space, LDMs use a 2D latent space. The decoder mirrors the encoder, and thus has four residual blocks and three upsampling convolutional layers.
The U-net consists of two downsampling residual blocks, one middle block with an attention mechanism \citep{attention}, and two upsampling residual blocks. The overall architecture follows the implementation of the original LDM in \citet{rombach2022highresolution}. Our implementation is developed using the Python libraries \texttt{diffusers} and \texttt{monai} (see the "Code availability" section for additional details). The detailed VAE and U-net architectures are shown in Figure~\ref{fig:arch_detailed}.

\subsection{LDM loss function and training}
\label{sec:training}
The encoder-decoder and the full diffusion network are trained in sequence. Three loss terms -- reconstruction loss, Kullback-Leibler (K-L) divergence loss, and hard data loss -- are considered in the encoder-decoder training.
Reconstruction loss is the error between the original model $\mathbf{m}_0$ and its reconstructed counterpart, $\hat{\mathbf{m}}_0 =\mathcal{D}(\mathcal{E}(\mathbf{m}_0))$:
\begin{equation}
    L_{recon}  = || \mathbf{m}_0 -  \hat{\mathbf{m}}_0 ||_2^2.
    \label{eqn:loss_recon}
\end{equation}
The latent variable $\boldsymbol{\xi}$ is intended to be standard normal \citep{rombach2022highresolution}, and the K-L divergence quantifies the difference between the distribution provided by the encoder and $\mathcal{N}(\mathbf{0}, \mathbf{I})$, i.e.,
\begin{equation}
    L_{KL} = D_{KL}\left[\mathcal{N}(\mathcal{E}_{\mu},\mathcal{E}_{\sigma^2}) || \mathcal{N}(\mathbf{0}, \mathbf{I}) \right].
    \label{eqn:loss_kl}
\end{equation}
The hard-data loss term acts to ensure that the decoded models honor conditioning at well locations. This loss is given by:
\begin{equation}
    L_h = \frac{1}{N_h} ||\mathbf{H}( \mathbf{m}_0 - \hat{\mathbf{m}}_0) ||_2^2,
    \label{eqn:loss_hd}
\end{equation}
where $\mathbf{H}$ is a matrix of zeros and ones that extracts the $N_h$ hard data locations. The overall loss is thus given by:
\begin{equation}
       L_{VAE}(\theta) = L_{recon} + \lambda_{KL} L_{KL} + \lambda_h L_h.
       \label{eqn:loss_vae}
\end{equation}
Here $\lambda_{KL}$ and $\lambda_{h}$ are weights that are found through numerical experimentation. 

After the encoded-decoder is trained, training is performed for the diffusion network. The loss function is of the same form as Eq.~\ref{eqn:loss_ddpm}, but it now involves the latent variable $\boldsymbol{\xi}$ instead of $\mathbf{m}$, i.e.,
\begin{equation}
        L_{LDM}(\theta) = \mathbb{E}_{t,\mathcal{E}(\mathbf{m}_0),\bm{\epsilon}} \left [ \Big|\Big| \bm{\epsilon} - \bm{\epsilon}_{\theta}\left( \sqrt{\bar\alpha_t} \boldsymbol{\xi}_0 + \sqrt{1 - \bar \alpha_t}\bm{\epsilon},t \right) \Big|\Big| ^2 _2\right ].
        \label{eqn:loss_ldm}
\end{equation}
In practice, during training, all losses are computed in batches of $N_{b}$ models. Pseudo-code for the U-net training is provided in Algorithm~\ref{alg:ldm_train} (modified from \citet{ho2020denoising}). For inference (generation), denoising starts from a lower dimensionality latent vector $\boldsymbol{\xi}_T$. The denoising in Eq.~\ref{eqn:sampling_ddim} is applied on $\boldsymbol{\xi}$. At the final step it is transformed by the decoder into a full-size geomodel $\mathbf{m}_{DM}$. This procedure is summarized in Algorithm~\ref{alg:ldm_generate} (modified from \citet{ho2020denoising}).

\begin{minipage}[t]{0.45\textwidth}
\begin{algorithm}[H]
\caption{Training}\label{alg:ldm_train}
\begin{algorithmic}[1]
\While {\text{not converged}}
\State $\mathbf{x}_0 \sim q(\mathbf{x}_0)$
\State $\boldsymbol{\xi}_0 = \mathcal{E}(\mathbf{x}_0)$
\State $t \sim U({1, \ldots, T})$
\State $\boldsymbol{\epsilon} \sim \mathcal{N}(\mathbf{0}, \mathbf{I})$
\State Gradient descent step on $\nabla_{\theta} L_{LDM}(\theta)$
\EndWhile
\end{algorithmic}
\end{algorithm}
\end{minipage}
\hfill
\begin{minipage}[t]{0.5\textwidth}
\begin{algorithm}[H]
\caption{Inference (generation)}\label{alg:ldm_generate}
\begin{algorithmic}[1]
\State $\boldsymbol{\xi}_T \sim \mathcal{N}(\mathcal{E}_{\mu}, \mathcal{E}_{\sigma^2})$
\For{${t = T, \ldots, 1}$}
        \State Apply Eq.~\ref{eqn:sampling_ddim} on $\boldsymbol{\xi}_{t-1}$ (instead of $\mathbf{m}_{t-1}$).
\EndFor
\State ${\mathbf{m}}_{DM} = \mathcal{D}(\boldsymbol{\xi}_0)$
\State \textbf{return} ${\mathbf{m}}_{DM}$
\end{algorithmic}
\end{algorithm}
\end{minipage}

\vspace{1cm}

\begin{figure}
  \centering
  \includegraphics[width=0.8\textwidth, trim = 0 0 5cm 0, clip]{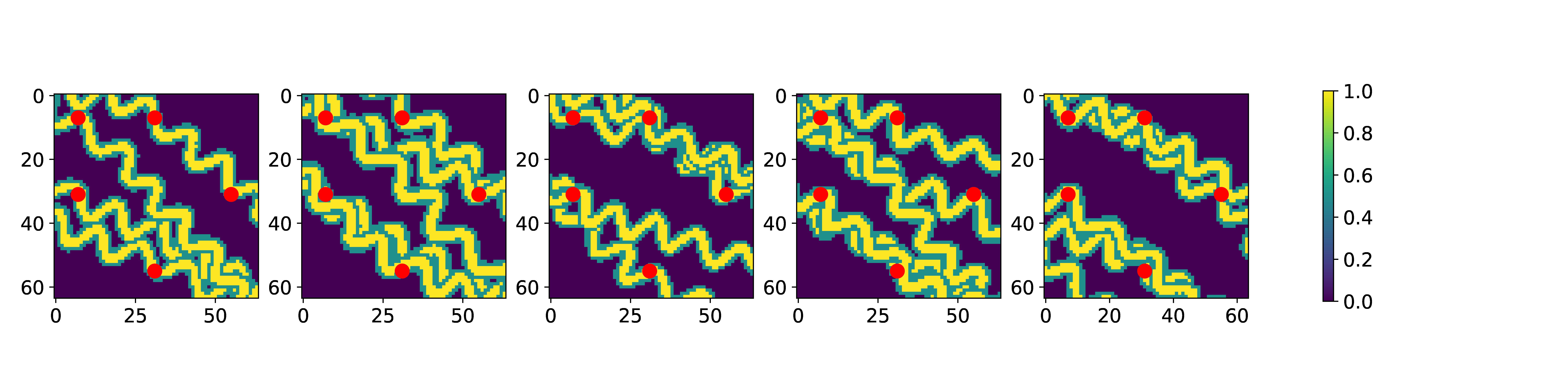}
  \caption{Training samples for three-facies models. The five conditioning points are shown in red. Colorbar in this figure applies to all subsequent figures of this type.}
  \label{fig:petrel_samples}
\end{figure}

The dataset used in this study consists of 4000 conditional realizations, of dimensions $64 \times 64$ ($N_x \times N_y = N_c = 4096$), of three-facies geomodels. The facies correspond to high-permeability channel sand, low-permeability background mud, and intermediate-permeability levee features. The training realizations are generated with Petrel software \citep{Schlumberger} using object-based modeling. These realizations are of similar style to those used by \citet{Song_Mukerji_Hou_2021}. A 70-20-10\% split for training, validation, and testing is applied. There are five hard-data locations, where the models are conditioned to channel facies. Five training realizations are shown in Figure~\ref{fig:petrel_samples}.

The total training time is about 2~h (of which 0.5~h is for VAE training and 1.5~h is for U-net training) on one Nvidia A100 GPU. We use the Adam optimizer \citep{adam}, with a learning rate of 1e-4 and a batch size of 16. The scheduler for the noising process is linear, with $\beta_1= 0.0001$ and $\beta_T = 0.02$, as in \citet{ho2020denoising}. A total of 100 DDIM steps are applied. The weights $\lambda_{KL}$ and $\lambda_h$ are set to $10^{-6}$ (as in \citet{rombach2022highresolution}) and $10$, respectively.

\section{Geomodel generation and flow simulations using diffusion models}
\label{sec:results_models}
We now utilize the diffusion model parameterization to generate realizations for the three-facies channelized system. Both visualizations and quantitative metrics, including spatial statistics and flow simulation results, are used to assess the similarity between a set of LDM-generated models and a set of reference models generated by geostatistical software. All comparisons involve 200 new LDM geomodels and 200 new Petrel geomodels.

\begin{figure}
    \centering
    \begin{subfigure}[b]{0.3\textwidth}
        \centering
        \includegraphics[width=\textwidth, trim=0 0 0 0, clip]{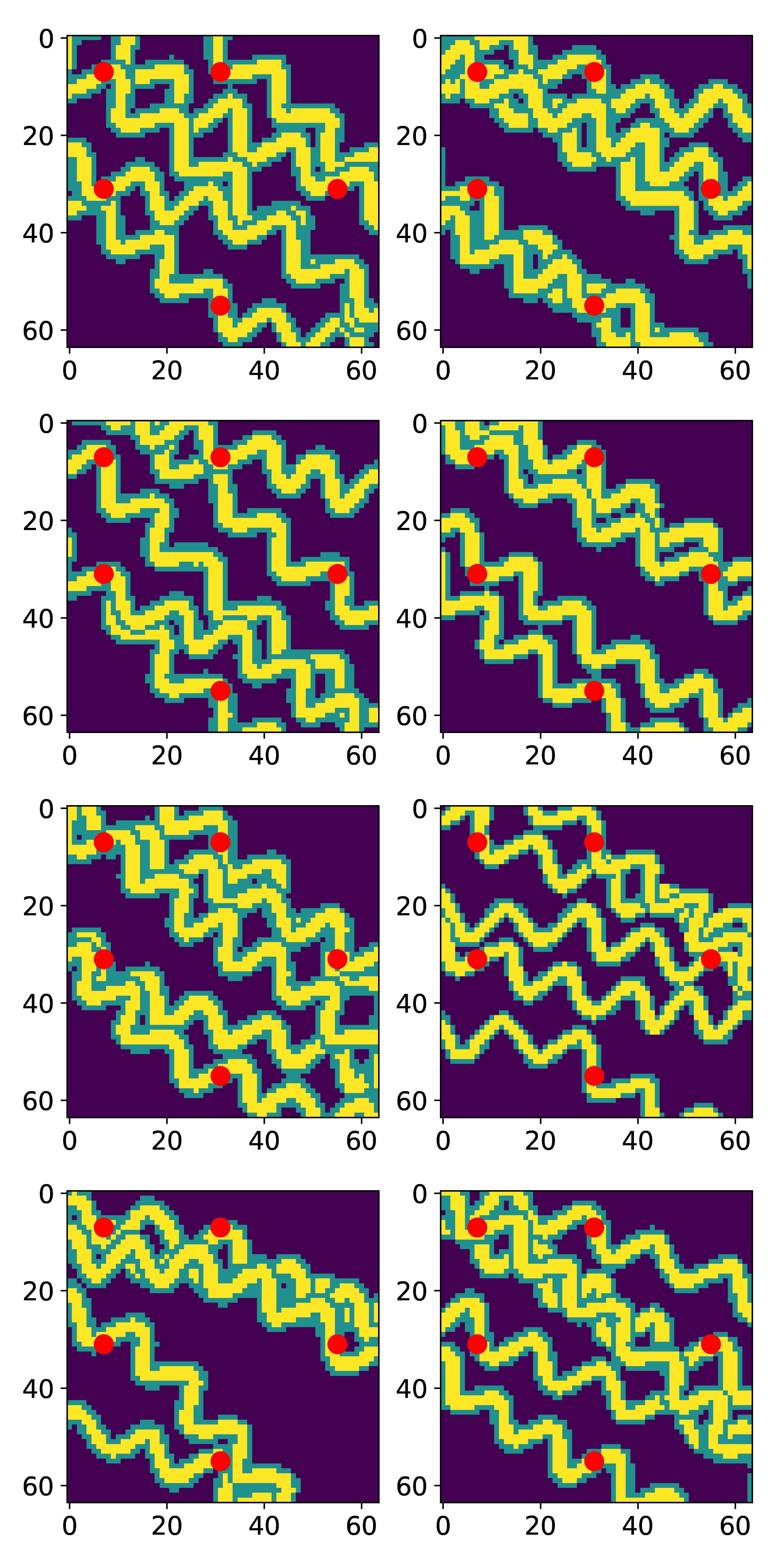}
        \caption{Petrel models}
        \label{fig:samples_petrel}
    \end{subfigure}
    \begin{subfigure}[b]{0.3\textwidth}
        \centering
        \includegraphics[width=\textwidth, trim=0 0 0 0, clip]{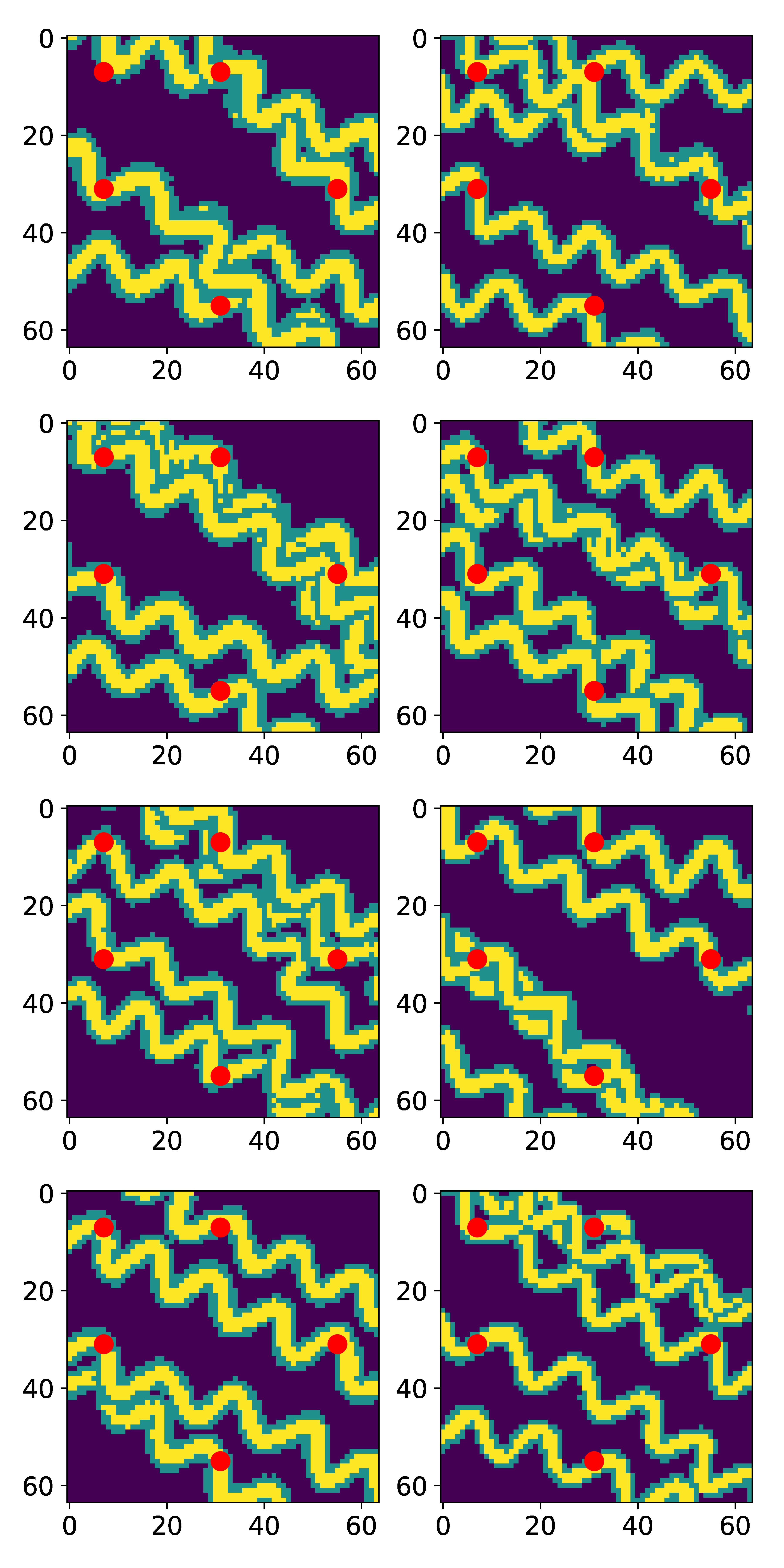}
        \caption{LDM-generated models}
        \label{fig:samples_dm}
    \end{subfigure}
    \caption{Randomly selected Petrel and LDM-generated realizations of the three-facies system. Conditioning points shown in red.}
    \label{fig:samples}
\end{figure}

Randomly selected LDM and Petrel geomodels are shown in Figure~\ref{fig:samples}. These models are all generated independently, so image-to-image correspondence between the Petrel (Figure~\ref{fig:samples}a) and LDM (Figure~\ref{fig:samples}b) realizations is not expected. We do, however, observe visual consistency in the key features, such as channel continuity, orientation, width and sinuosity. The facies ordering and thickness of the levee features are also consistent. The well facies conditioning is honored to 98\% accuracy in the LDM realizations.

\begin{figure}
    \centering
    \begin{subfigure}[b]{0.45\textwidth}
        \centering
        \includegraphics[width=\textwidth, trim=0 0 0 0, clip]{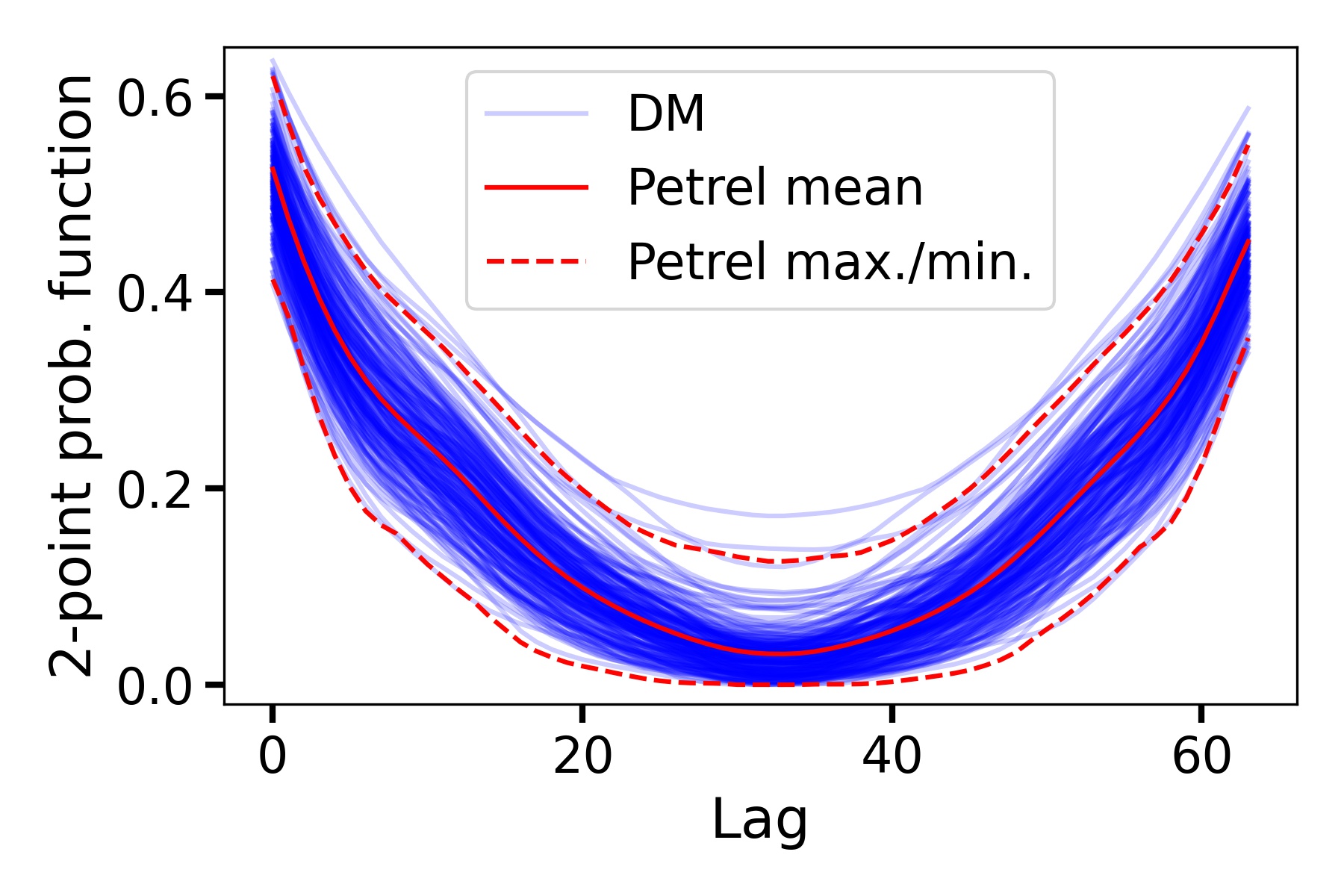}
        \caption{Mud facies, along horizontal axis}
        \label{fig:2p_prob_fun_x_background}
    \end{subfigure}
    \begin{subfigure}[b]{0.45\textwidth}
        \centering
        \includegraphics[width=\textwidth, trim=0 0 0 0, clip]{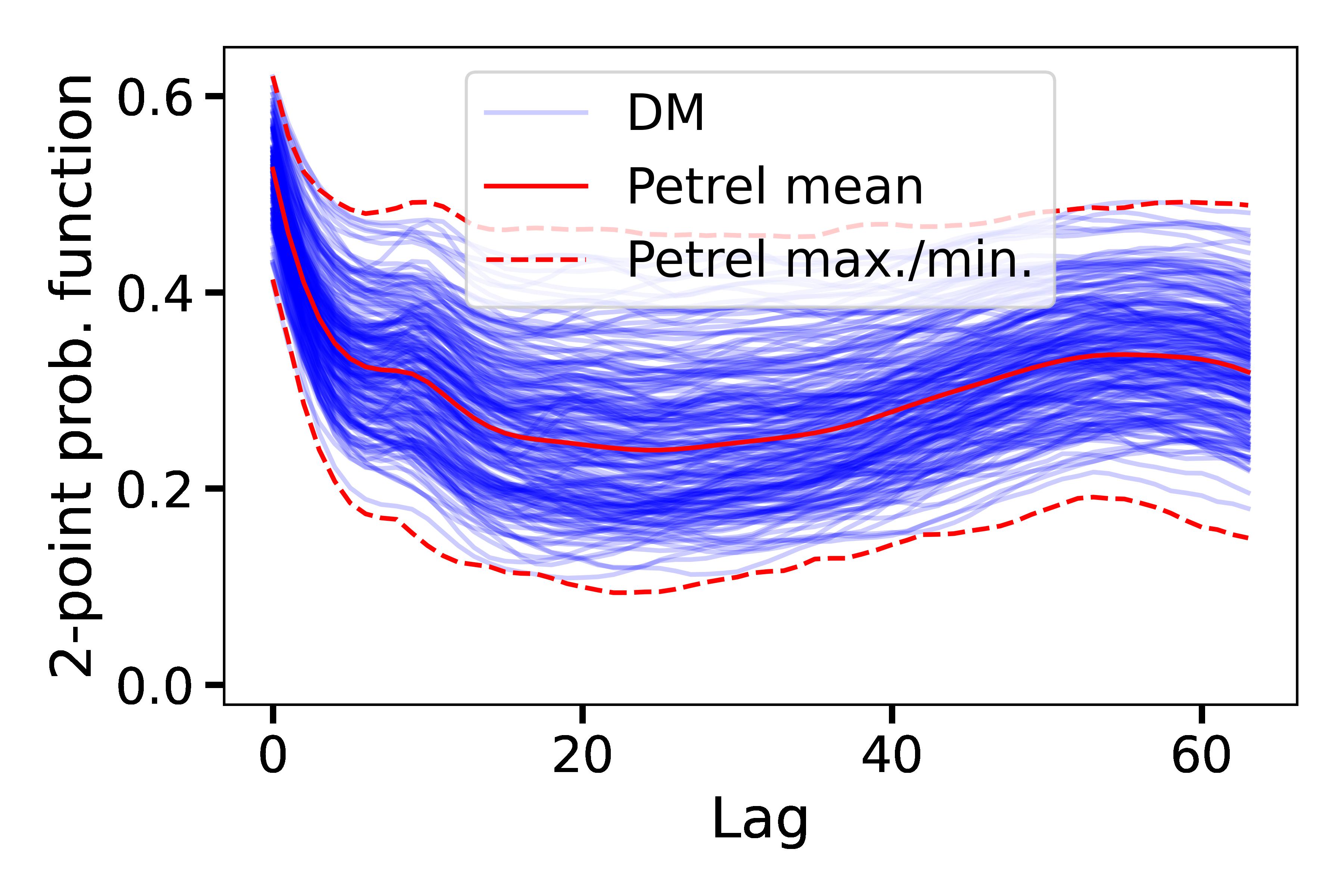}
        \caption{Mud facies, along diagonal}
        \label{fig:2p_prob_fun_dxy_background}
    \end{subfigure}
    \begin{subfigure}[b]{0.45\textwidth}
        \centering
        \includegraphics[width=\textwidth, trim=0 0 0 0, clip]{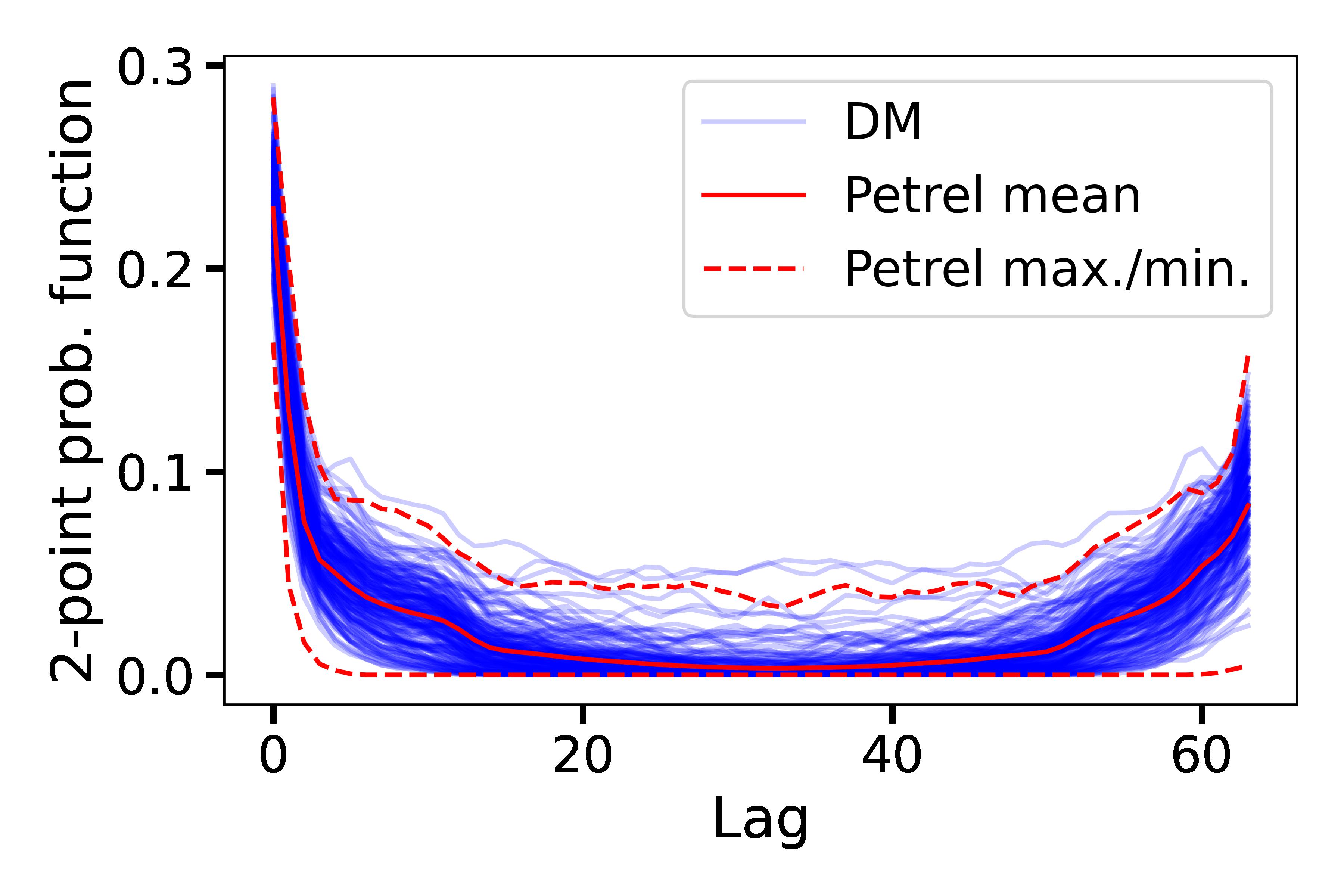}
        \caption{Levee facies, along horizontal axis}
        \label{fig:2p_prob_fun_x_levee}
    \end{subfigure}
    \begin{subfigure}[b]{0.45\textwidth}
        \centering
        \includegraphics[width=\textwidth, trim=0 0 0 0, clip]{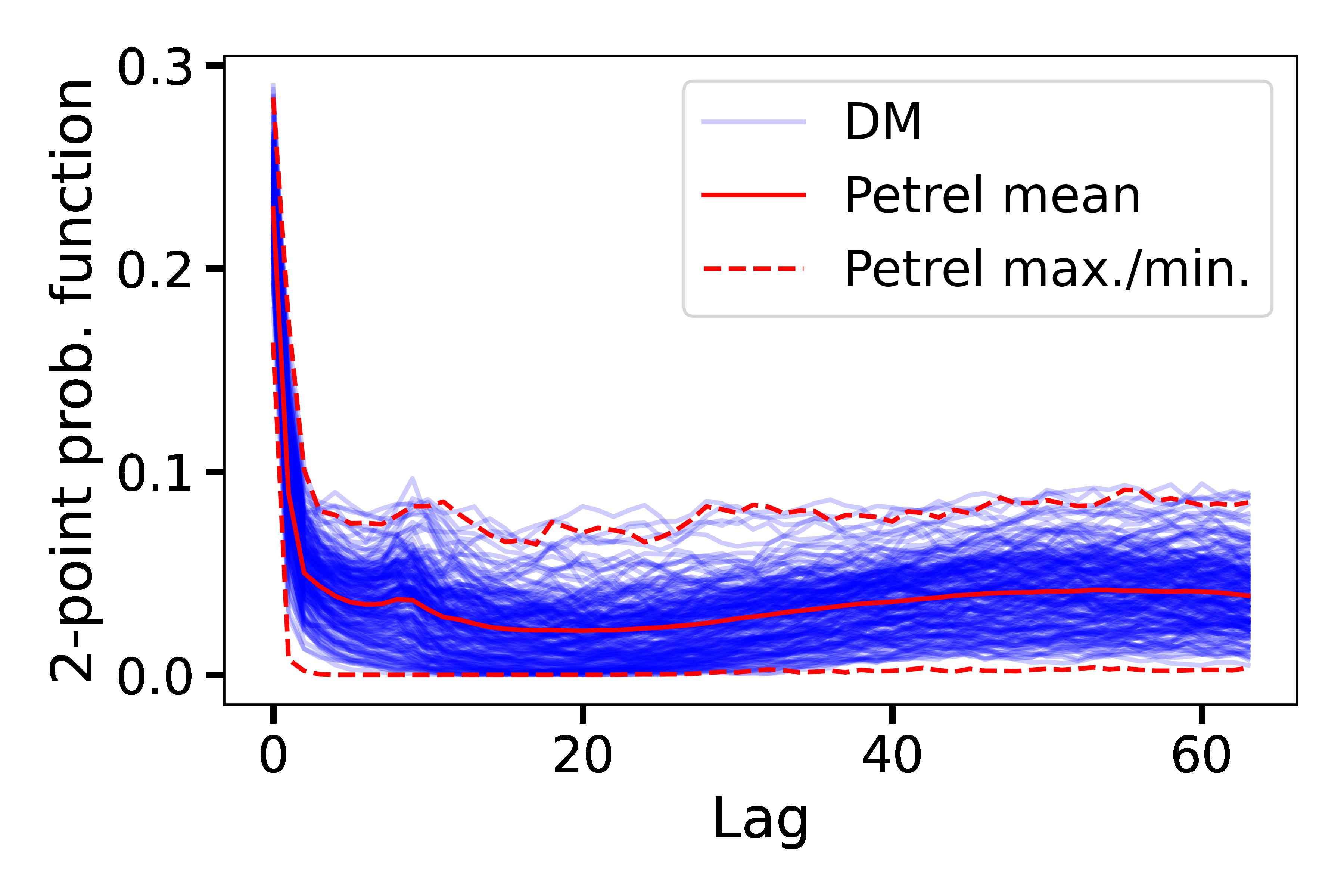}
        \caption{Levee facies, along diagonal}
        \label{fig:2p_prob_fun_dxy_levee}
    \end{subfigure}
    \begin{subfigure}[b]{0.45\textwidth}
        \centering
        \includegraphics[width=\textwidth, trim=0 0 0 0, clip]{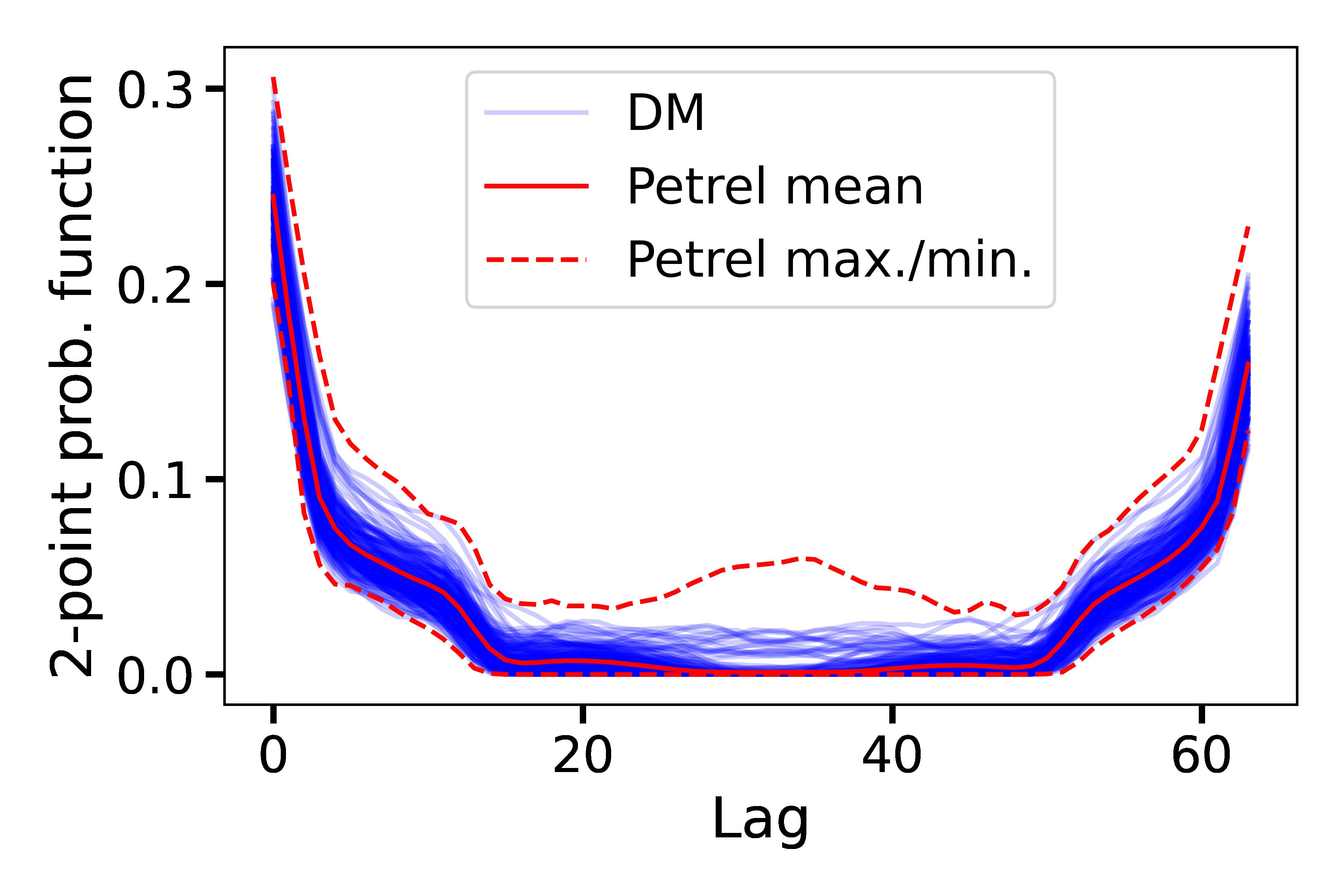}
        \caption{Channel facies, along horizontal axis}
        \label{fig:2p_prob_fun_x_channel_all}
    \end{subfigure}
    \begin{subfigure}[b]{0.45\textwidth}
        \centering
        \includegraphics[width=\textwidth, trim=0 0 0 0, clip]{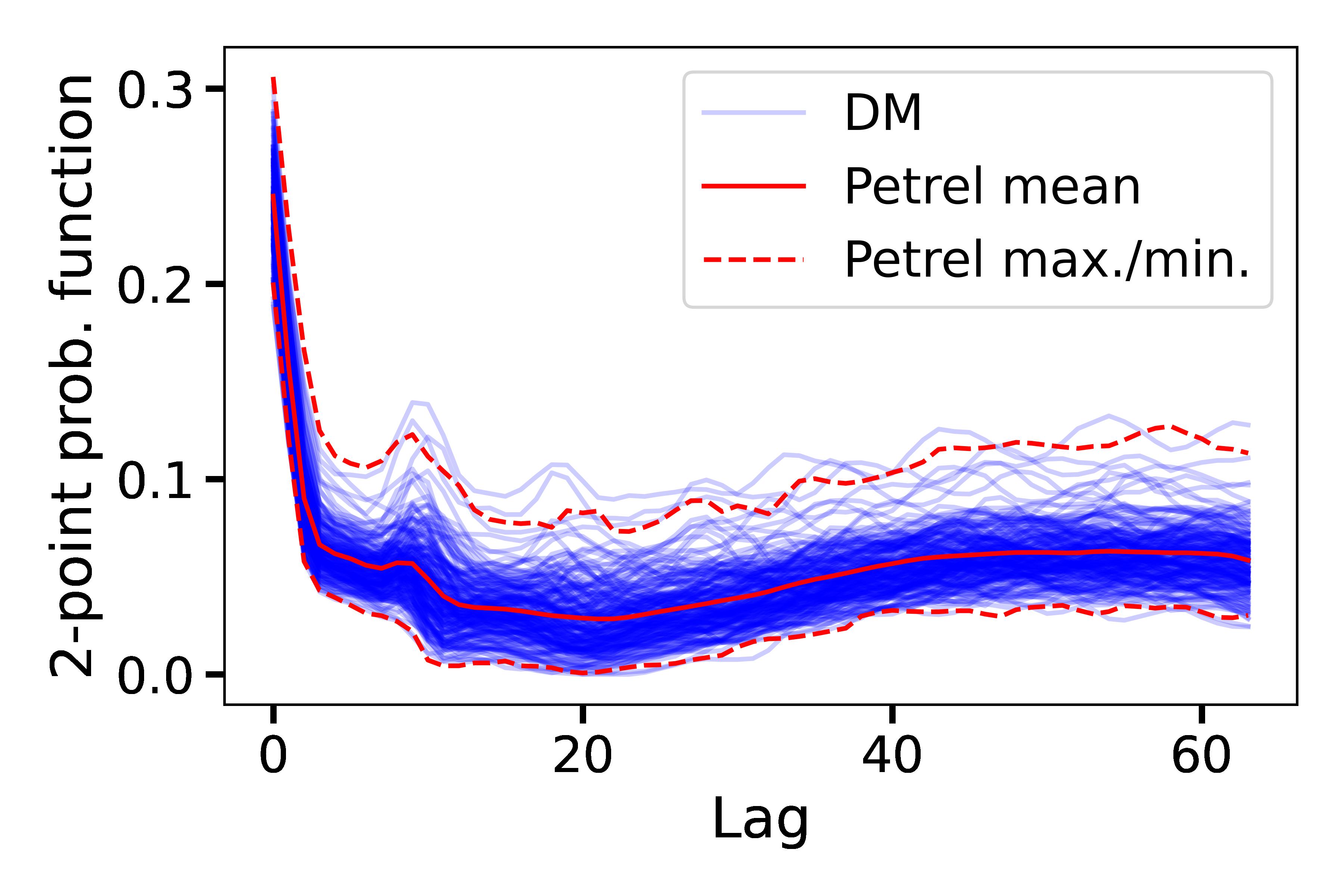}
        \caption{Channel facies, along diagonal}
        \label{fig:2p_prob_fun_dxy_channel_all}
    \end{subfigure}
    \caption{Comparisons of two-point probability functions, with lag distance in grid blocks. Red solid and dashed curves denote mean, maximum, and minimum value at each lag distance for 200 new Petrel models, while blue curves show individual results for 200 new LDM-generated models.
    \label{fig:prob_fun}
}
\end{figure}

We now consider quantitative similarity metrics. First, we compute two-point connectivity functions for the two sets of realizations. This function gives the probability that two points separated by a given lag distance, along a specified direction, belong to the same facies~\citep{Laloy_2018}. Results are presented in Figure~\ref{fig:prob_fun} for the mud, levee, and channel facies along the horizontal and diagonal ($y=x$) directions. We observe near-perfect agreement for the P$_{10}$, P$_{50}$, P$_{90}$ (percentile) results for these quantities. Rather than show these percentile results, we follow \citet{Laloy_2018} and present mean, maximum, and minimum values at each lag for the reference Petrel models, along with individual curves for LDM-generated models. The general agreement between the two sets of results in  Figure~\ref{fig:prob_fun} indicates that the overall orientation and connectivity for the mud, levee, and channel facies are accurately captured in the LDM realizations.

To achieve a controllable (and thus efficient) exploration of the latent space during history matching, the generative model should respond smoothly to small perturbations in the input. To test the smoothness of the LDM parameterization, we consider two random latent inputs, $\bm{\xi}_{T}^{(1)}$ and $\bm{\xi}_{T}^{(2)}$.
Note that, since subscripts were originally defined for the forward (noising) process, as in Figure~\ref{fig:noising_denoising}, generation (denoising) starts from step $T$. Their corresponding generated models are denoted $\mathbf{m}_{DM}^{(1)}$ and $\mathbf{m}_{DM}^{(2)}$. A linear interpolation between the two inputs gives $\bm{\xi}_T(\delta) = \bm{\xi}_{T}^{(1)}(1-\delta) + \bm{\xi}_T^{(2)} \delta$, where the parameter $\delta$ can take any value between 0 and 1. If the parameterization is relatively smooth, we expect to see a degree of continuity in the variation from $\mathbf{m}_{DM}^{(1)}$ and $\mathbf{m}_{DM}^{(2)}$, even though these models involve discrete features.

\begin{figure} 
    \centering
    \begin{subfigure}[b]{0.9\textwidth}
        \centering
        \includegraphics[width=\textwidth, trim=0 0 0 5cm, clip]{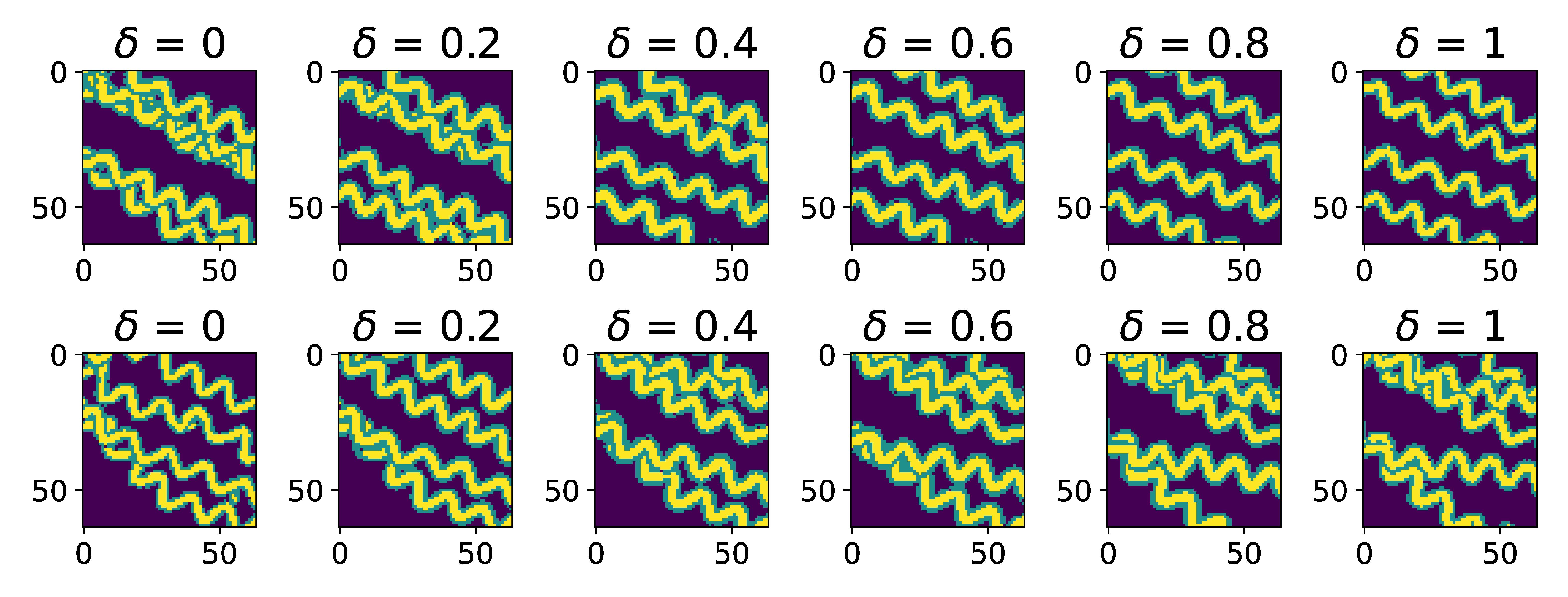}
        \caption{Example~1}
        \label{fig:interp1_models}
    \end{subfigure}
    \begin{subfigure}[b]{0.9\textwidth}
        \centering
        \includegraphics[width=\textwidth, trim=0 5cm 0 0, clip]{interpolation.jpg}
        \caption{Example~2}
        \label{fig:interp2_models}
    \end{subfigure}
    \caption{Interpolation tests on the LDM parameterization. Intermediate models generated by linearly interpolating the latent variable between the two end-member realizations ($\delta = 0$ and $\delta = 1$).
    \label{fig:interpolation_models}
}
\end{figure}

Two interpolation examples are shown in Figure~\ref{fig:interpolation_models}. There are clear differences between the end-members $\mathbf{m}_{DM}^{(1)}$ and $\mathbf{m}_{DM}^{(2)}$. In these examples we specify $\delta=0.2$, 0.4, 0.6 and 0.8. Reasonably smooth transitions are observed in both cases. More specifically, in Example~1, we see the gradual aggregation of the upper two channels along with the gradual spreading of most portions of the lower two channels. In Example~2, channels aggregate at both the top and bottom in $\mathbf{m}_{DM}^{(1)}$. These two sets of aggregated channels separate gradually into four distinct channels with increasing $\delta$.

\begin{figure}
    \centering
    \begin{subfigure}[b]{0.45\textwidth}
        \centering
        \includegraphics[width=\textwidth, trim=0 0 0 0, clip]{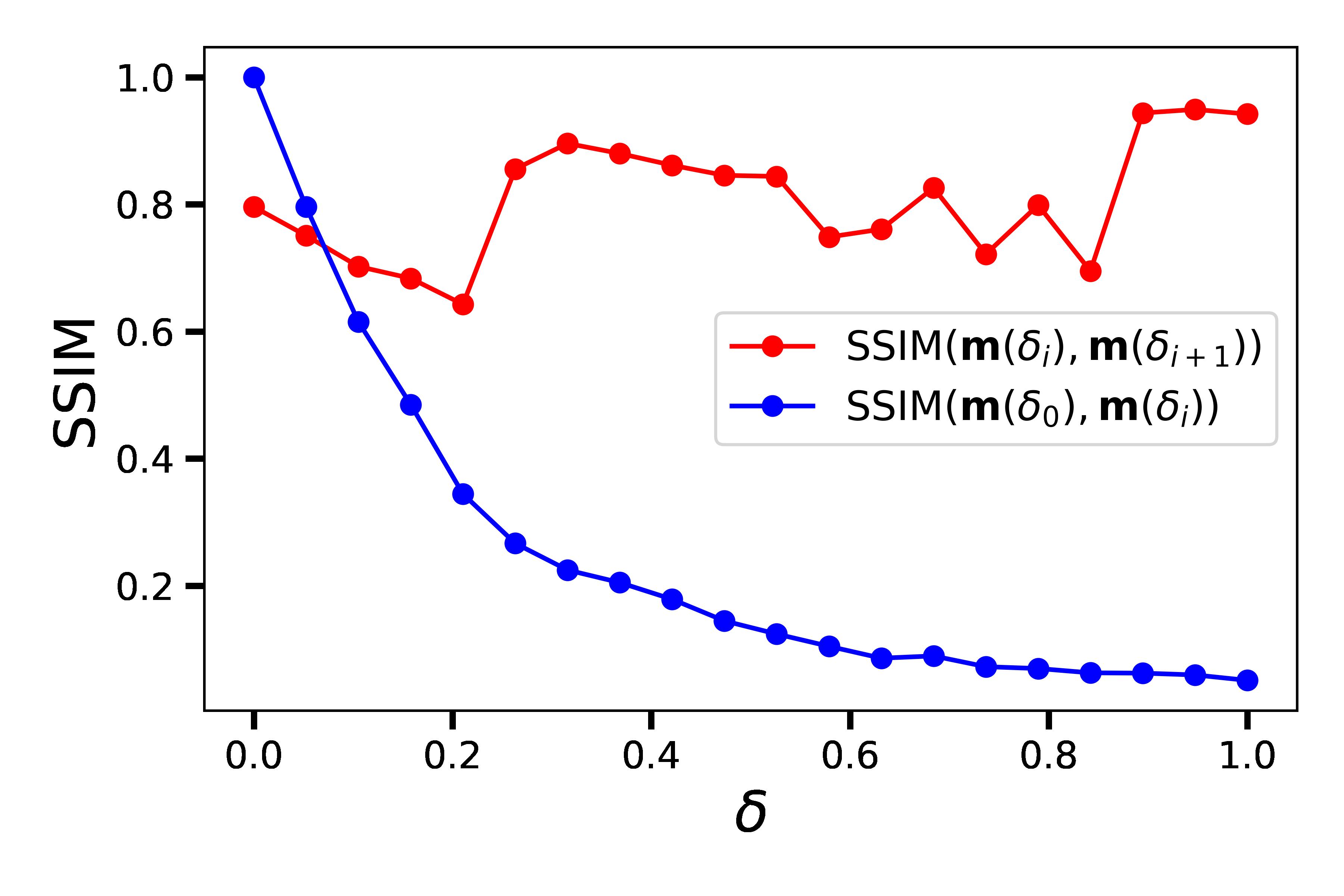}
        \caption{Example~1}
        \label{fig:interp1_ssim}
    \end{subfigure}
    \begin{subfigure}[b]{0.45\textwidth}
        \centering
        \includegraphics[width=\textwidth, trim=0 0 0 0, clip]{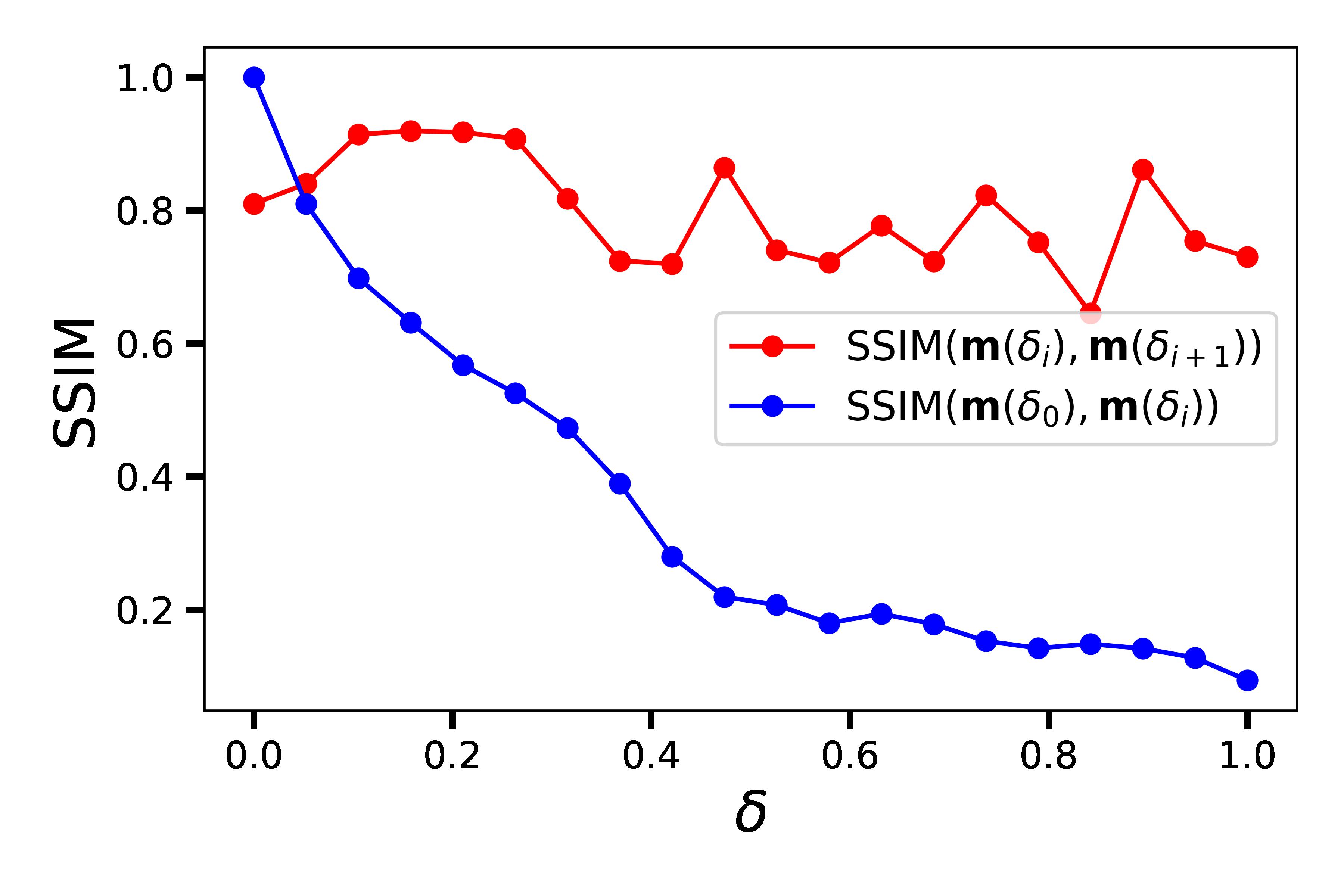}
        \caption{Example~2}
        \label{fig:interp2_ssim}
    \end{subfigure}
    \caption{Structural similarity index measure (SSIM) for the interpolation tests. Red curve shows the SSIM computed between consecutive models, and blue curve shows SSIM computed between the first model and models at each $\delta$ value.
    \label{fig:interpolation_ssim}
}
\end{figure}

The similarity between models can be quantified using the structural similarity index measure (SSIM), a widely applied image similarity metric~\citep{ssim}. SSIM ranges from 1 (for identical images) to -1. Results for the two cases are shown in Figure~\ref{fig:interpolation_ssim}. In these results we use increments in $\delta$ of 0.05. The red curves provide SSIM computed between consecutive models (i.e., separated by $\delta=0.05$), while the blue curves give SSIM between the first model $\mathbf{m}_{DM}^{(1)}$ and the model corresponding to each $\delta$ value. Although there is some variability in the red curves we do observe clear consistency from model to model, with SSIM values of around 0.8. The blue curves display a smooth transition away from $\mathbf{m}_{DM}^{(1)}$. Both sets of behaviors are desirable in history matching settings and suggest that the LDM should be applicable in this context.

The next assessment involves comparisons of flow responses. This type of test is important, as it allows us to gauge whether the LDM realizations provide flow results that are similar, in terms of the distributions in injection and production rates through time, to those of the reference Petrel geomodels. 

\begin{figure} 
    \centering
    \begin{subfigure}[b]{0.4\textwidth}
        \centering
        \includegraphics[width=\textwidth, trim=0 0 0 0, clip]{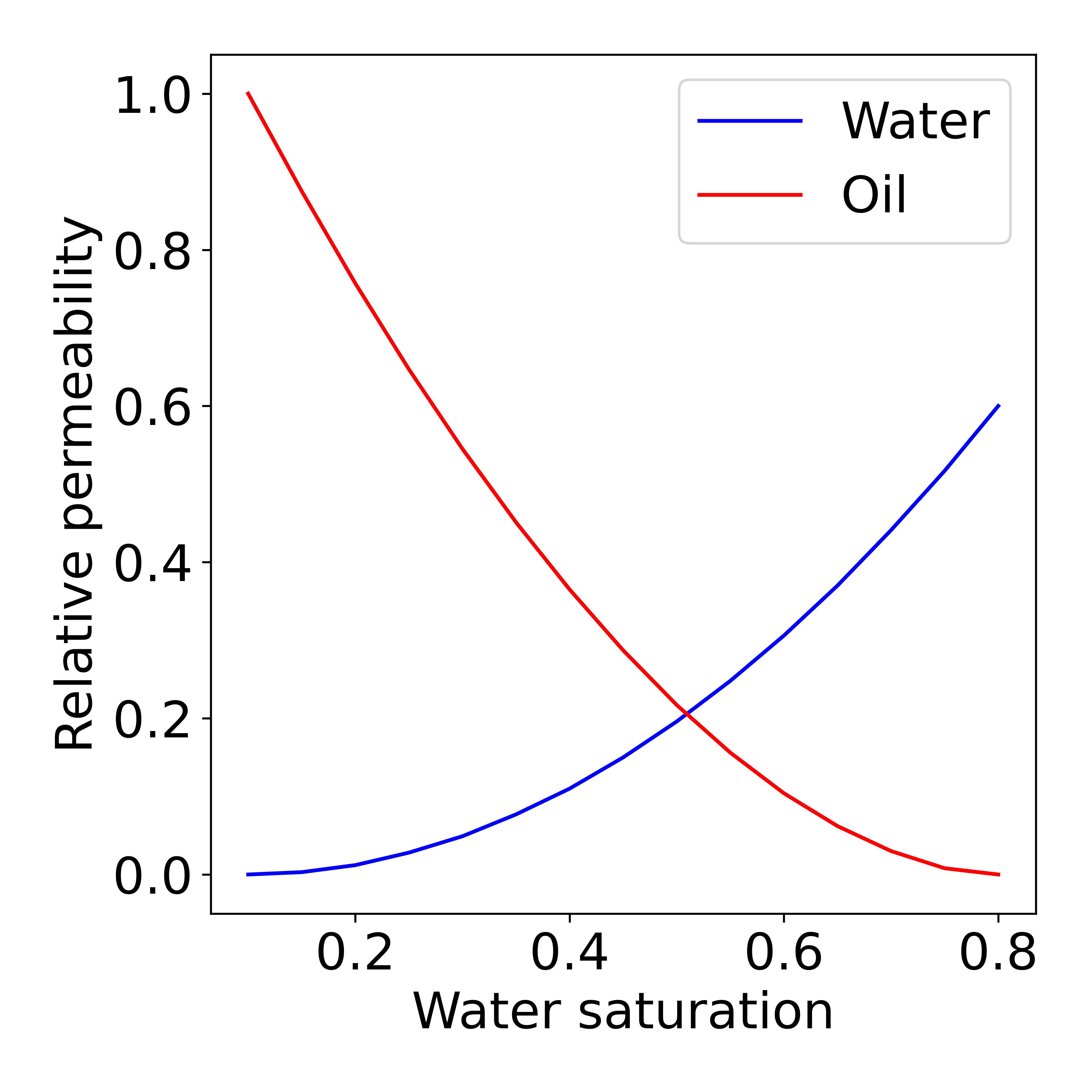}
        \caption{Oil-water relative permeabilities}
        \label{fig:rel_perms}
    \end{subfigure}
    \begin{subfigure}[b]{0.3\textwidth}
        \centering
        \includegraphics[width=\textwidth, trim=1cm 0 1.8cm 0, clip]{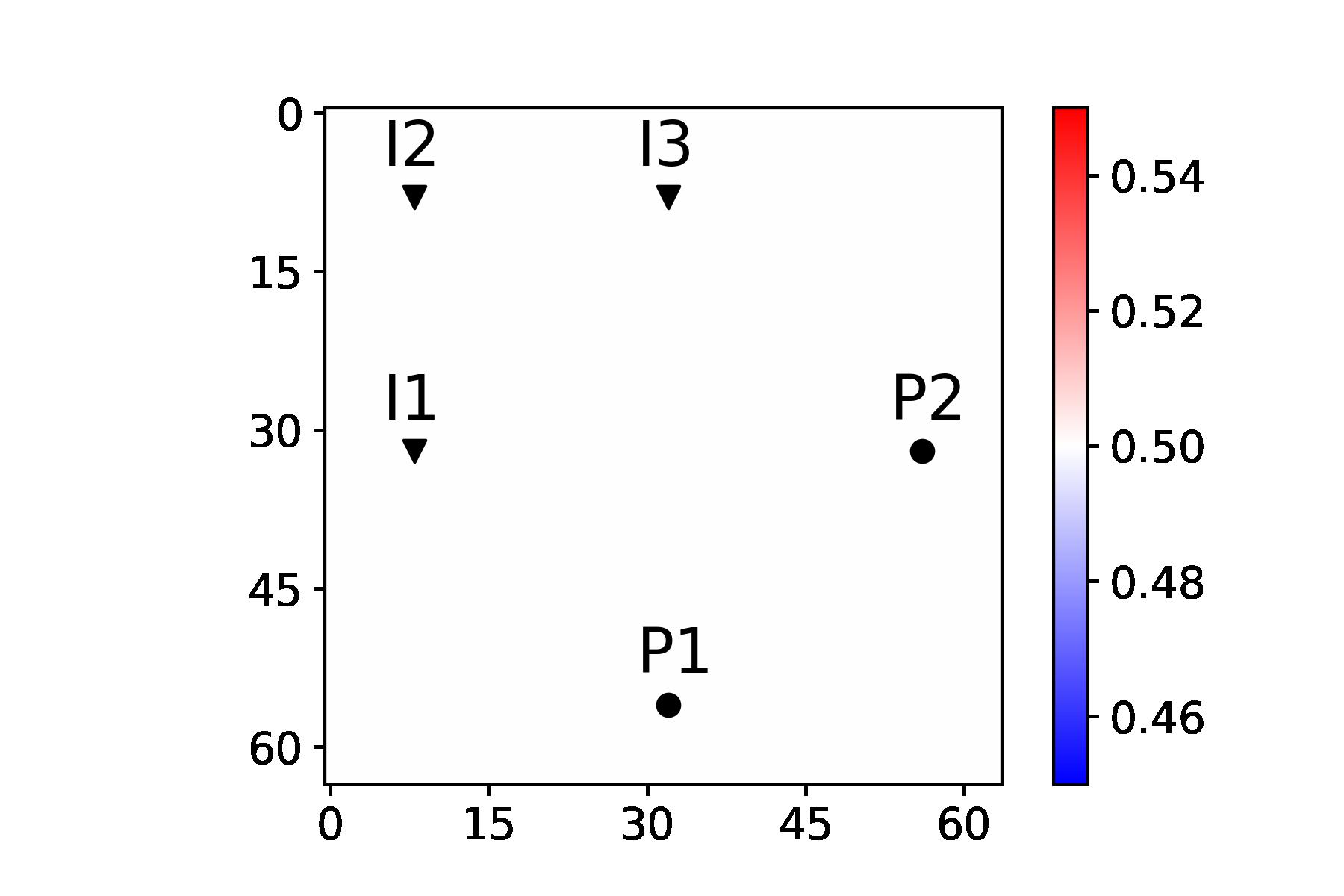}
        \vspace{0.7cm}
        \caption{Injection and production wells}
        \label{fig:well_locs}
    \end{subfigure}
    \caption{Relative permeability curves and well locations used for all simulations.
    \label{fig:sim_info}
}
\end{figure}

The setup for the flow problem is as follows. The geomodels contain $64 \times 64$ cells, with each grid block of dimensions 20~m $\times$ 20~m $\times$ 5~m (in the $x$, $y$ and $z$ directions, respectively). We consider two-phase oil-water flow. The relative permeability curves used in all simulations are shown in Figure~\ref{fig:rel_perms}. Water viscosity is constant at 0.31~cP. Oil viscosity is a function of pressure, with a value of 1.09~cP at 325~bar. The initial water saturation and initial reservoir pressure are 0.1 and 310~bar. Porosity and permeability are constant within each facies, at 0.2 and 2500~mD for the channel facies, 0.15 and 400~mD for the levee facies, and 0.05 and 50~mD for the mud facies. There are a total of five wells -- three injection wells and two production wells -- at the conditioning locations, as shown in Figure~\ref{fig:well_locs}. Wells are BHP controlled, with injectors at 330~bar and producers at 300~bar. The simulations are run for a period of 2500~days, with maximum time steps of 50~days. All runs are performed using Stanford’s Automatic Differentiation General Purpose Research Simulator, ADGPRS~\citep{zhou2012parallel}.

\begin{figure}
    \centering
    \begin{subfigure}[b]{0.45\textwidth}
        \centering
        \includegraphics[width=\textwidth]{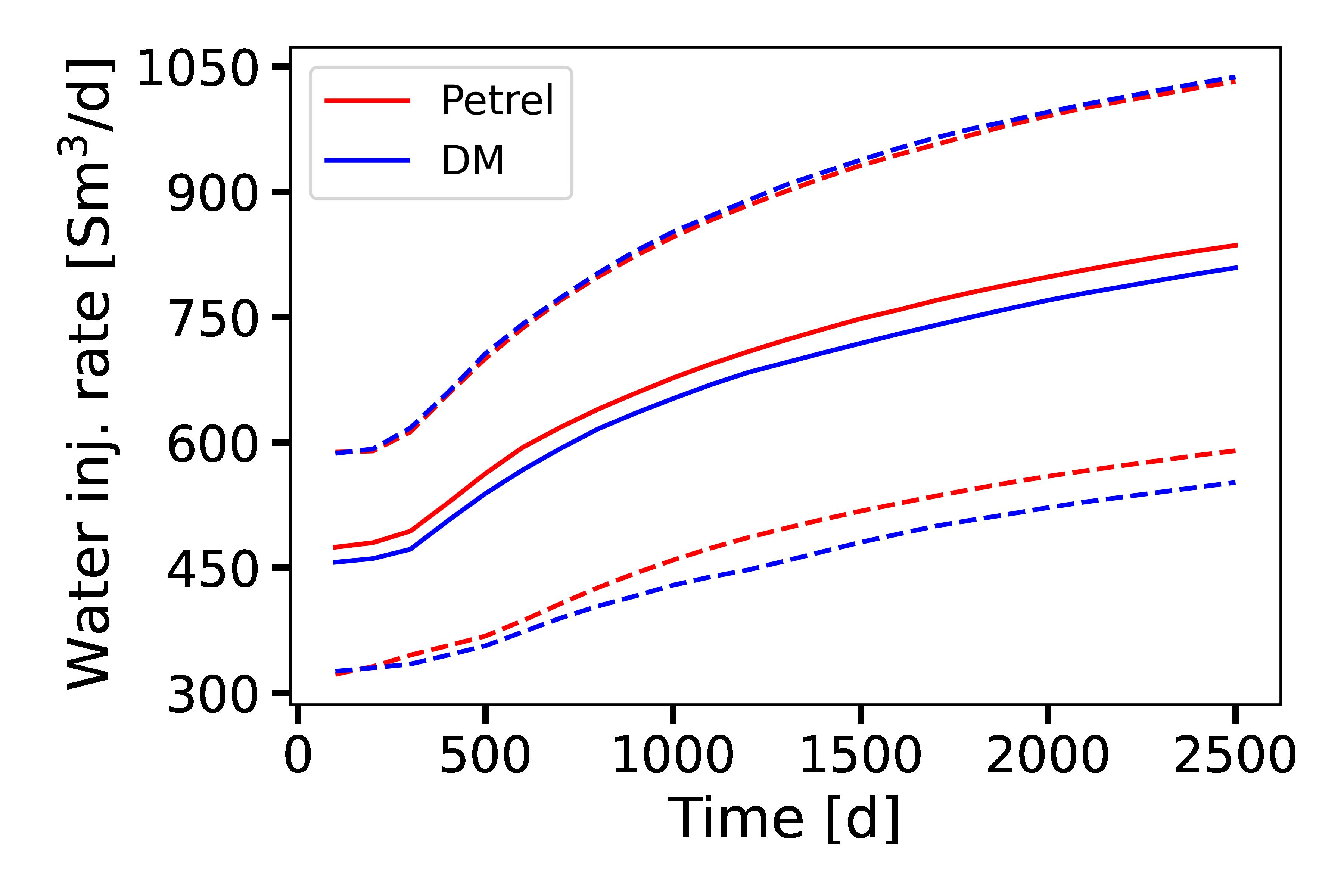}
        \caption{Field water injection rate}
        \label{fig:field_inj}
    \end{subfigure}
    \begin{subfigure}[b]{0.45\textwidth}
        \centering
        \includegraphics[width=\textwidth]{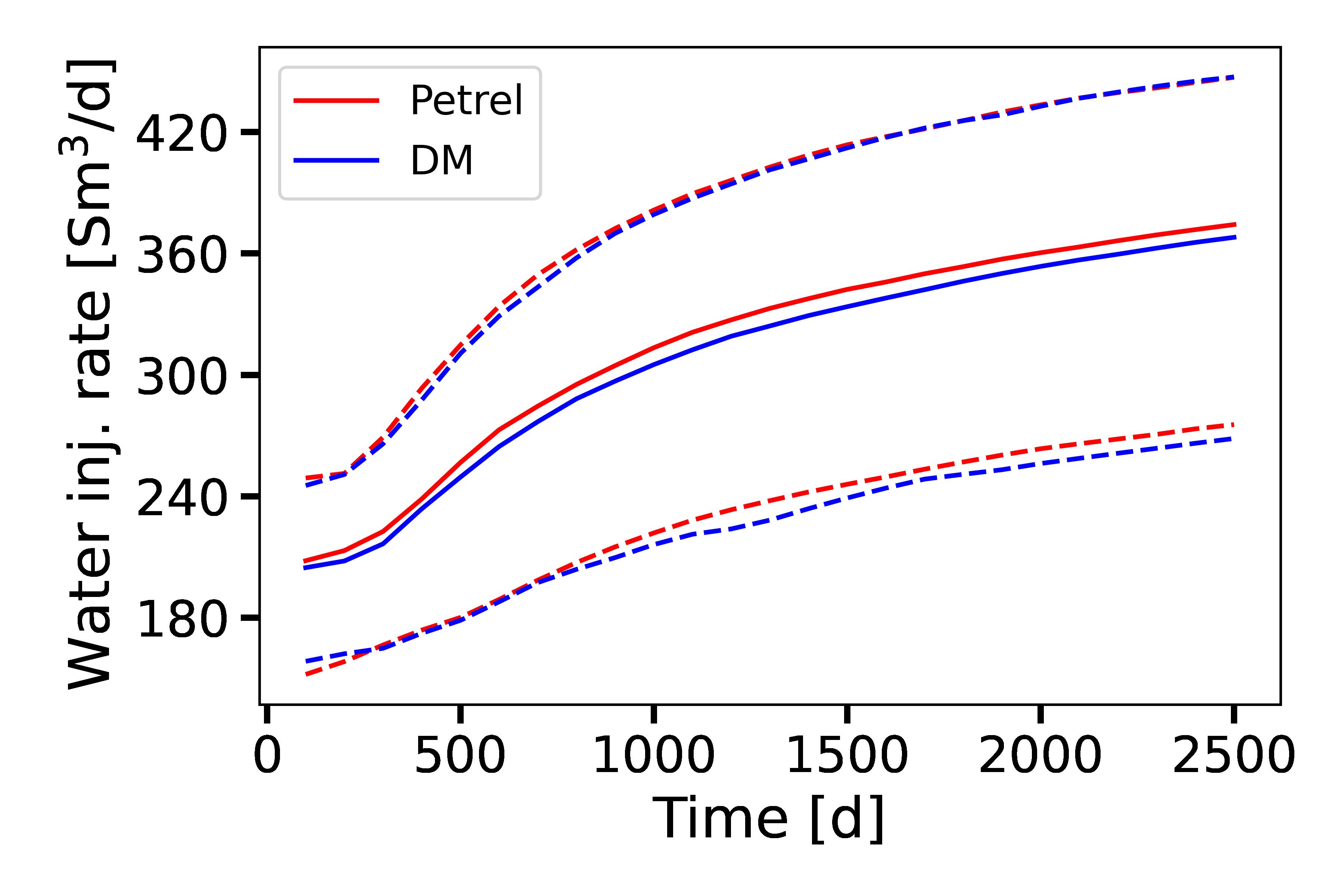}
        \caption{I1 water injection rate (highest injection)}
        \label{fig:I1_WAT}
    \end{subfigure}
    \begin{subfigure}[b]{0.45\textwidth}
        \centering
        \includegraphics[width=\textwidth]{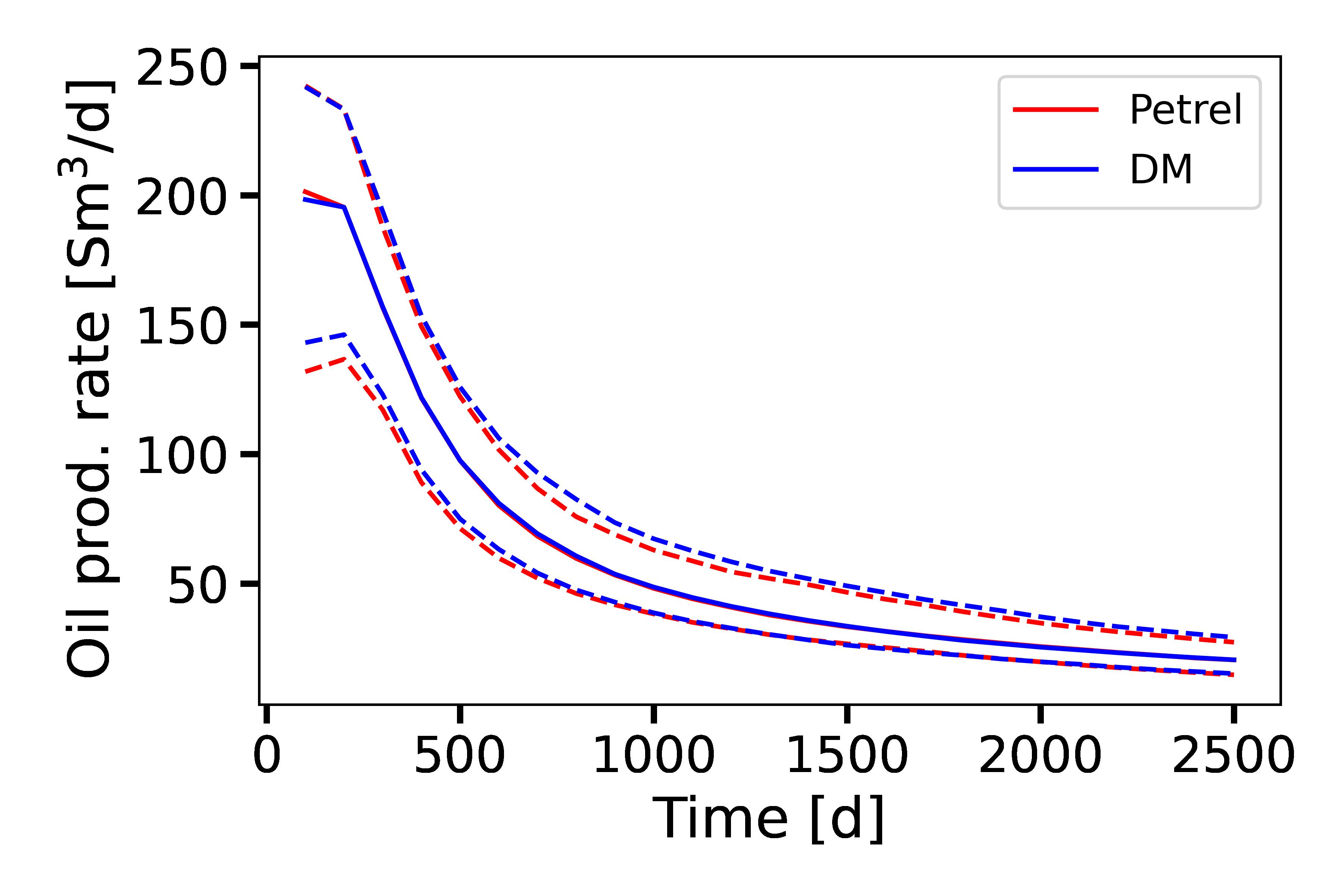}
        \caption{P1 oil production rate}
        \label{fig:P1_OIL}
    \end{subfigure}
    \begin{subfigure}[b]{0.45\textwidth}
        \centering
        \includegraphics[width=\textwidth]{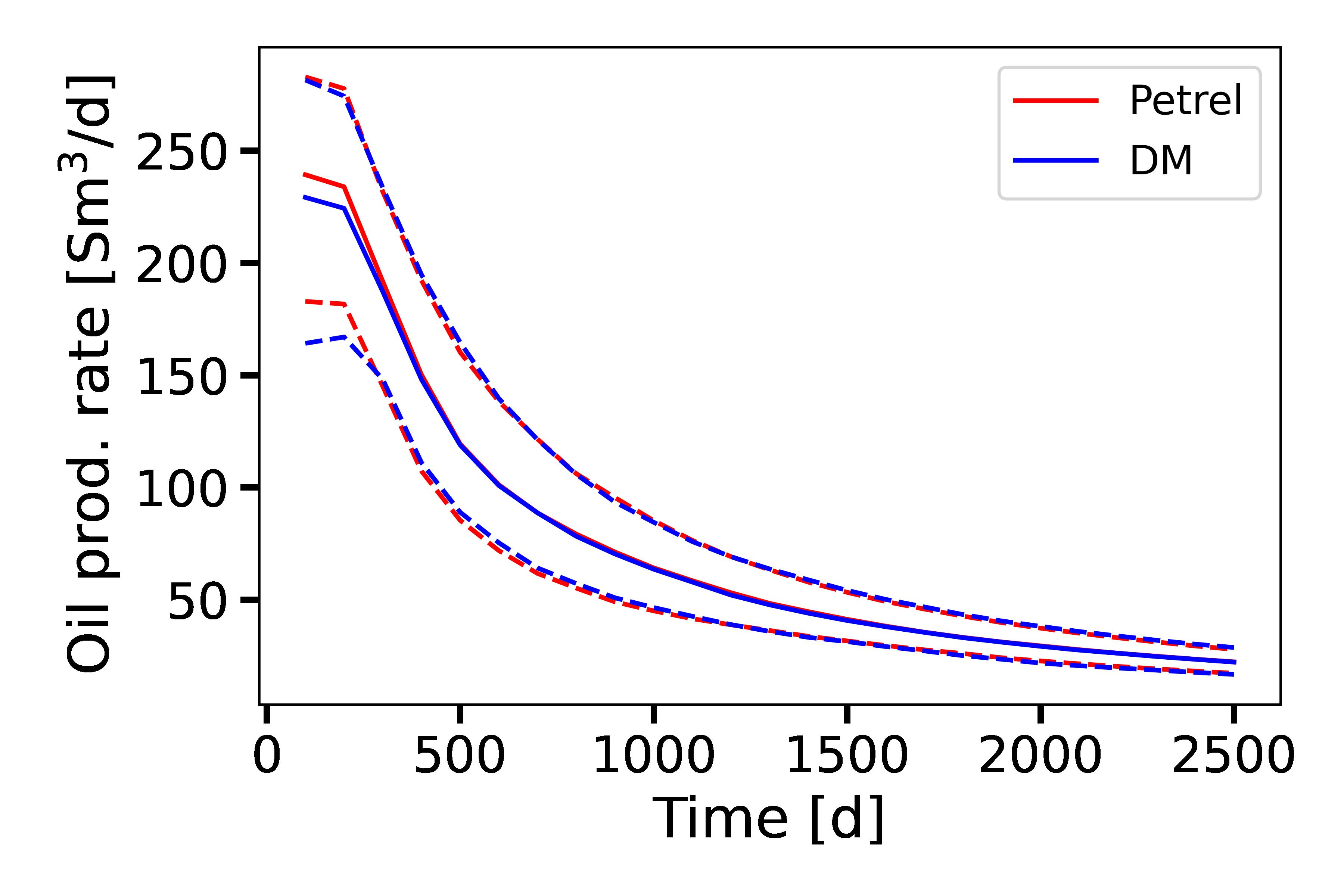}
        \caption{P2 oil production rate}
        \label{fig:P2_OIL}
    \end{subfigure}
    \begin{subfigure}[b]{0.45\textwidth}
        \centering
        \includegraphics[width=\textwidth]{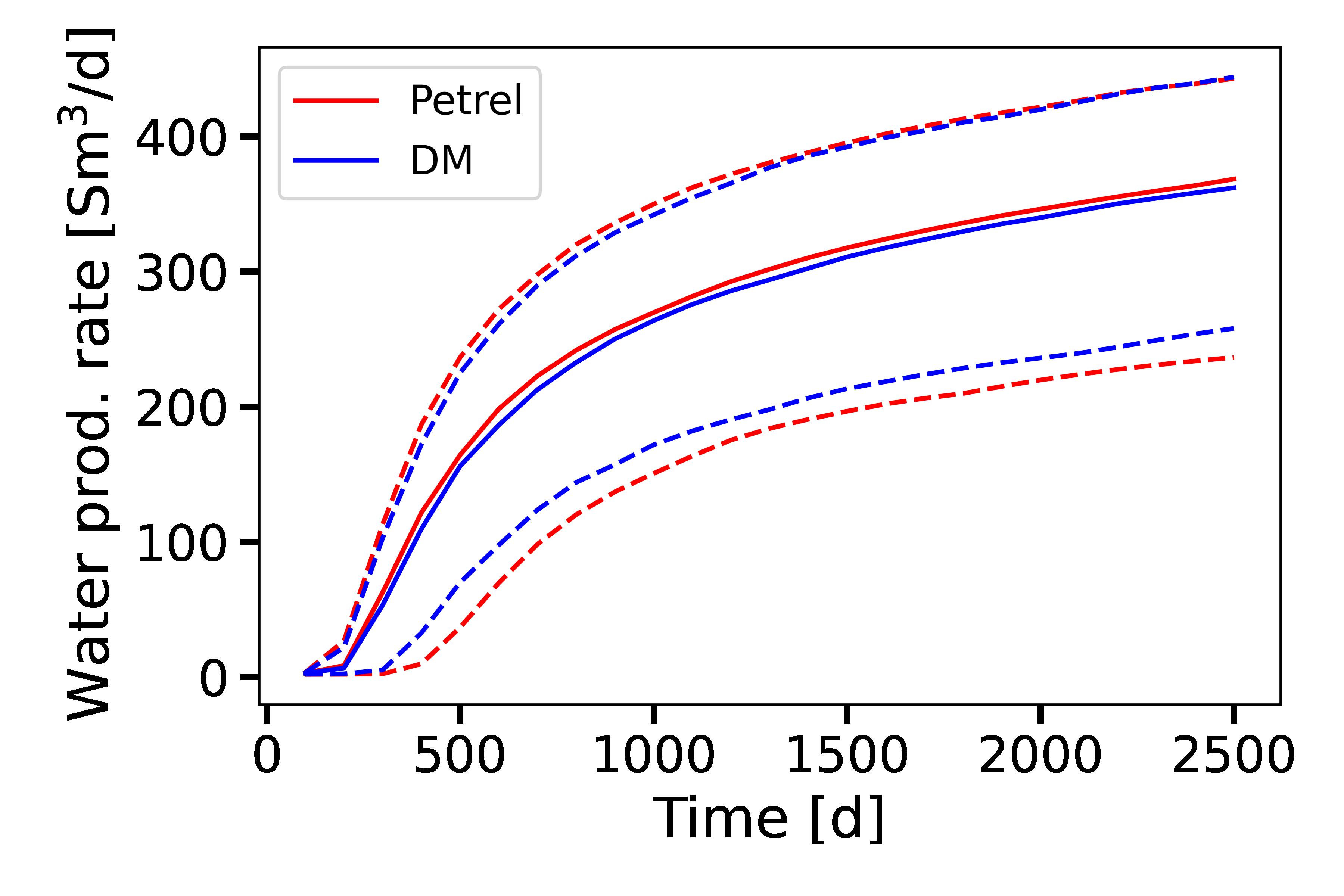}
        \caption{P1 water production rate}
        \label{fig:P1_WAT}
    \end{subfigure}
    \begin{subfigure}[b]{0.45\textwidth}
        \centering
        \includegraphics[width=\textwidth]{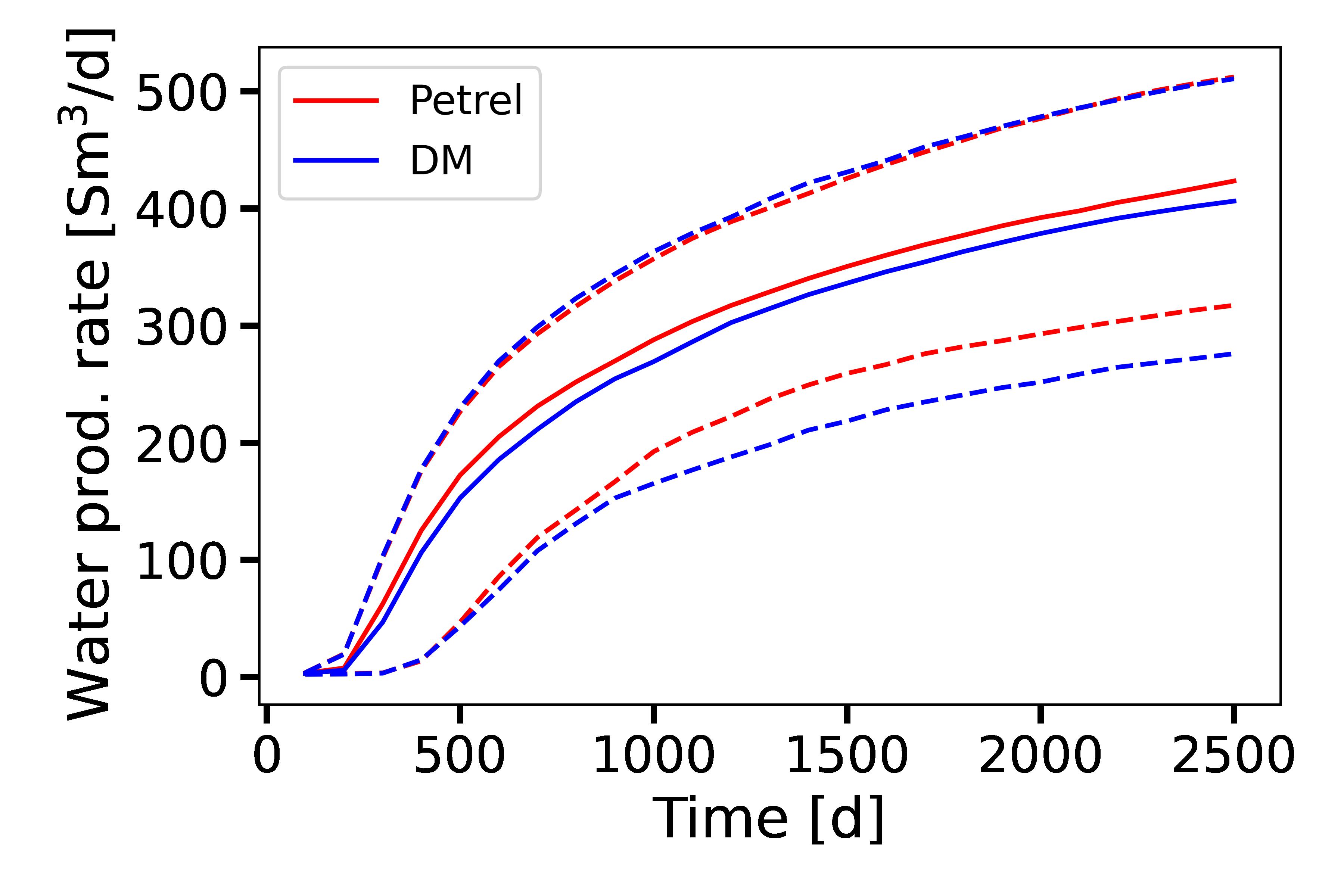}
        \caption{P2 water production rate}
        \label{fig:P2_WAT}
    \end{subfigure}
    \caption{Comparison of flow statistics for Petrel realizations (red curves) and LDM realizations (blue curves) over ensembles of 200 (new) test models. Solid curves denote P$_{50}$ results, lower and upper dashed curves are P$_{10}$ and P$_{90}$ results.
    \label{fig:flow_stats}
}
\end{figure}

Flow results, in terms of the P$_{10}$, P$_{50}$ and P$_{90}$ percentile curves for both sets of models, are shown in Figure~\ref{fig:flow_stats}. These curves are constructed from the results over the 200 models at each time step (different points on the curves correspond to different realizations, in general). Results are shown for field injection rate (Figure~\ref{fig:flow_stats}a), for the highest cumulative injection well, I1 (Figure~\ref{fig:flow_stats}b), and for oil and water production rates for both production wells (Figure~\ref{fig:flow_stats}c-f). We see nearly a factor of two difference between the P$_{10}$ and P$_{90}$ field injection rates, indicating a reasonable amount of variation between the realizations in terms of flow response. Importantly, the LDM geomodels accurately capture all of these flow responses in an overall sense. This suggests that the parameterization represents the key spatial features and associated variability in terms of their impact on flow. Some discrepancy in the P$_{10}$ water rates is evident, particularly for well P2 (Figure~\ref{fig:flow_stats}f). This may be due to a slight underestimation of the frequency or continuity of the channels connecting wells I3 and P2.

\section{History matching using the LDM}
\label{sec:results_hm}
We now apply the diffusion model parameterization within an ensemble-based history matching framework. Results will be presented for two cases. We denote the variables to be determined during history matching as $\boldsymbol{\xi}_{hm}$. In the first case the porosity and permeability for each facies are specified, so only the LDM variables ($\boldsymbol{\xi}_T$) are updated during data assimilation. Thus we have $\boldsymbol{\xi}_{hm} = \boldsymbol{\xi}_{T}$. In the second case, porosity and permeability in each facies are treated as uncertain (though uniform for the facies). In this case we have $\boldsymbol{\xi}_{hm} = [\boldsymbol{\xi}_T, \boldsymbol{\phi}, \bf{k}]$, where $\boldsymbol{\phi} = [\phi_{mud}, \phi_{levee}, \phi_{channel}]$ and ${\bf k} = [\log{k}_{mud}, \log{k}_{levee}, \log{k}_{channel}]$. Hence, the number of variables to be determined is $N_{hm} = n_c = n_x \times n_y = 64$ in the first case, and $N_{hm} = n_c + 6 = 70$ in the second case. 

\begin{figure}
    \centering
    \begin{subfigure}[b]{0.25\textwidth}
        \centering
        \includegraphics[width=\textwidth, trim=2.5cm 0 2.5cm 0cm, clip]{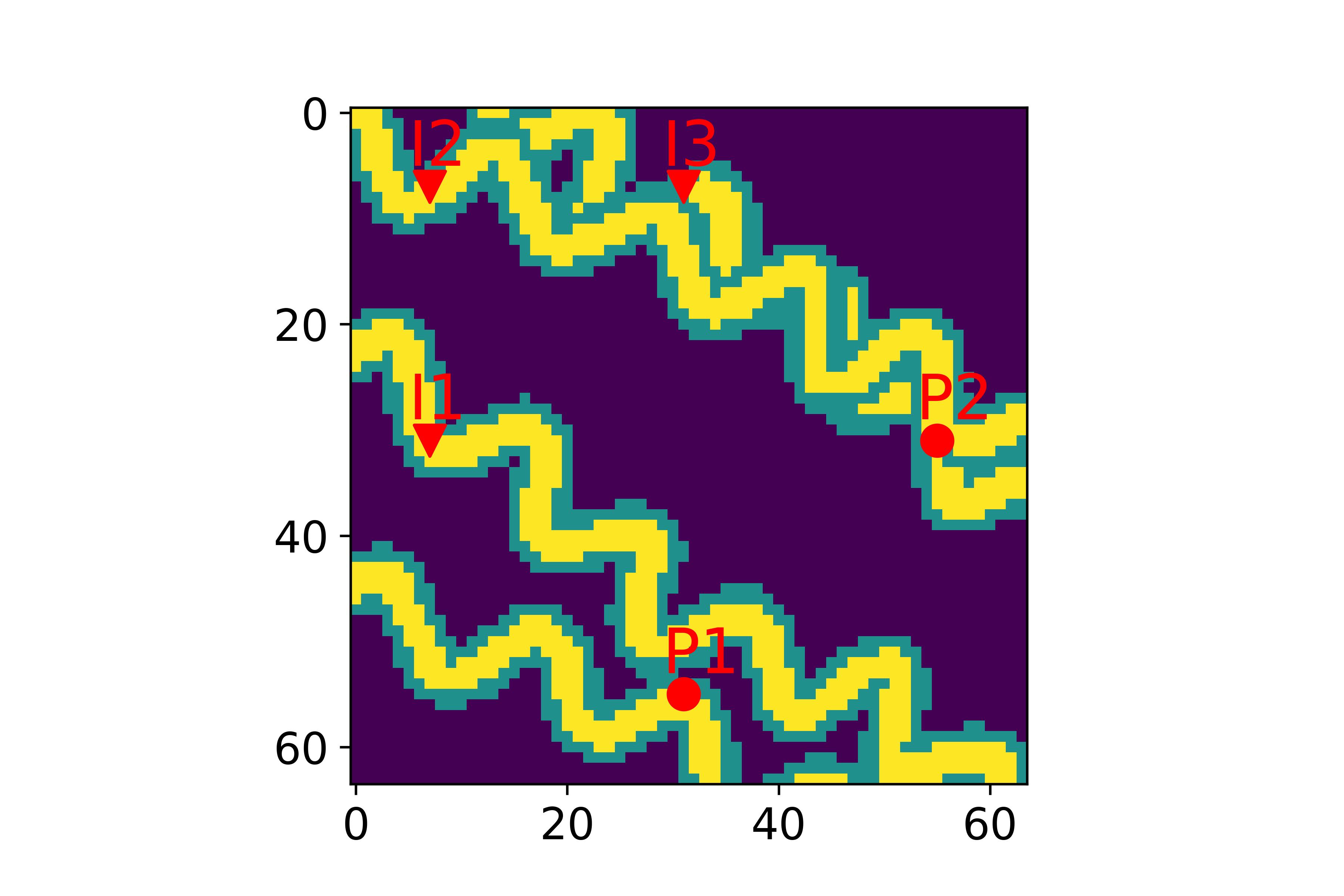}
        \subcaption{True model 1}
        \label{fig:true}
    \end{subfigure}
    \begin{subfigure}[b]{0.25\textwidth} 
        \centering
        \includegraphics[width=\textwidth, trim=2.5cm 0 2.5cm 0cm, clip]{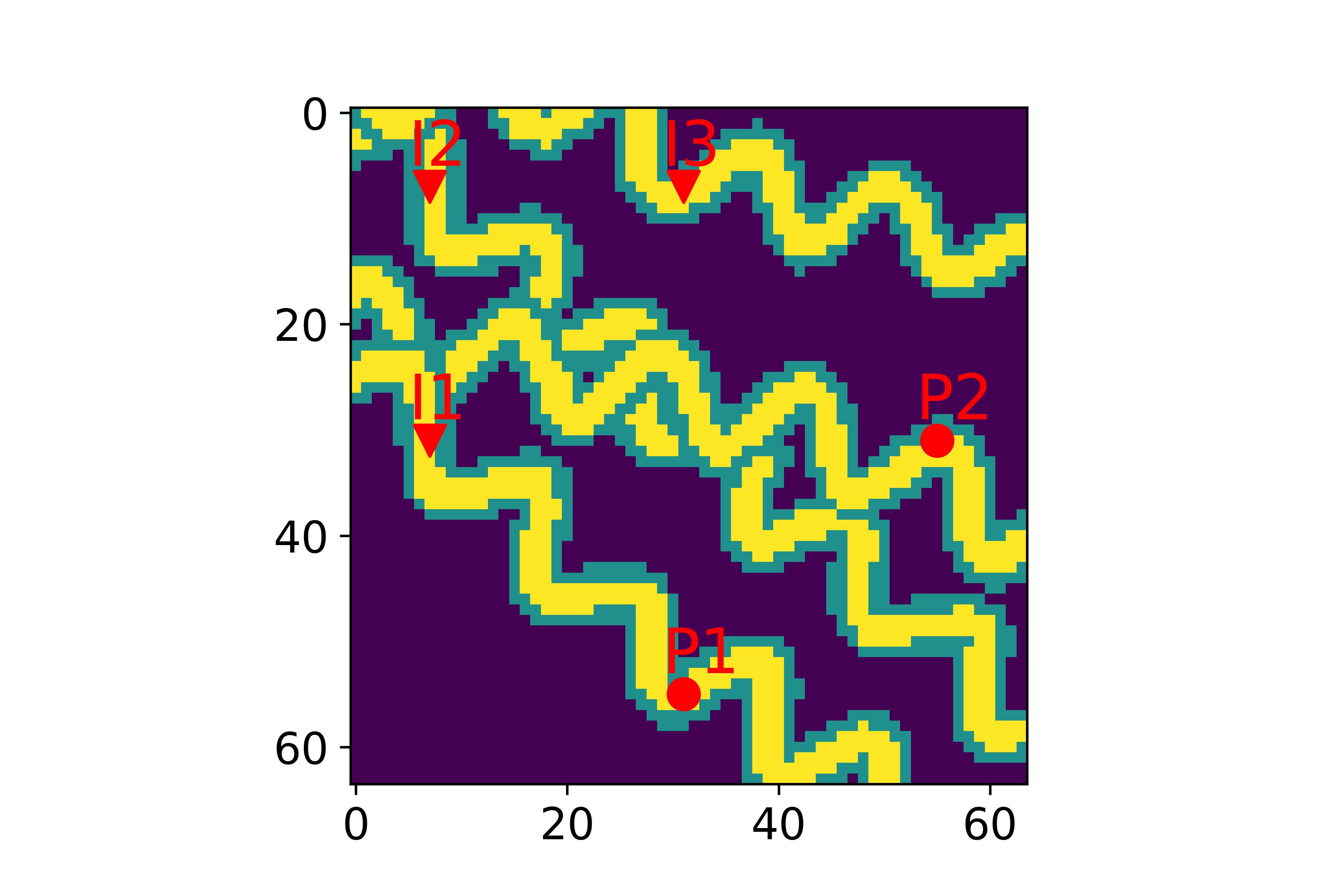}
        \subcaption{True model 2}
        \label{fig:true_var}
    \end{subfigure}
    \caption{Synthetic (Petrel-generated) true models used for history matching.}
    \label{fig:trues}
\end{figure}

Flow simulations are performed in the full geomodel space, while updates occur in the latent space. The LDM generates the facies model corresponding to the updated latent variables at each data assimilation step. Porosity and permeability values for each facies are then assigned -- these are specified (fixed) in the first case and determined during history matching in the second case. Different randomly selected ``true'' (Petrel) models, shown in Figure~\ref{fig:trues}, are considered for the two cases. Observed data are generated by running the flow simulator with the true model, and then adding noise (described below) to the simulated data.

Oil and water production rates at the two producers, and water injection rates at the three injectors, collected every 100~days for the first 1000~days, are used as observed data. This results in a total of $N_d = 70$ observations. In this way, the full simulation time frame of 2500~days is divided into a historical period (up to day~1000), and a forecast period (from day~1000 to day~2500). Well locations and the rest of the simulation setup are the same as in the flow statistics evaluation in Section~\ref{sec:results_models}. 

The history matching algorithm used in this work is the ensemble smoother with multiple data assimilation, ESMDA, developed by \citet{EMERICK20133}. 
ESMDA is widely applied for history matching with high-dimensional models, i.e., in the absence of any geological parameterization. In such cases, heuristic treatments such as localization are commonly used. ESMDA has also been applied for history matching with parameterized models. Examples include \citet{Canchumuni_2019, Canchumuni_2020}, \citet{LIU2021104676}, and \citet{sujiang_2021}. ESMDA is well-suited for use with the LDM parameterization because the latent variable is (close to) standard normal, consistent with the Gaussian assumptions inherent in ESMDA.

ESMDA updates an ensemble of prior uncertain variables by assimilating simulated data to observed data with inflated measurement errors. The prior ensemble consists of $N_e$ realizations of the latent variable, of dimension ${n_c}$, sampled from the standard normal distribution. Since, as mentioned earlier, LDMs rely on a lower-dimensional 2D latent space, the associated history matching variables are in practice the flattened (1D) latent input used for LDM inference. We continue to use $\boldsymbol{\xi}_T$ to refer to these variables. 
At each of the $N_a$ data assimilation steps, the history matching variables are updated (superscript $u$) through application of
\begin{equation}
    \boldsymbol{\xi}^{u,j}_{hm} = \boldsymbol{\xi}^{j}_{hm} + C_{\xi d} (C_{dd} + \alpha C_d)^{-1}(\mathbf{d}^{j} - \mathbf{d}^*_{obs} ), \quad \text{ for } j = 1, \dots, N_e.
\end{equation}
Here $\mathbf{d}^{j} \in \mathbb{R}^{N_d}$ represents the simulated injection and production data and $\mathbf{d}^*_{obs} \in \mathbb{R}^{N_d}$ is the randomly perturbed observed data sampled from $N( \mathbf{d}_{obs}, \alpha C_d$), where $C_d \in \mathbb{R}^{N_d \times N_d}$ is the covariance matrix for the measurement error and $\alpha$ is the inflation factor. Measurement error is assumed to be uncorrelated in space and time (meaning $C_d$ is diagonal), with standard deviation of 2\% of the corresponding simulated value. The matrices $C_{\xi d} \in \mathbb{R}^{n_c \times N_d}$ and $C_{d d} \in \mathbb{R}^{N_d \times N_d} $ are the cross-covariance between $\boldsymbol{\xi}_{hm}$ and $\mathbf{d}$ and the autocovariance of $\mathbf{d}$. They are computed as:
\begin{equation}
    C_{\xi d}=\frac{1}{N_{e}-1} \sum_{j=1}^{N_e} (\boldsymbol{\xi}^j_{hm}-\bar {\boldsymbol{\xi}}_{hm})(\mathbf{d}^j-\bar {\mathbf{d}})^T \quad \text{ and } \quad   C_{d d}=\frac{1}{N_{e}-1} \sum_{j=1}^{N_e} (\mathbf{d}^j-\bar {\mathbf{d}})(\mathbf{d}^j-\bar {\mathbf{d}})^T,
\end{equation}
where the overbar denotes the average over the $N_e$ samples in the ensemble at the current iteration. At each iteration, the observed data are randomly perturbed and the inflation factor is varied. Inflation factors in ESMDA must satisfy $\sum_{i=1}^{N_a} {\alpha_i}^{-1} = 1$. In this work, we use an ensemble size of $N_e = 200$, $N_a =10$ assimilation steps, and inflation coefficients $\alpha_i$, $i=1,\ldots,10$, of [57.017, 35.0, 25.0, 20.0, 18.0, 15.0, 12.0, 8.0, 5.0, 3.0], as suggested by \citet{EMERICK20133}.

\subsection{Case~1: fixed facies properties}
\label{sec:results_hm_fixed}

\begin{figure} 
    \centering
    \begin{subfigure}[b]{0.45\textwidth}
        \centering
        \includegraphics[width=\textwidth]{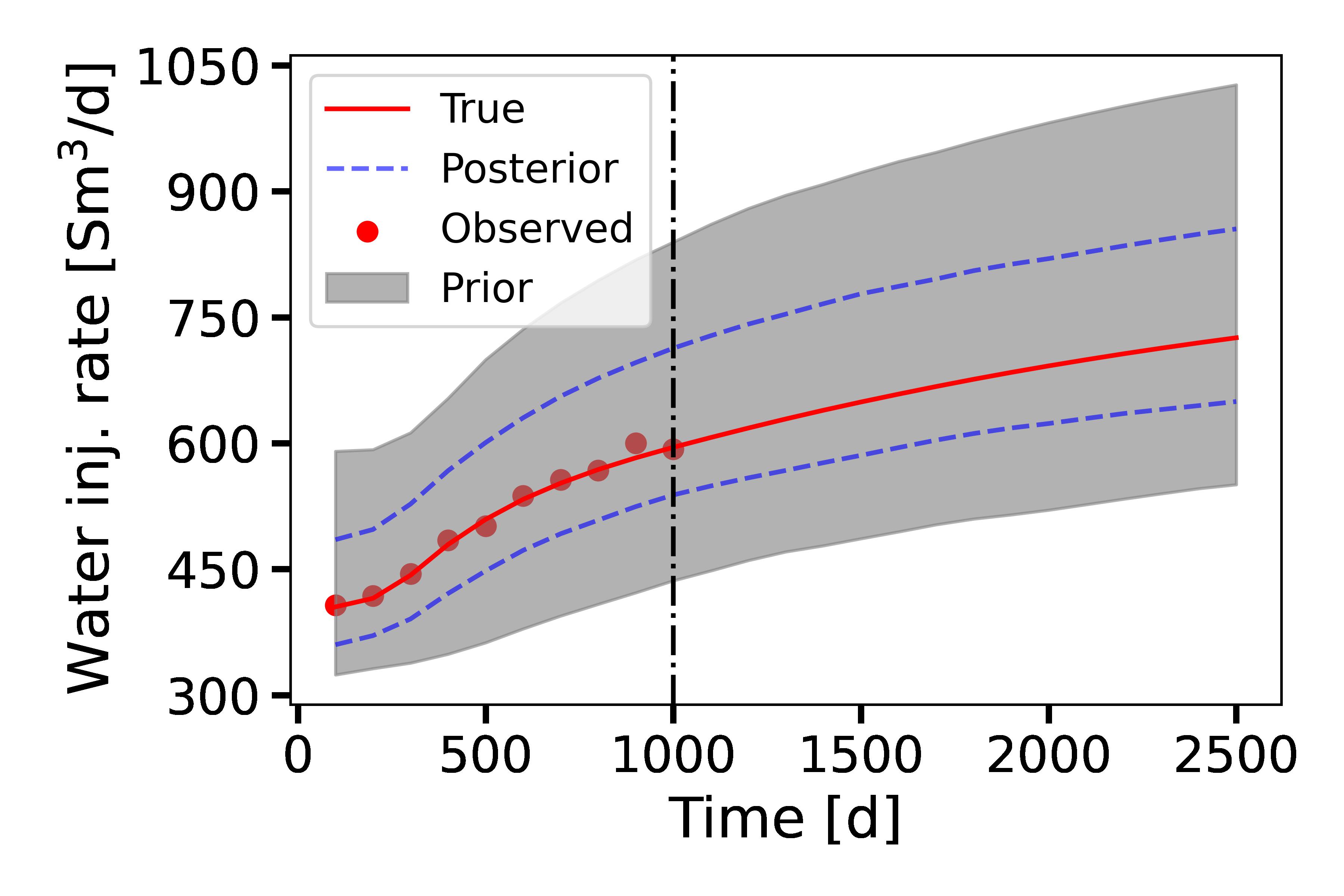}
        \caption{Field water injection rate}
        \label{fig:rates_field_inj}
    \end{subfigure}
    \begin{subfigure}[b]{0.45\textwidth}
        \centering
        \includegraphics[width=\textwidth]{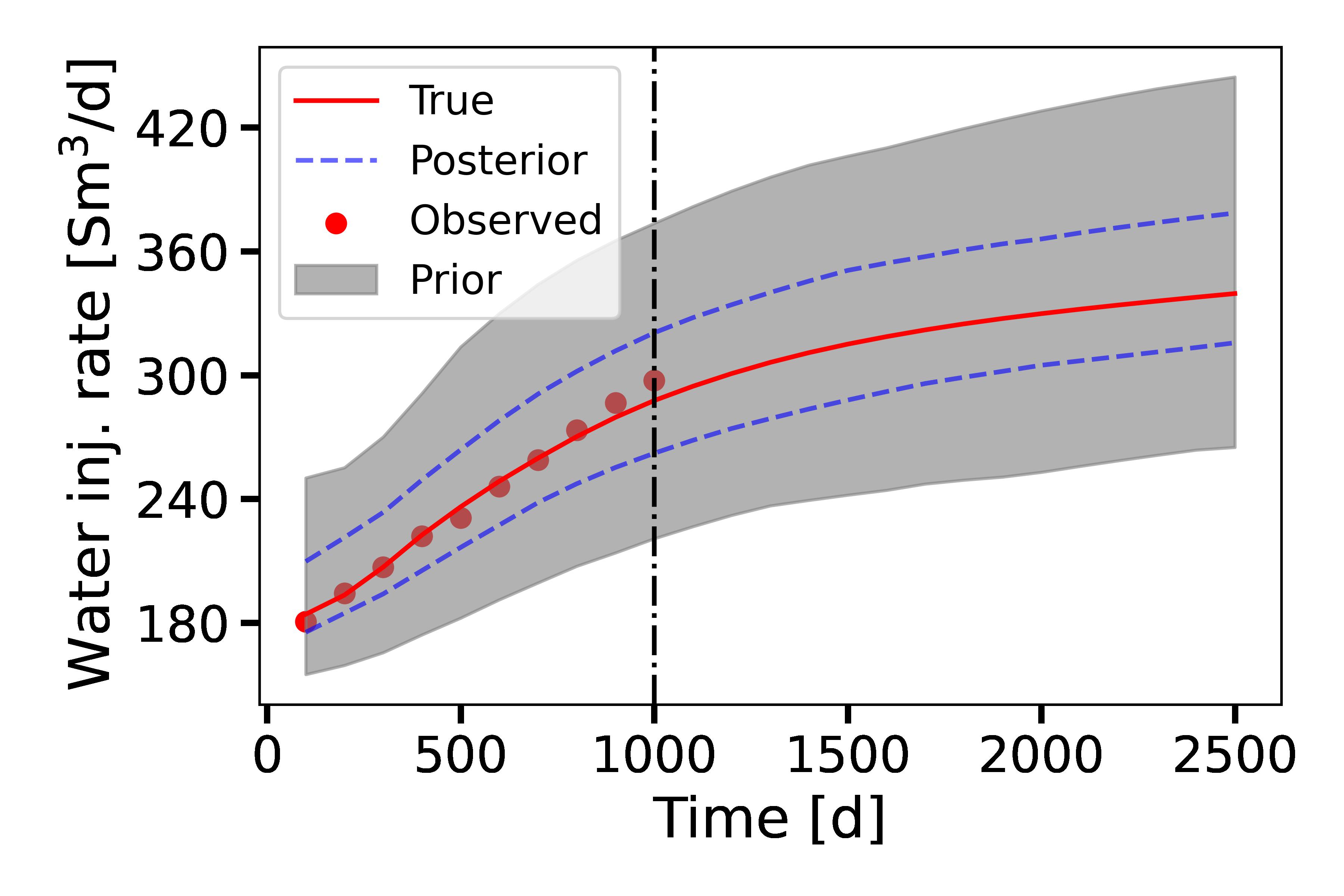}
        \caption{I1 water injection rate (highest injection)}
        \label{fig:rates_I1_WAT}
    \end{subfigure}
    \begin{subfigure}[b]{0.45\textwidth}
        \centering
        \includegraphics[width=\textwidth]{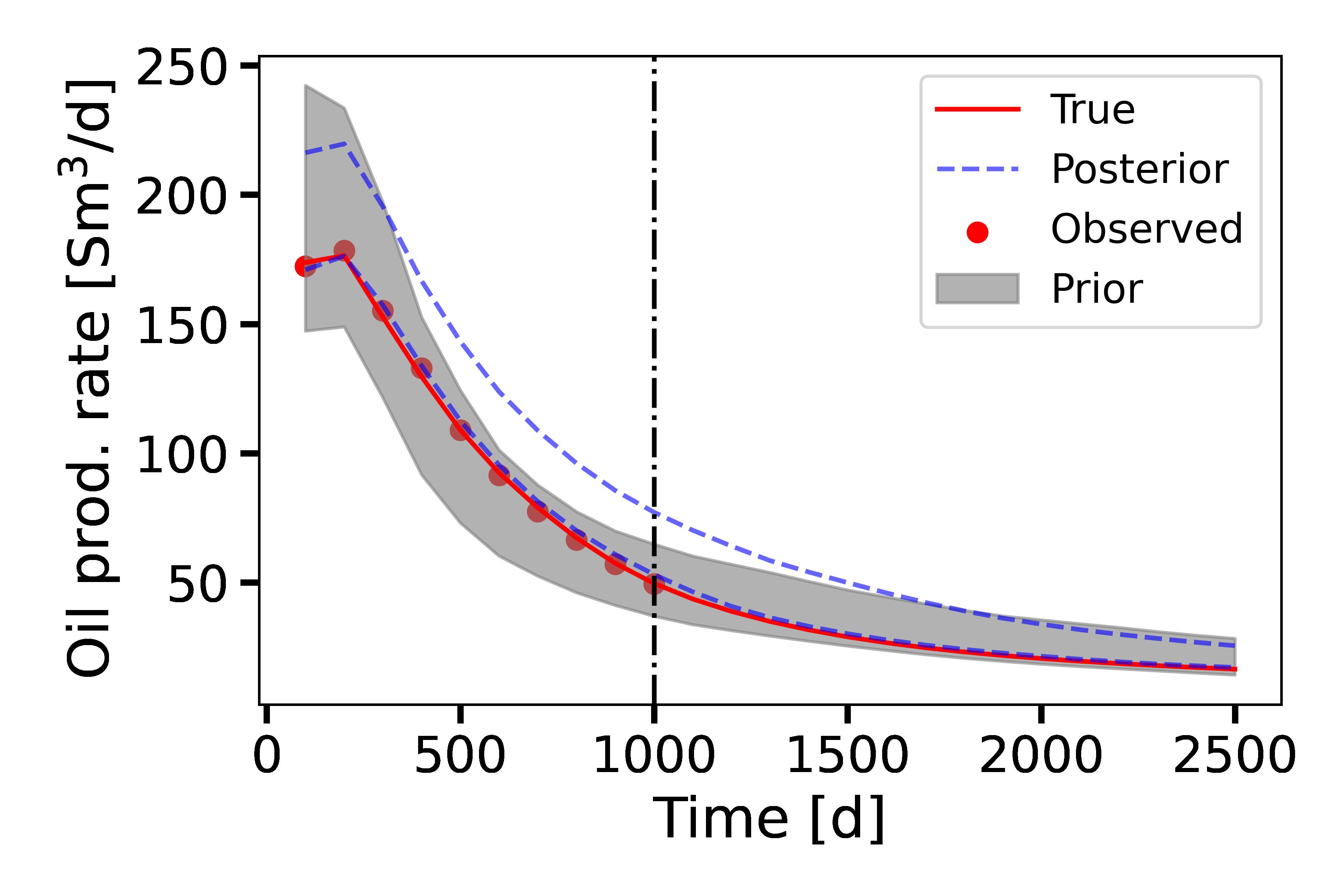}
        \caption{P1 oil production rate}
        \label{fig:rates_P1_OIL}
    \end{subfigure}
    \begin{subfigure}[b]{0.45\textwidth}
        \centering
        \includegraphics[width=\textwidth]{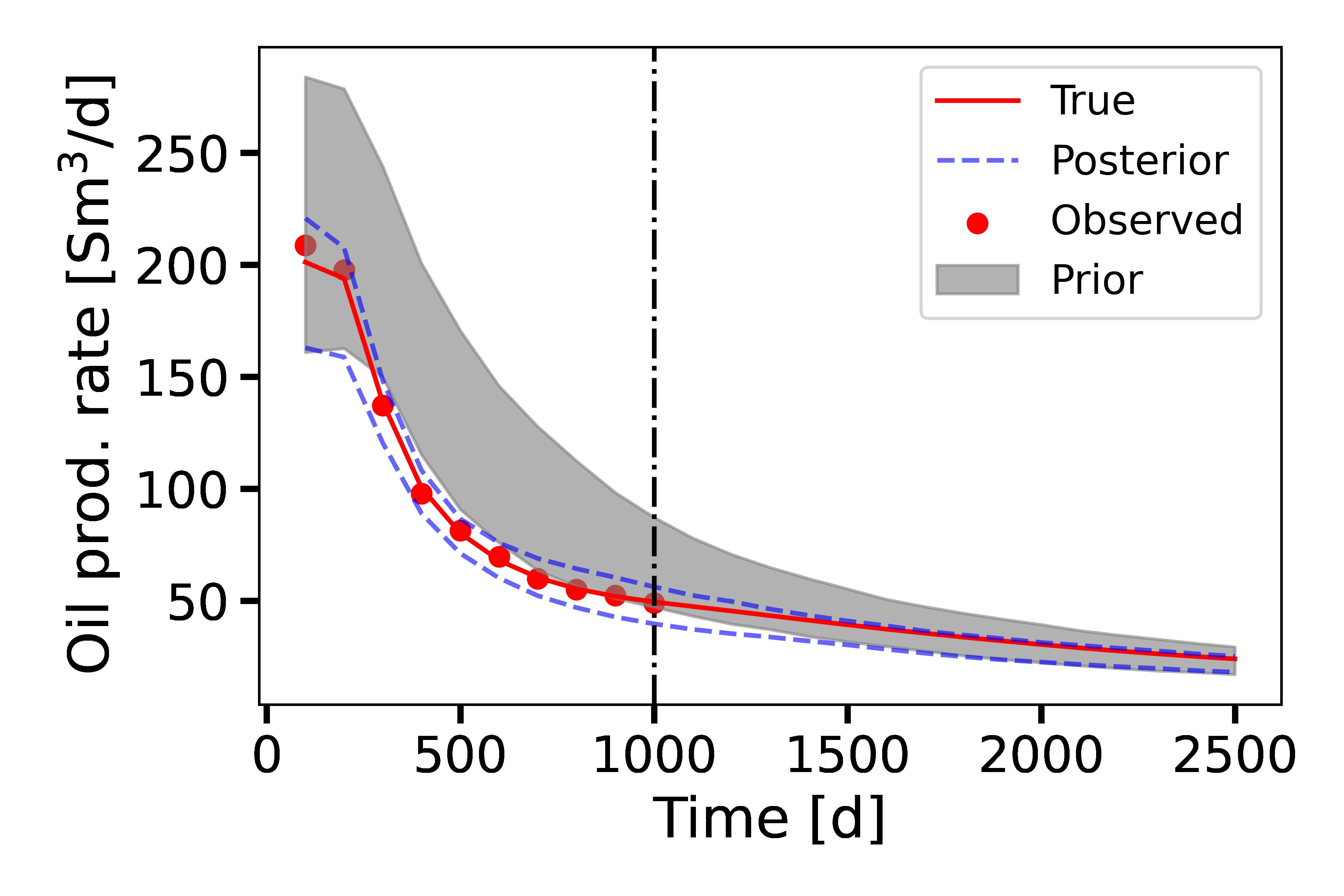}
        \caption{P2 oil production rate}
        \label{fig:rates_P2_OIL}
    \end{subfigure}
    \begin{subfigure}[b]{0.45\textwidth}
        \centering
        \includegraphics[width=\textwidth]{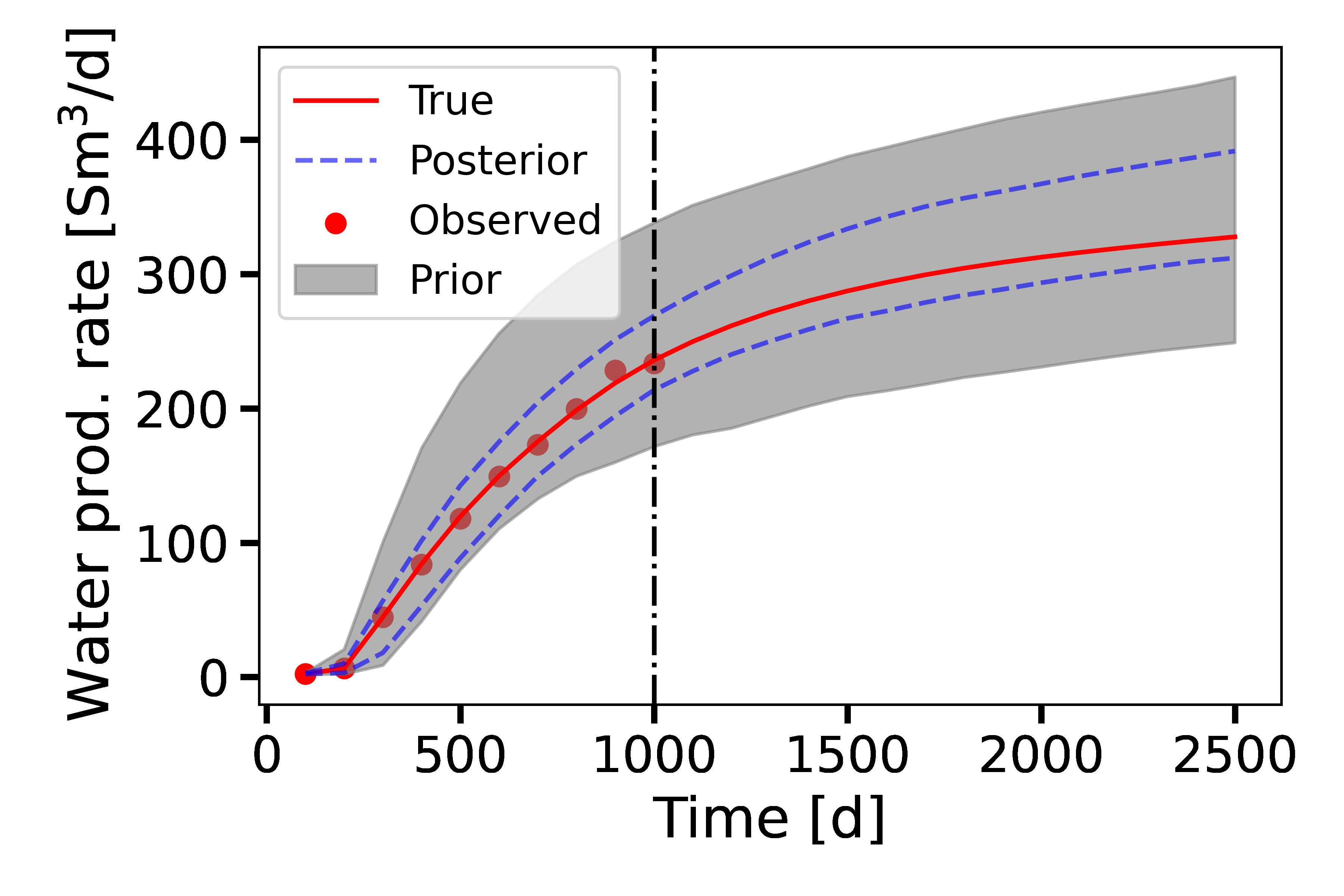}
        \caption{P1 water production rate}
        \label{fig:rates_P1_WAT}
    \end{subfigure}
    \begin{subfigure}[b]{0.45\textwidth}
        \centering
        \includegraphics[width=\textwidth]{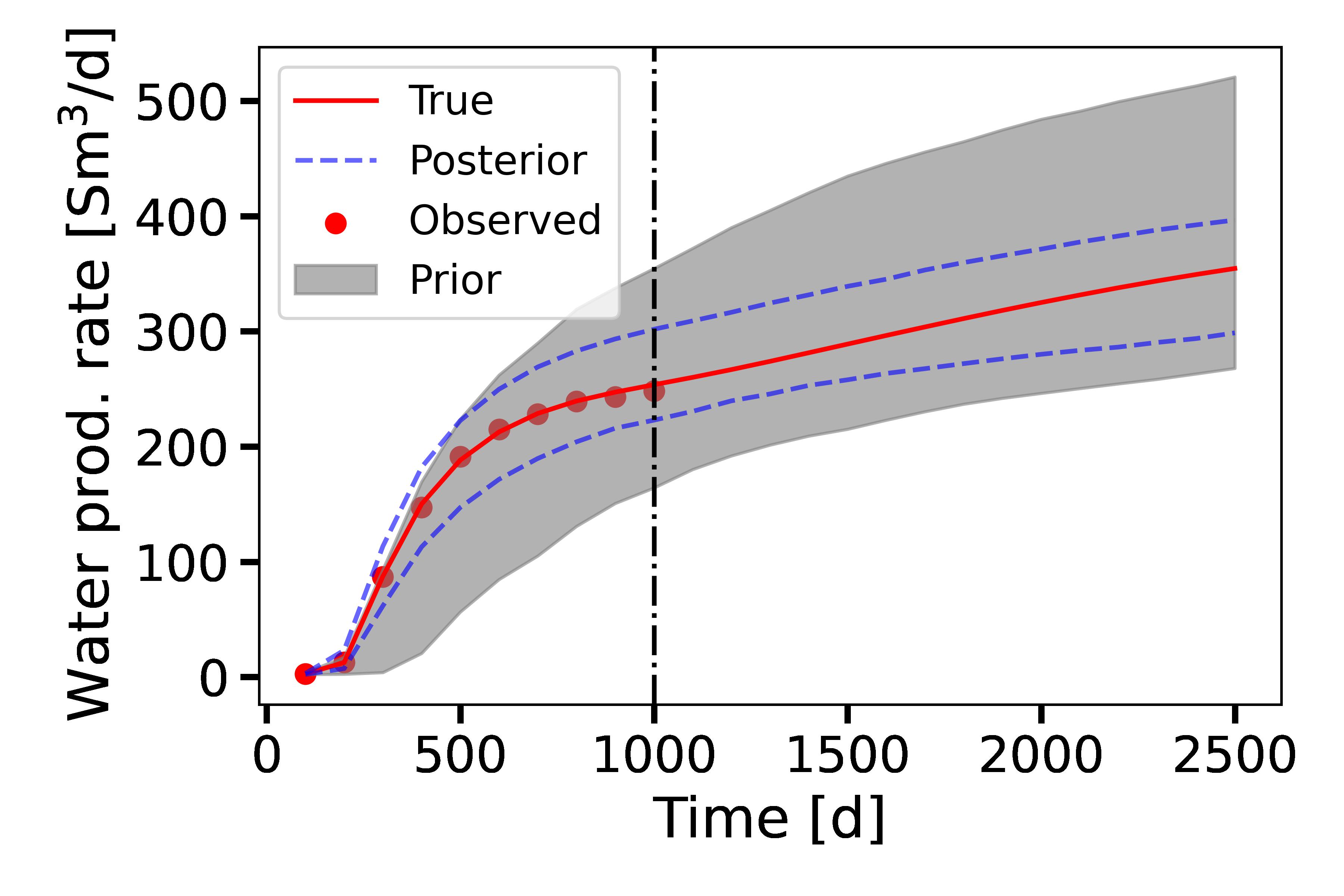}
        \caption{P2 water production rate}
        \label{fig:rates_P2_WAT}
    \end{subfigure}
    \caption{History matching results (Case~1) for key flow quantities. Gray regions show the prior P$_{10}$--P$_{90}$ range, blue dashed curves denote the posterior P$_{10}$--P$_{90}$ range, and red points and red curves represent observed and true data. The vertical dot-dash line indicates the end of the history matching period.
    \label{fig:rates_hm}
}
\end{figure}

History matching results for oil and water rates for this case are shown in Figure~\ref{fig:rates_hm}. The vertical black line in each plot indicates the end of the historical period and the start of the forecasting period. The gray shaded regions denote the P$_{10}$--P$_{90}$ range for the prior ensemble, while the blue dashed curves show the P$_{10}$--P$_{90}$ range for the posterior ensemble. These regions/curves can correspond to different realizations at each time step. The red points represent the perturbed data (used as observations), and the red curves are the simulation results for the true model (Figure~\ref{fig:trues}a). It is evident that a degree of uncertainty reduction is achieved for all flow quantities shown in Figure~\ref{fig:rates_hm}, with the true results (red curves) generally falling between the P$_{10}$--P$_{90}$ for the posterior ensemble. Some of the points in Figure~\ref{fig:rates_hm}c are, however, just below the posterior P$_{10}$ curve.

Next, we present representative prior and posterior geomodels. Representative prior models are obtained by performing $k$-means clustering, with $k=5$, on the prior ensemble, and then selecting the medoid model (closest to the centroid) for each cluster. An analogous procedure is applied on the posterior ensemble. 

\begin{figure}
    \centering
    \begin{subfigure}[b]{0.18\textwidth}
        \centering
        \includegraphics[width=\textwidth, trim=2cm 0 2cm 0, clip]{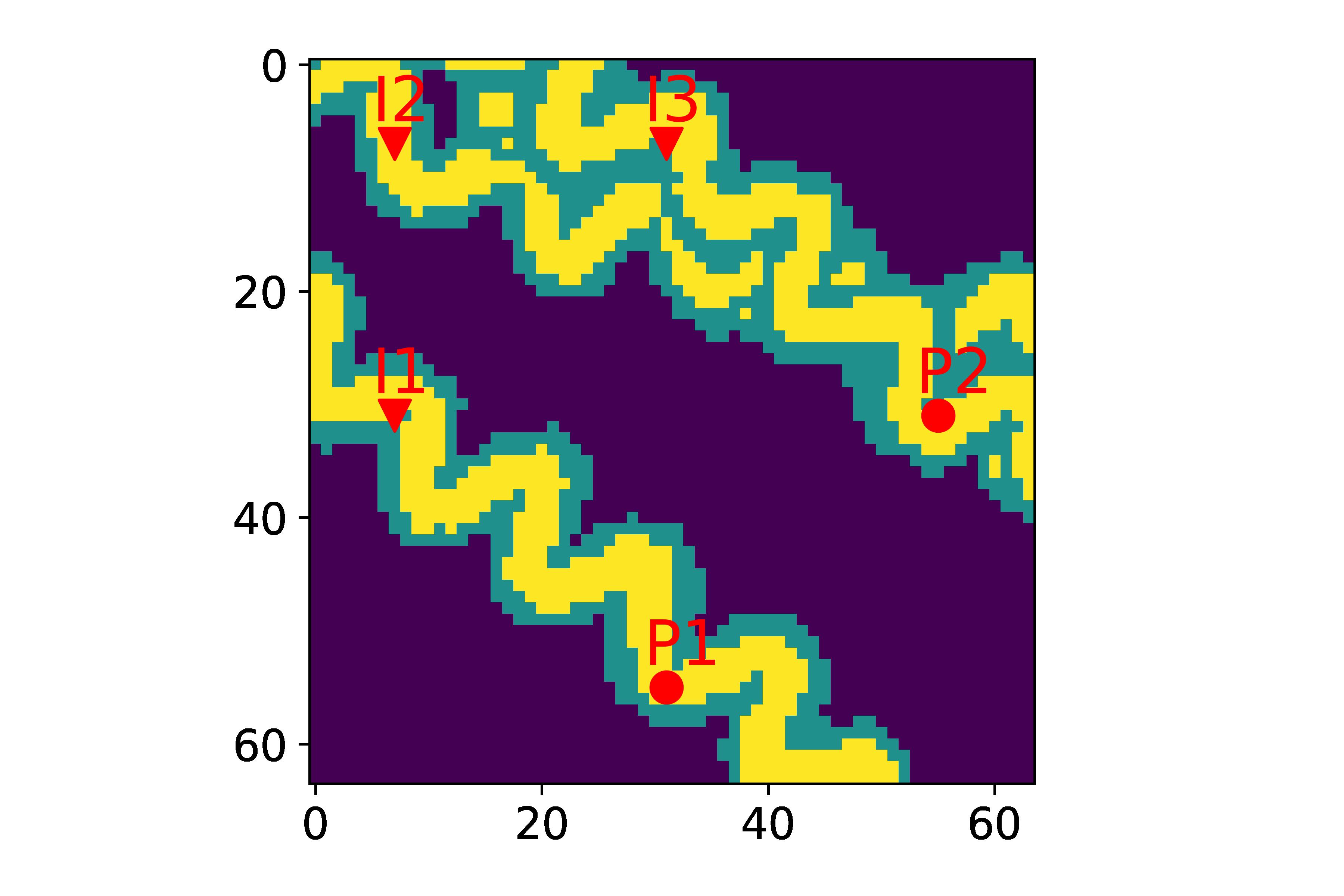}
        \caption{Prior 1}
        \label{fig:medoid_cluster_0_prior}
    \end{subfigure}
    \begin{subfigure}[b]{0.18\textwidth}
        \centering
        \includegraphics[width=\textwidth, trim=2cm 0 2cm 0, clip]{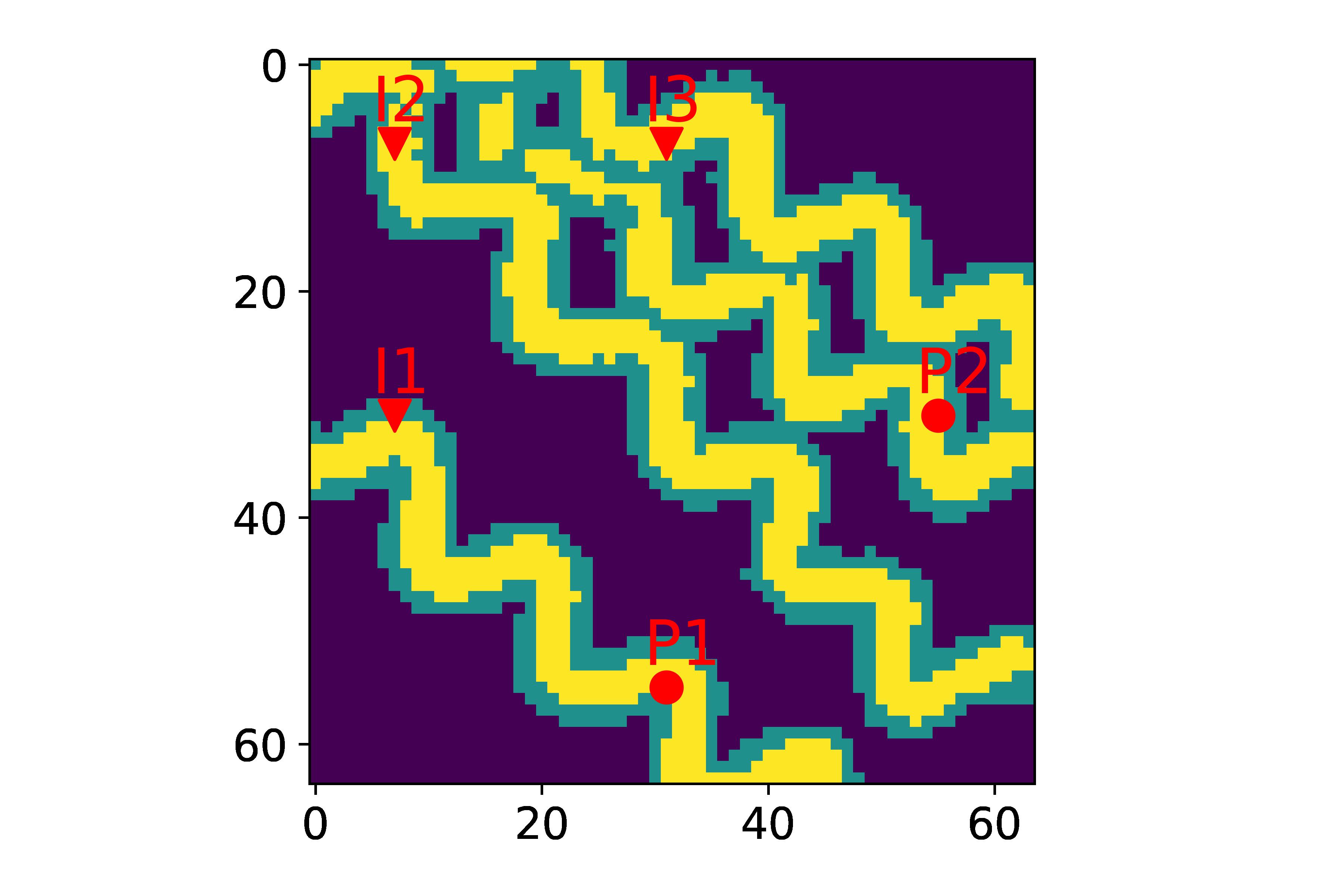}
        \caption{Prior 2}
        \label{fig:medoid_cluster_1_prior}
    \end{subfigure}
    \begin{subfigure}[b]{0.18\textwidth}
        \centering
        \includegraphics[width=\textwidth, trim=2cm 0 2cm 0, clip]{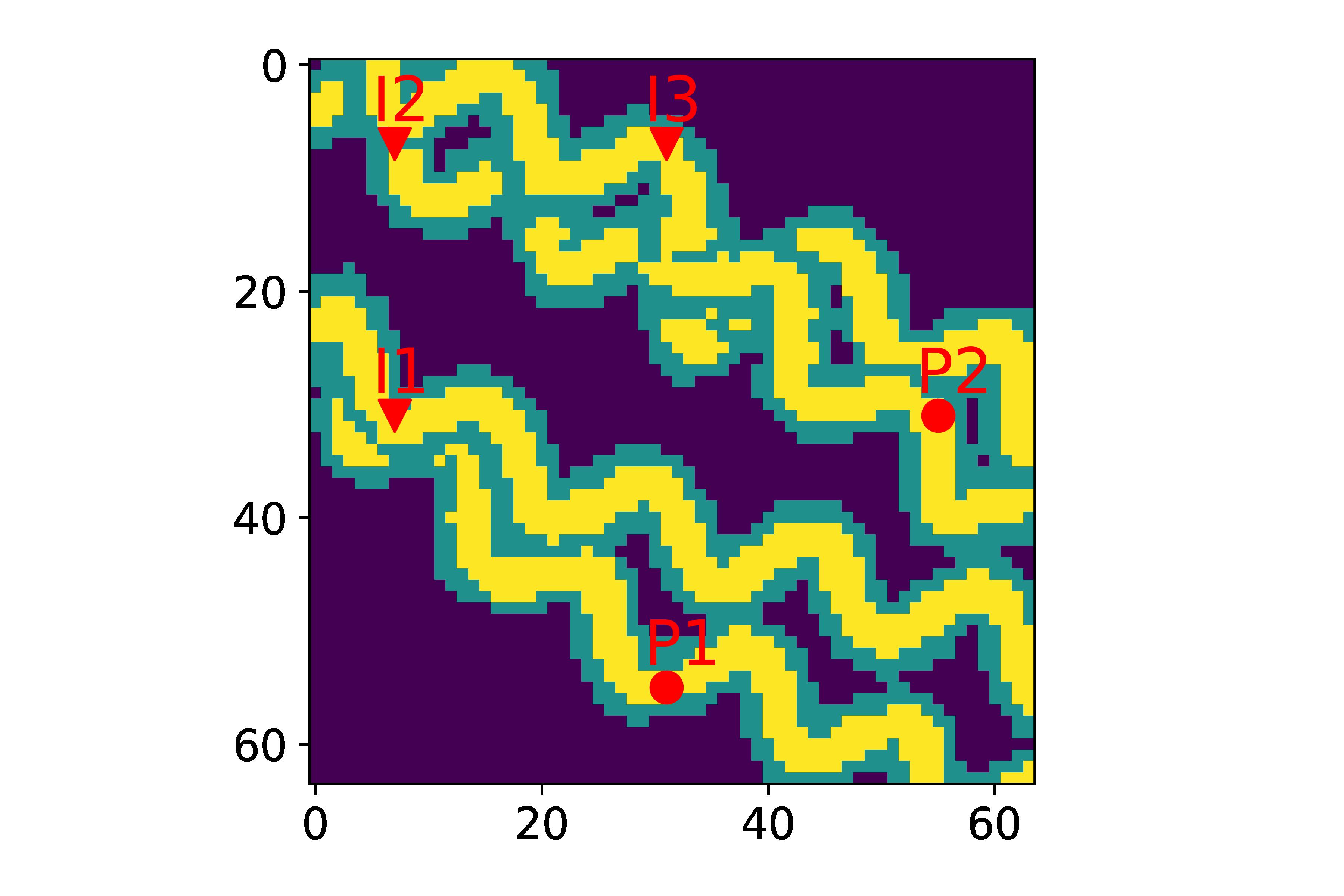}
        \caption{Prior 3}
        \label{fig:medoid_cluster_2_prior}
    \end{subfigure}
    \begin{subfigure}[b]{0.18\textwidth}
        \centering
        \includegraphics[width=\textwidth, trim=2cm 0 2cm 0, clip]{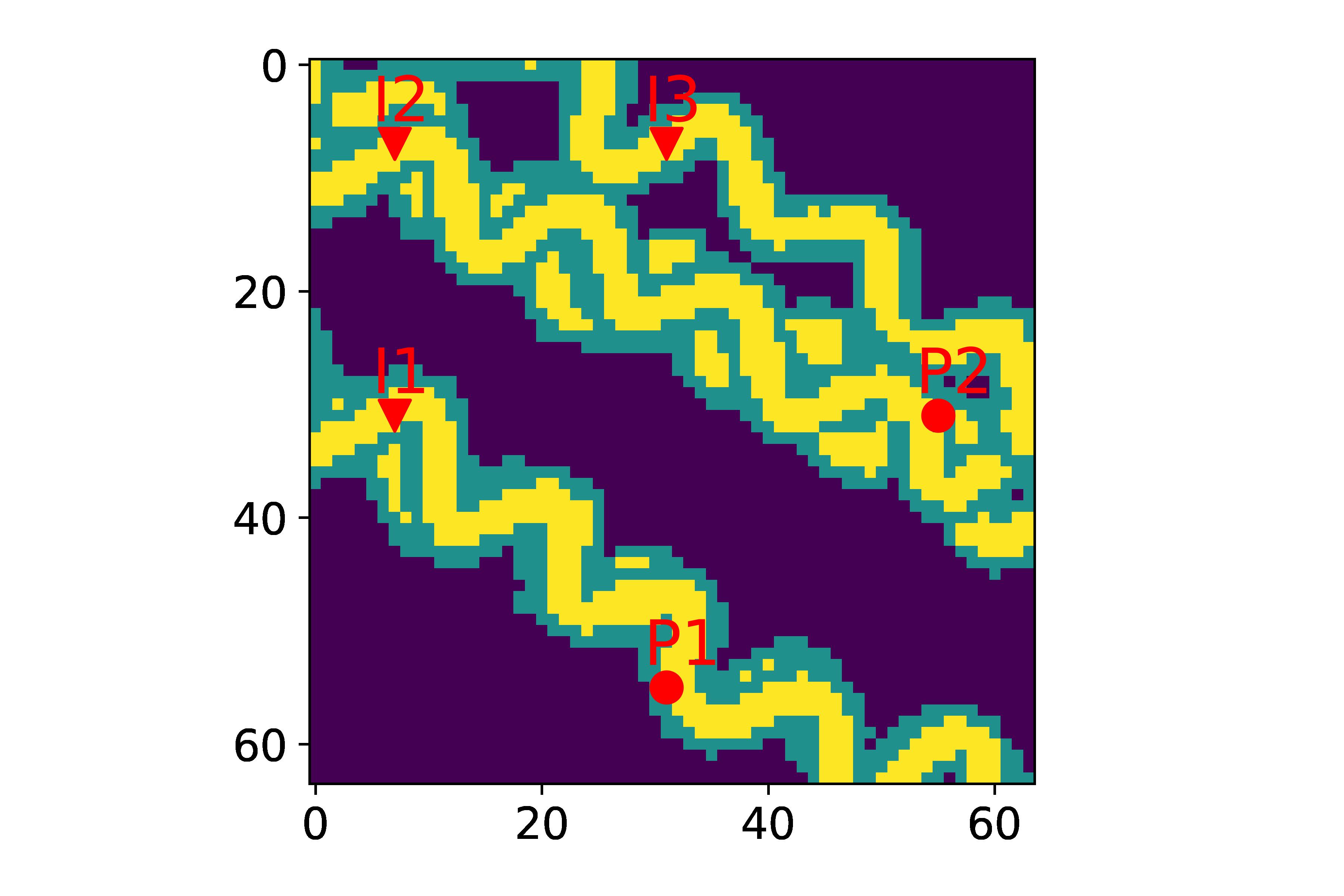}
        \caption{Prior 4}
        \label{fig:medoid_cluster_3_prior}
    \end{subfigure}
    \begin{subfigure}[b]{0.18\textwidth}
        \centering
        \includegraphics[width=\textwidth, trim=2cm 0 2cm 0, clip]{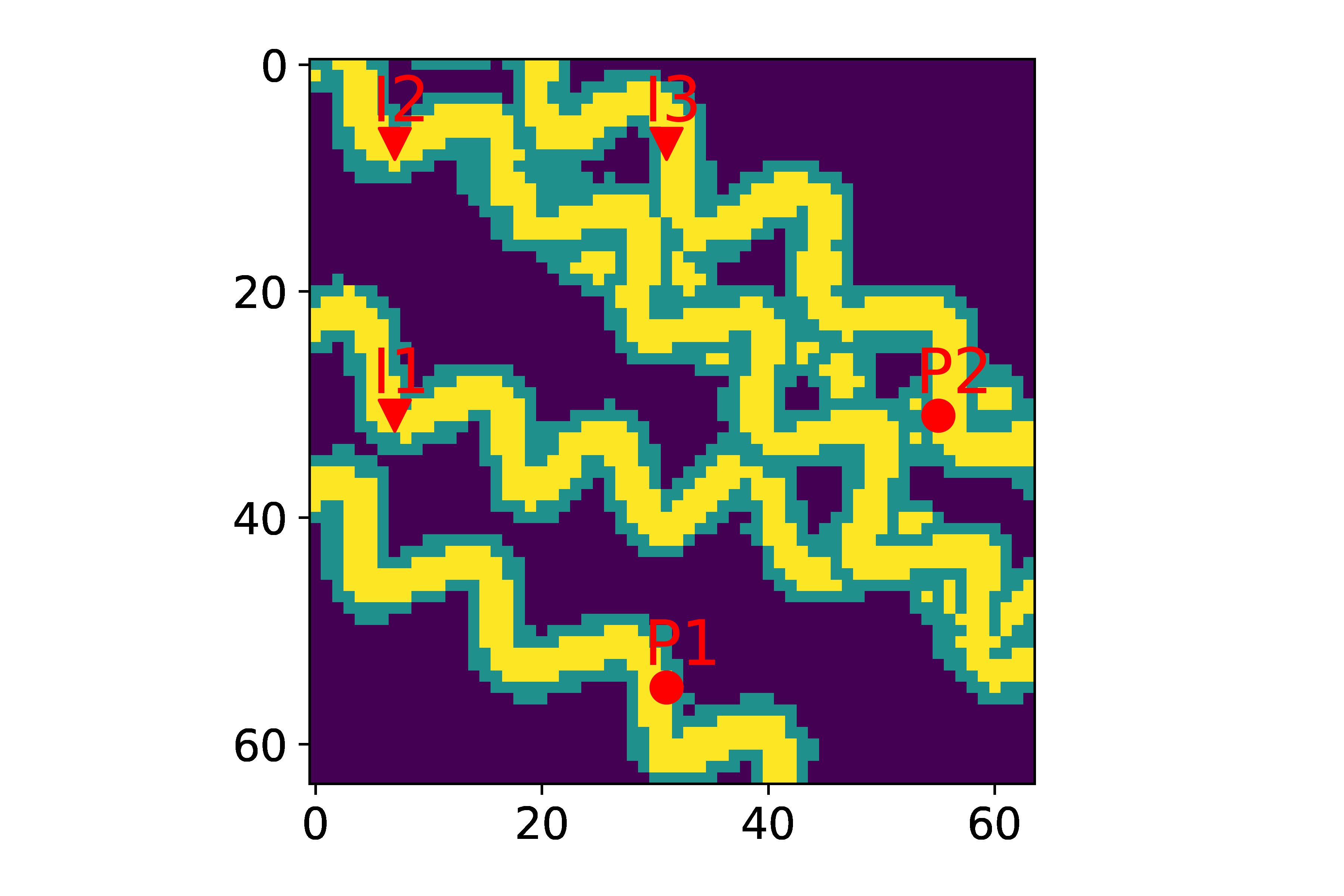}
        \caption{Prior 5}
        \label{fig:medoid_cluster_4_prior}
    \end{subfigure}
    \begin{subfigure}[b]{0.18\textwidth}
        \centering
        \includegraphics[width=\textwidth, trim=2cm 0 2cm 0, clip]{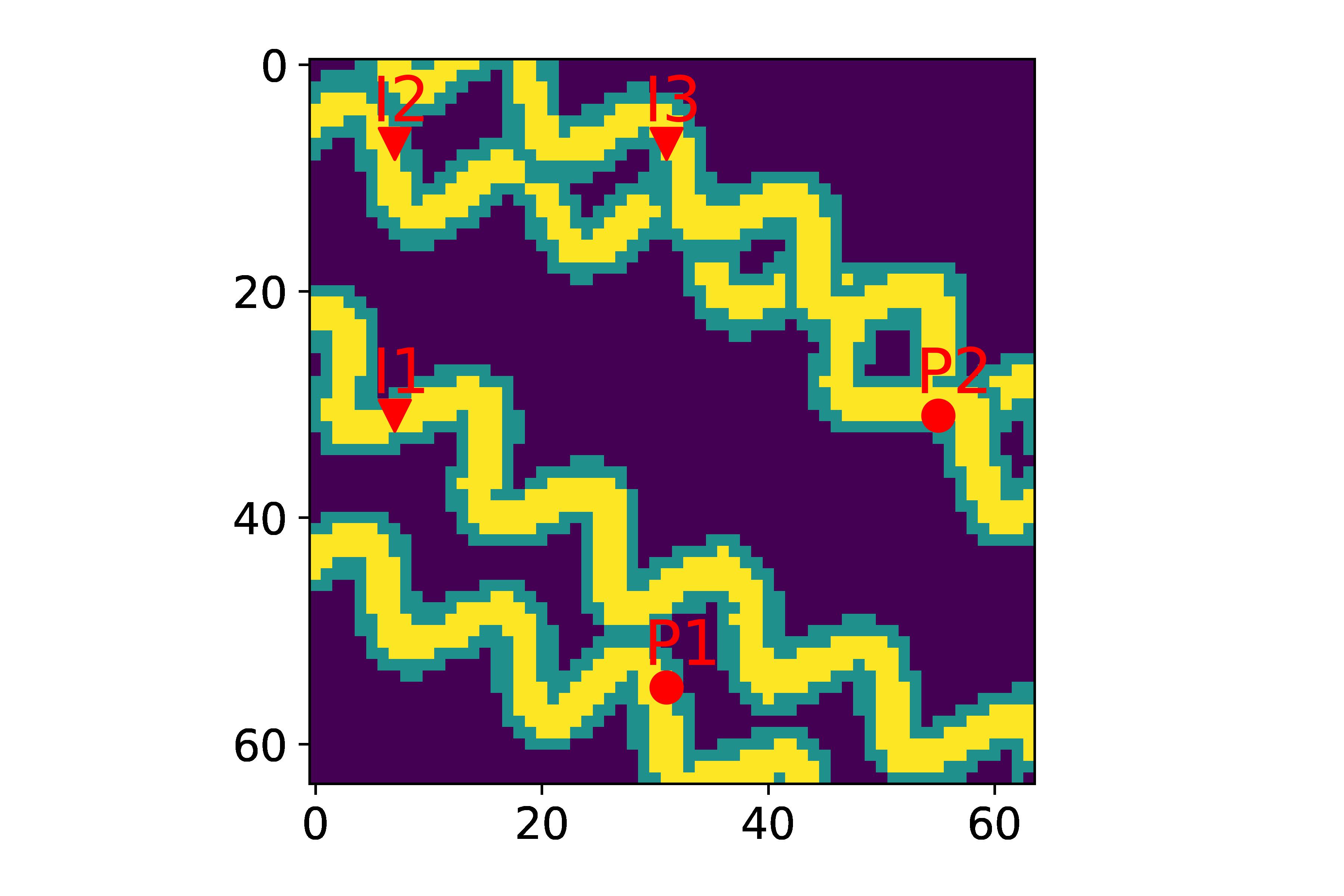}
        \caption{Post.~1}
        \label{fig:medoid_cluster_0_post}
    \end{subfigure}
    \begin{subfigure}[b]{0.18\textwidth}
        \centering
        \includegraphics[width=\textwidth, trim= 2cm 0 2cm 0, clip]{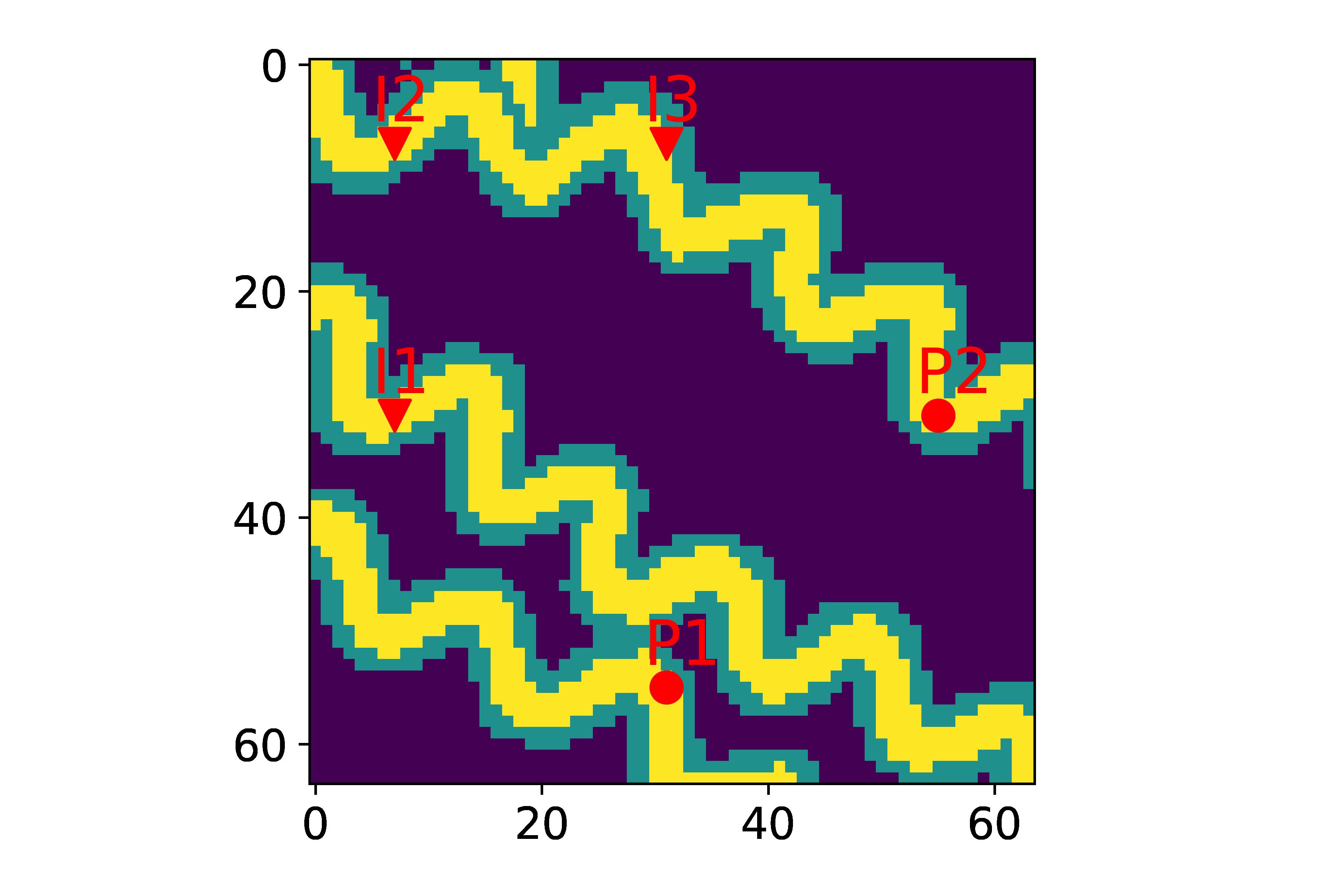}
        \caption{Post.~2}
        \label{fig:medoid_cluster_1_post}
    \end{subfigure}
    \begin{subfigure}[b]{0.18\textwidth}
        \centering
        \includegraphics[width=\textwidth, trim=2cm 0 2cm 0, clip]{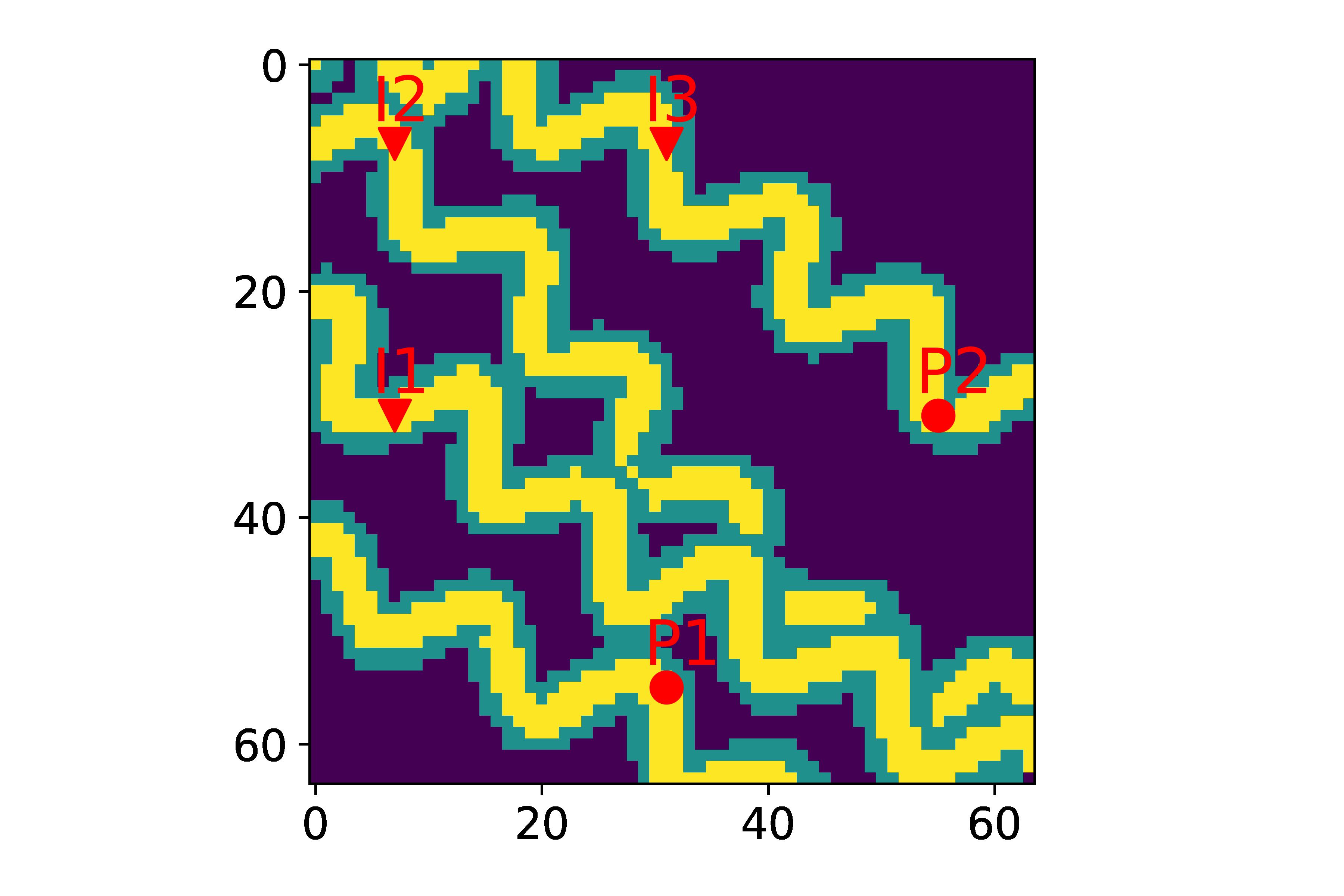}
        \caption{Post.~3}
        \label{fig:medoid_cluster_2_post}
    \end{subfigure}
    \begin{subfigure}[b]{0.18\textwidth}
        \centering
        \includegraphics[width=\textwidth, trim=2cm 0 2cm 0, clip]{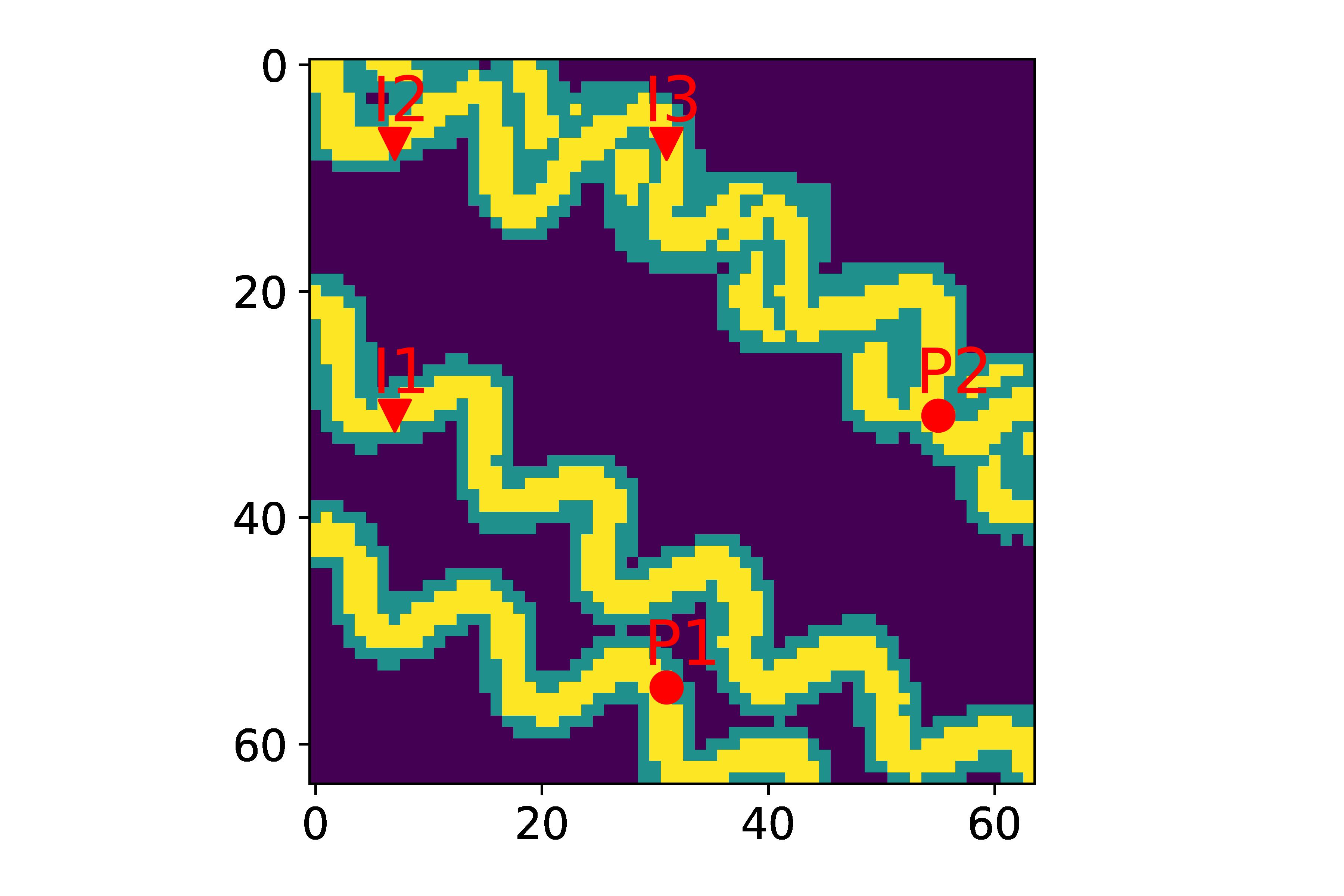}
        \caption{Post.~4}
        \label{fig:medoid_cluster_3_post}
    \end{subfigure}
    \begin{subfigure}[b]{0.18\textwidth}
        \centering
        \includegraphics[width=\textwidth, trim=2cm 0 2cm 0, clip]{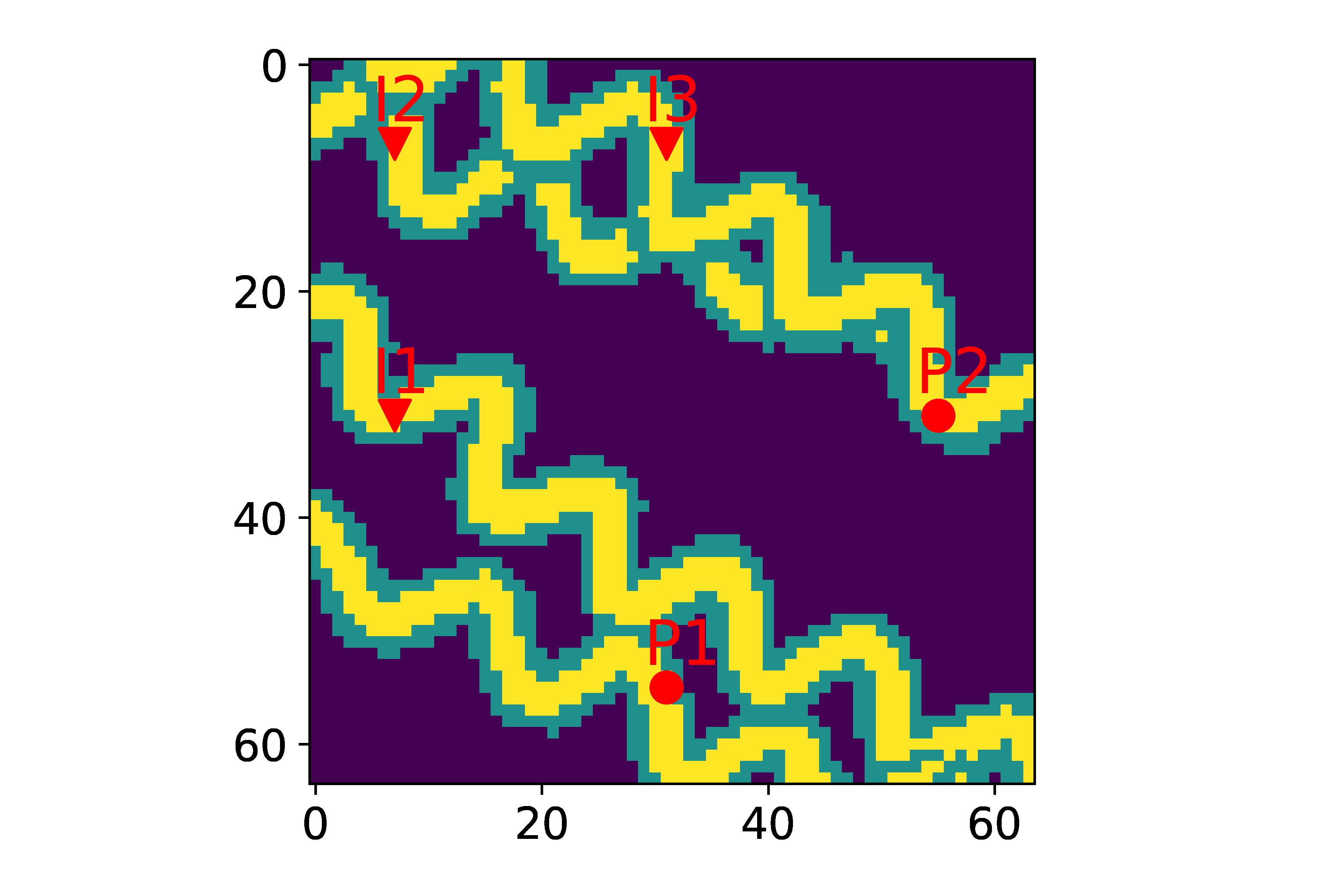}
        \caption{Post.~5}
        \label{fig:medoid_cluster_4_post}
    \end{subfigure}
    \caption{Prior and posterior realizations (Case~1) determined by $k$-means clustering. Each model is the medoid of its cluster. True model shown in Figure~\ref{fig:trues}a.
    \label{fig:medoids}
}
\end{figure}

The resulting representative models are shown in Figure~\ref{fig:medoids}. Note that there is no direct correspondence between prior realizations in the top row and posterior realizations in the bottom row -- each is simply a cluster-medoid model. The prior realizations display clear differences. For example, most prior realizations have a channel connection between wells I1 and P1, but this is not the case in Figure~\ref{fig:medoids}e. Differences are also observed along the I2--P2 direction. The posterior realizations, by contrast, display more consistent geological features and connections. In addition, they share aspects of the true model (Figure~\ref{fig:trues}a), such as the two distinct channels that just touch one another along the I1--P1 direction and, other than in Figure~\ref{fig:medoids}h, intertwined channels toward the top of the model.

\subsection{Case~2: uncertain facies properties}
\label{sec:results_hm_var}

The history matching workflow is now applied for a case with uncertain facies porosity and permeability values. The true model is shown in Figure~\ref{fig:trues}b, and the same simulation setup as in Case~1 is used. Prior realizations are constructed by sampling the components of $\boldsymbol{\xi}_T$ from the standard normal distribution and the facies properties ($\boldsymbol{\phi}$ and ${\bf k}$) from separate normal distributions. The mean and standard deviation for each facies property are given in Table~\ref{table:poro_perm_combined}. The distributions are truncated (at plus and minus two standard deviations) to avoid nonphysical values. Updated values at each data assimilation step are similarly truncated. Facies properties for the true model are determined by sampling from these distributions.

\begin{table}[h]
\centering
\caption{Mean, standard deviation, minimum, and maximum values for porosity and permeability (in $\log_e k$, with $k$ in mD) for the three facies for Case~2.}
\label{table:poro_perm_combined}
\begin{tabular}{lcccc|cccc}
\hline
\multirow{2}{*}{\textbf{Facies}} & \multicolumn{4}{c}{\textbf{Porosity}} & \multicolumn{4}{c}{\textbf{Permeability [$\bm{\log_e k}$~mD]}} \\
\cline{2-9}
 & {Mean} & {Std. dev.} & {Min.} & {Max.} & {Mean} & {Std. dev.} & {Min.} & {Max.} \\
\hline
{Mud} & 0.075 & 0.0125 & 0.05 & 0.10 & 3.45 & 0.23 & 3.00 & 3.91 \\
\hline
{Levee} & 0.16 & 0.02 & 0.12 & 0.20 & 5.41 & 0.40 & 4.61 & 6.21 \\
\hline
{Channel} & 0.26 & 0.02 & 0.22 & 0.30 & 7.46 & 0.27 & 6.91 & 8.00 \\
\hline
\end{tabular}
\end{table}

\begin{figure}
    \centering
    \begin{subfigure}[b]{0.45\textwidth}
        \centering
        \includegraphics[width=\textwidth]{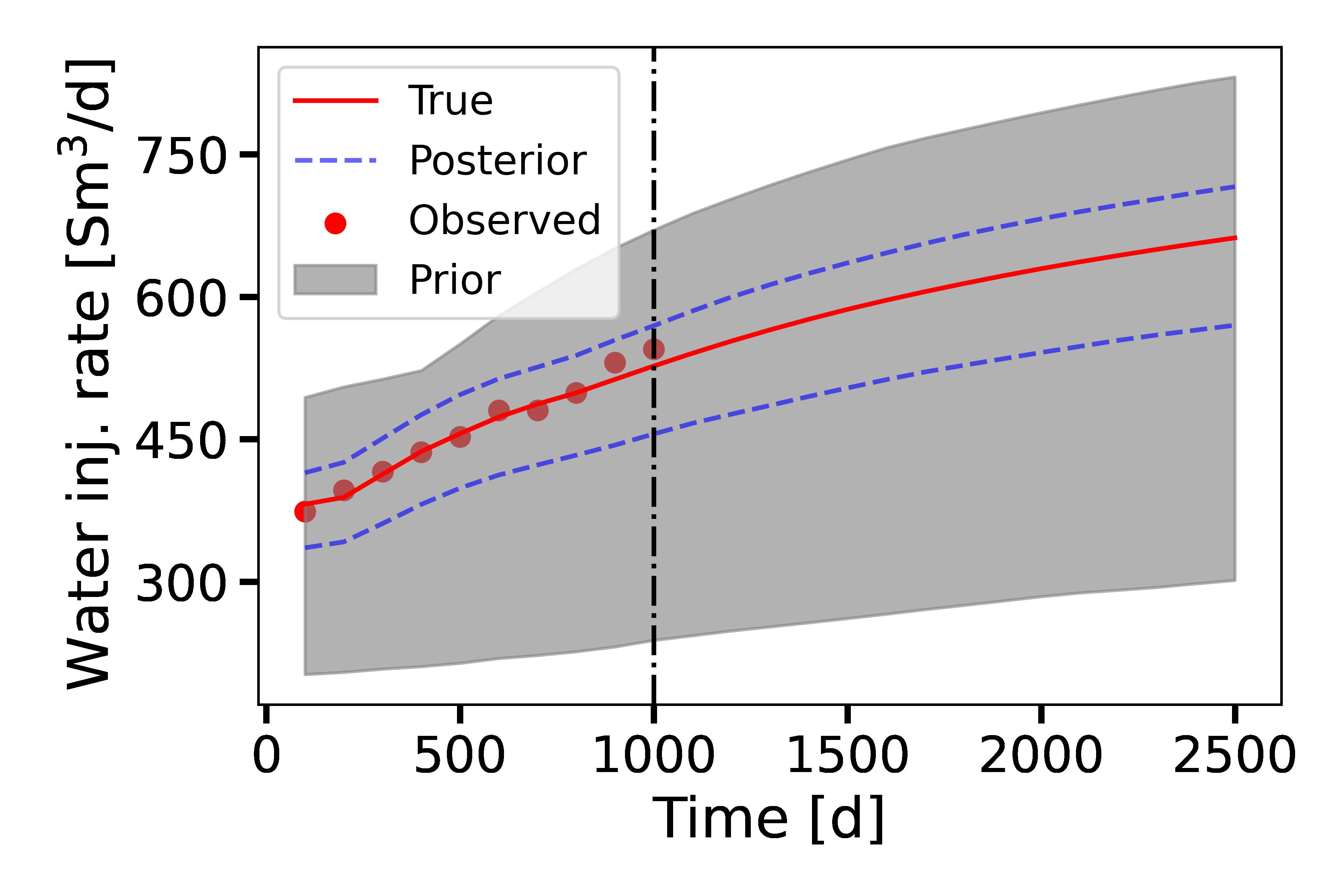}
        \caption{Field water injection rate}
        \label{fig:rates_field_inj_var}
    \end{subfigure}
    \begin{subfigure}[b]{0.45\textwidth}
        \centering
        \includegraphics[width=\textwidth]{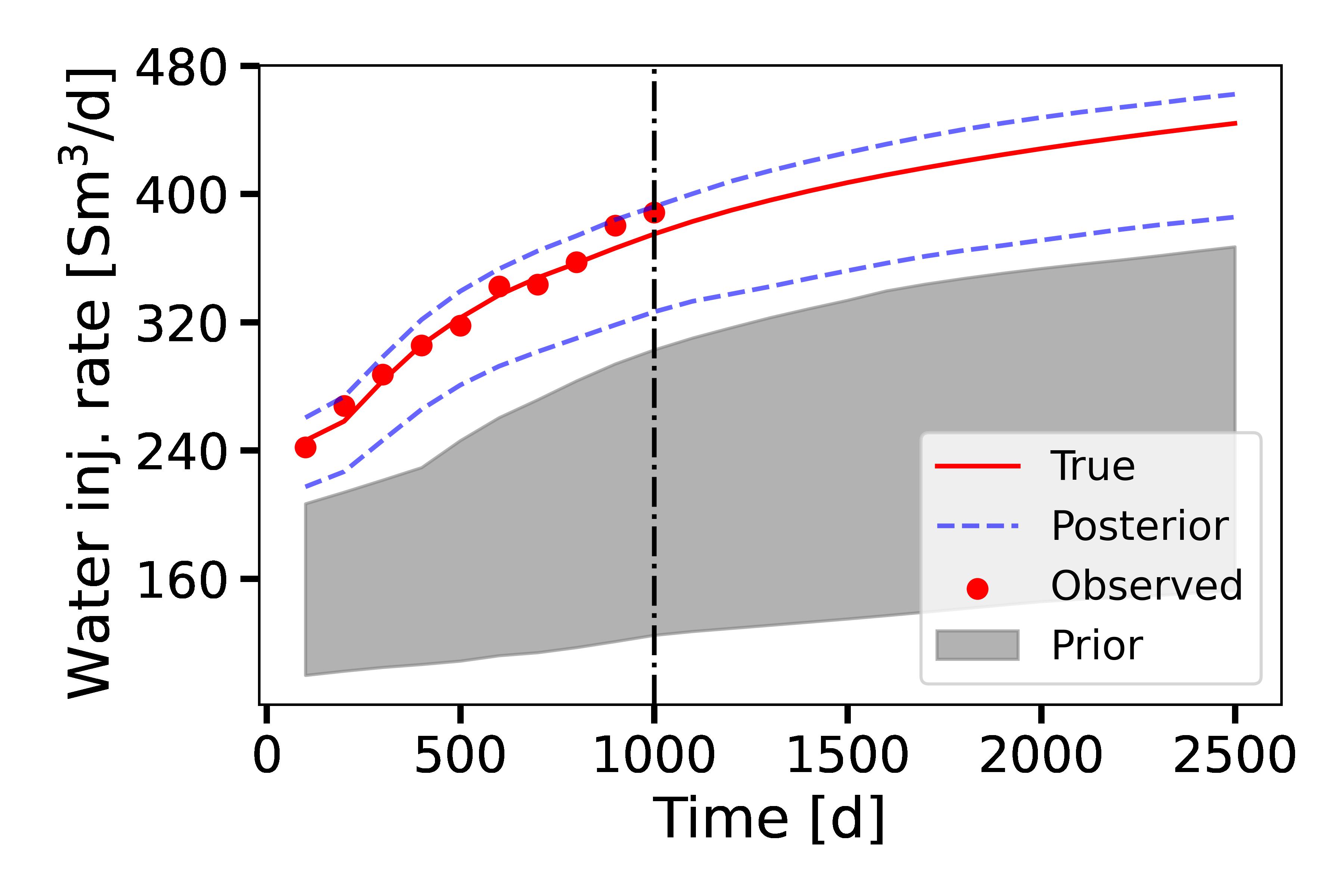}
        \caption{I1 water injection rate (highest injection)}
        \label{fig:rates_I1_WAT_var}
    \end{subfigure}
    \begin{subfigure}[b]{0.45\textwidth}
        \centering
        \includegraphics[width=\textwidth]{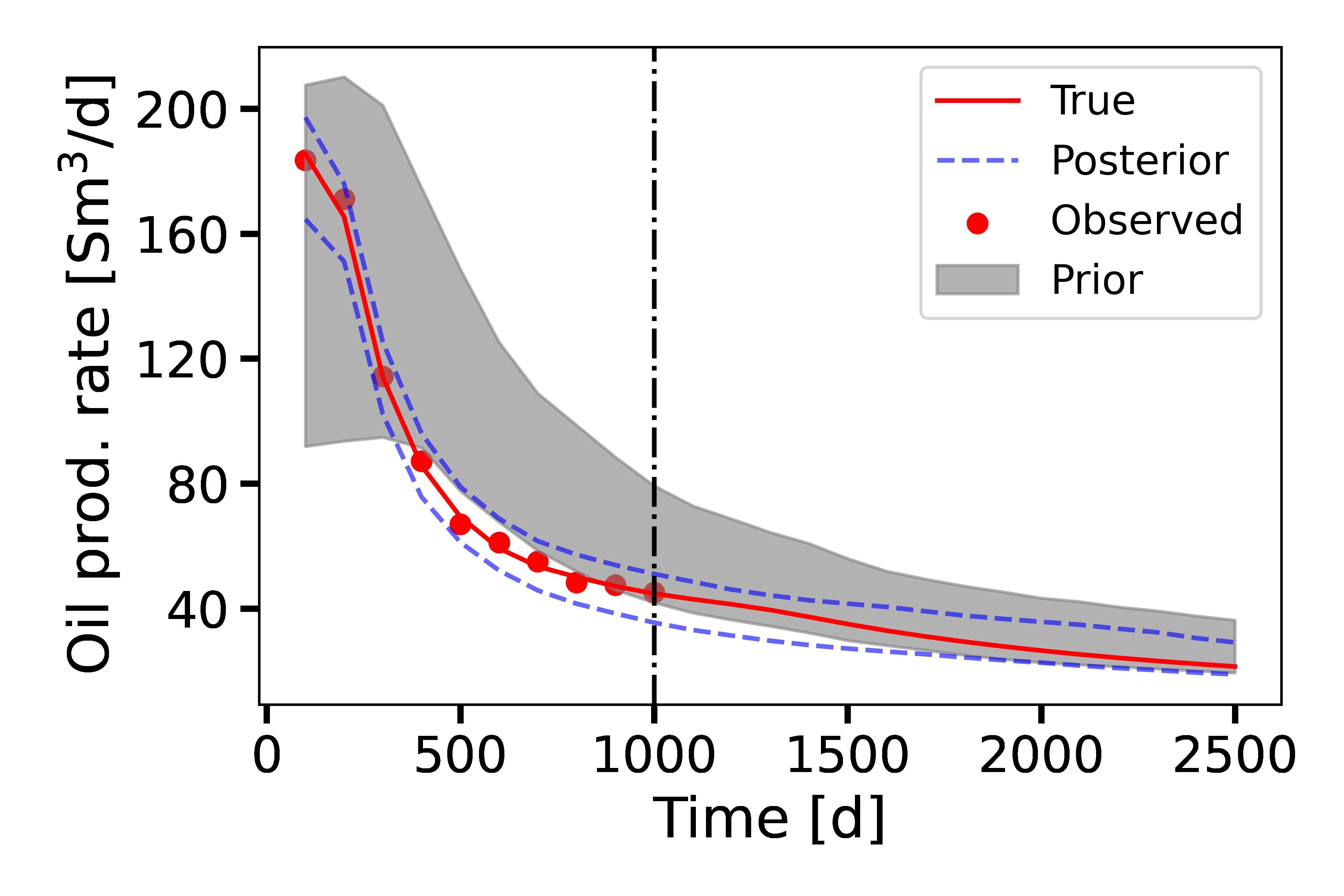}
        \caption{P1 oil production rate}
        \label{fig:rates_P1_OIL_var}
    \end{subfigure}
    \begin{subfigure}[b]{0.45\textwidth}
        \centering
        \includegraphics[width=\textwidth]{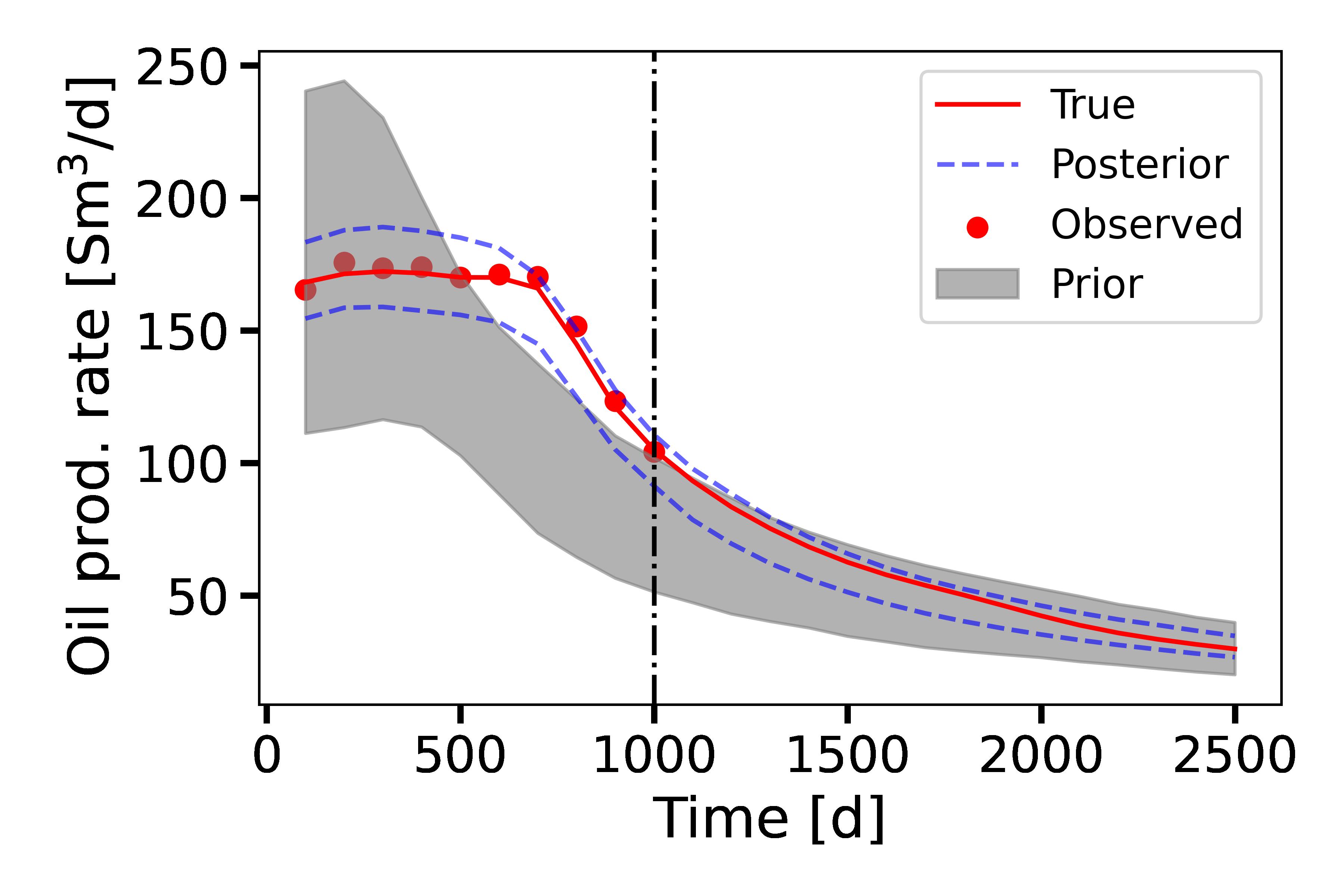}
        \caption{P2 oil production rate}
        \label{fig:rates_P2_OIL_var}
    \end{subfigure}
    \begin{subfigure}[b]{0.45\textwidth}
        \centering
        \includegraphics[width=\textwidth]{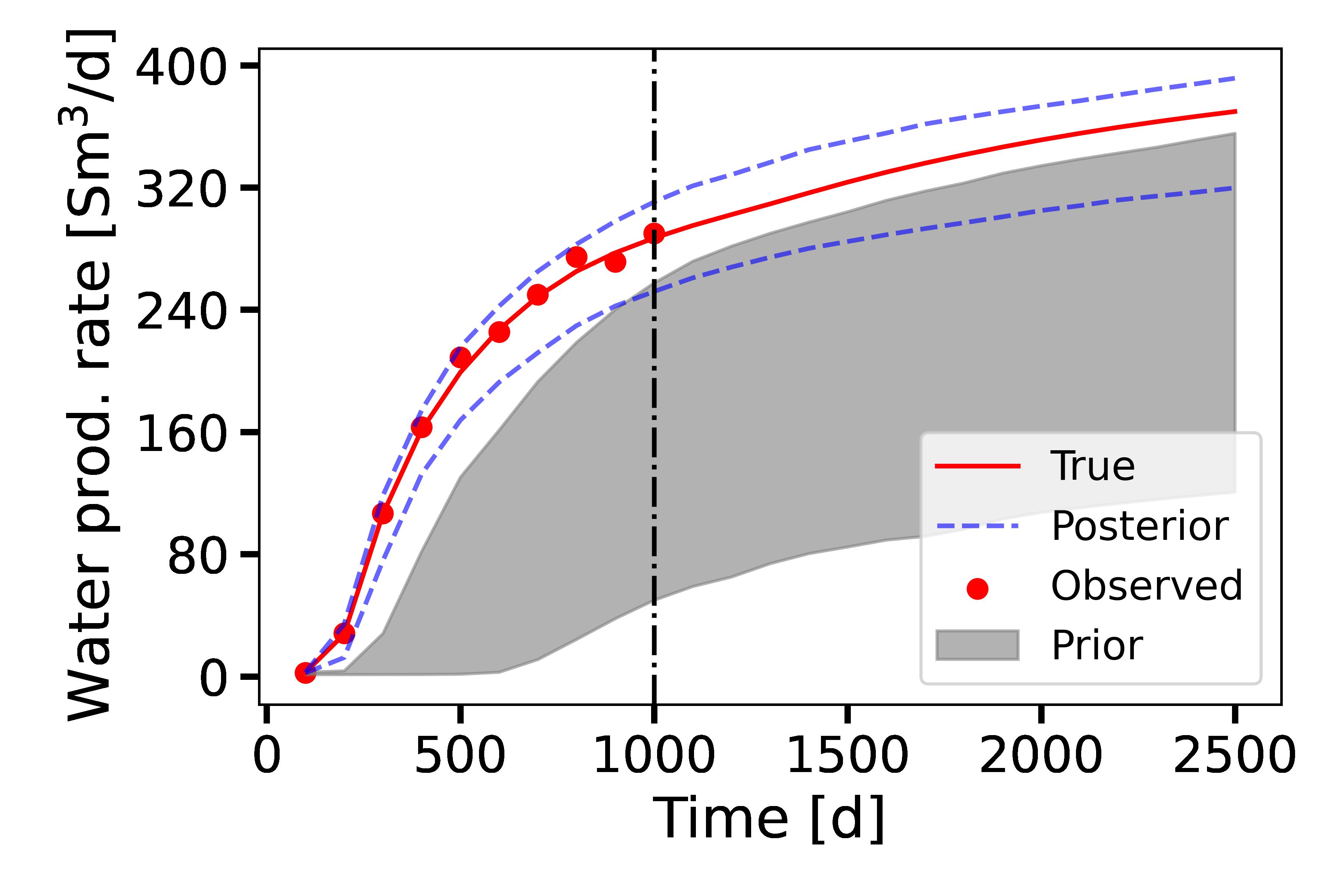}
        \caption{P1 water production rate}
        \label{fig:rates_P1_WAT_var}
    \end{subfigure}
    \begin{subfigure}[b]{0.45\textwidth}
        \centering
        \includegraphics[width=\textwidth]{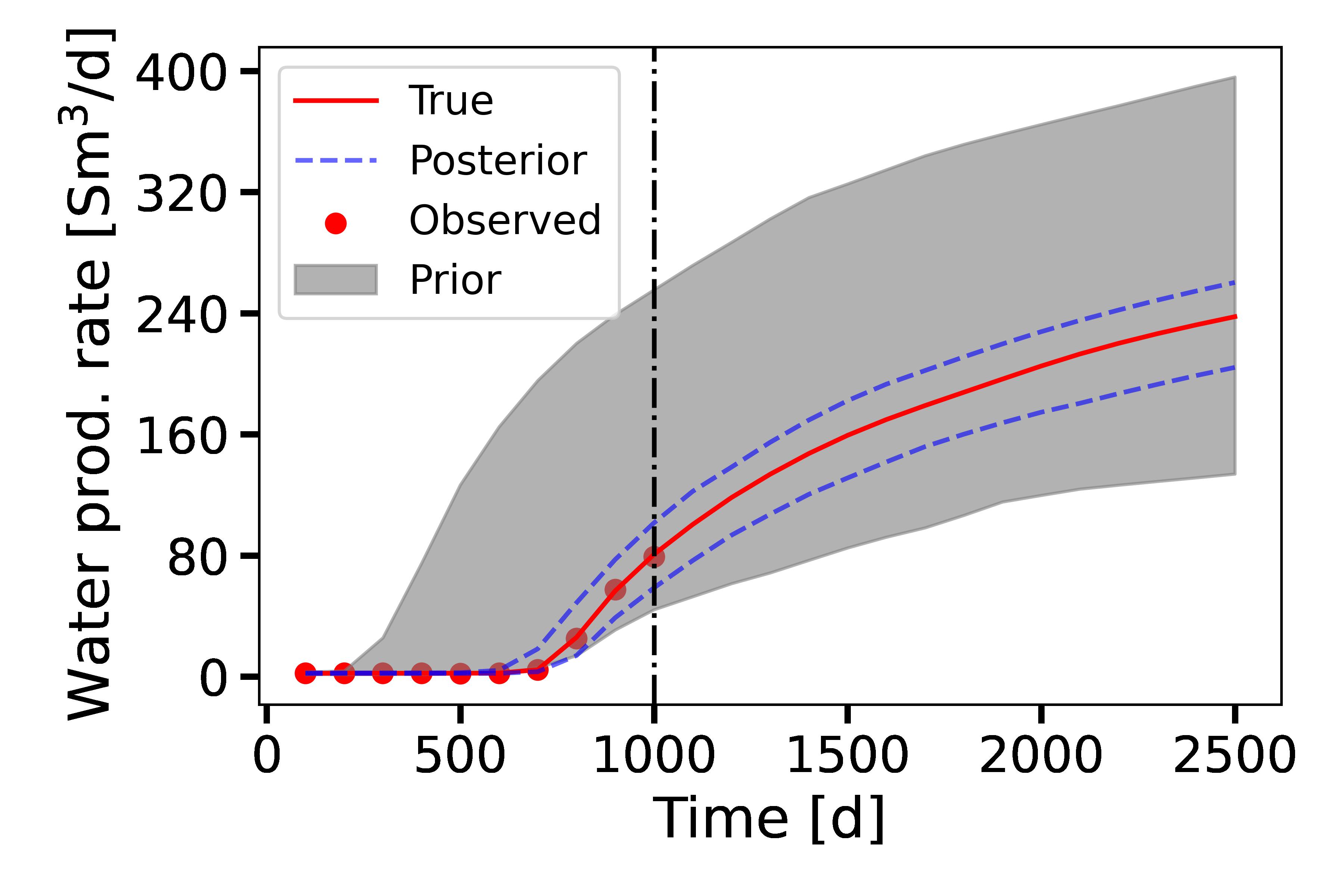}
        \caption{P2 water production rate}
        \label{fig:rates_P2_WAT_var}
    \end{subfigure}

    \caption{History matching results (Case~2) for key flow quantities. Gray regions show the prior P$_{10}$--P$_{90}$ range, blue dashed curves denote the posterior P$_{10}$--P$_{90}$ range, and red points and red curves represent observed and true data. The vertical dot-dash line indicates the end of the history matching period.
    \label{fig:rates_hm_var}
}
\end{figure}

Figure~\ref{fig:rates_hm_var} shows history matching results for oil and water rates. The curves, points and shaded regions have the same meaning as in Figure~\ref{fig:rates_hm}. The different prior ranges compared to Case~1 are due to the variation in the facies properties. Data assimilation leads to substantial uncertainty reduction in all flow quantities, and the true data consistently fall within the posterior P$_{10}$--P$_{90}$ ranges. This is also the case for quantities outside the P$_{10}$--P$_{90}$ prior ranges, such as the I1 water injection rate (Figure~\ref{fig:rates_hm_var}b) and the P1 water production rate (Figure~\ref{fig:rates_hm_var}e).

\begin{figure}
    \centering
    \begin{subfigure}[b]{0.18\textwidth}
        \centering
        \includegraphics[width=\textwidth, trim=2cm 0 2cm 0, clip]{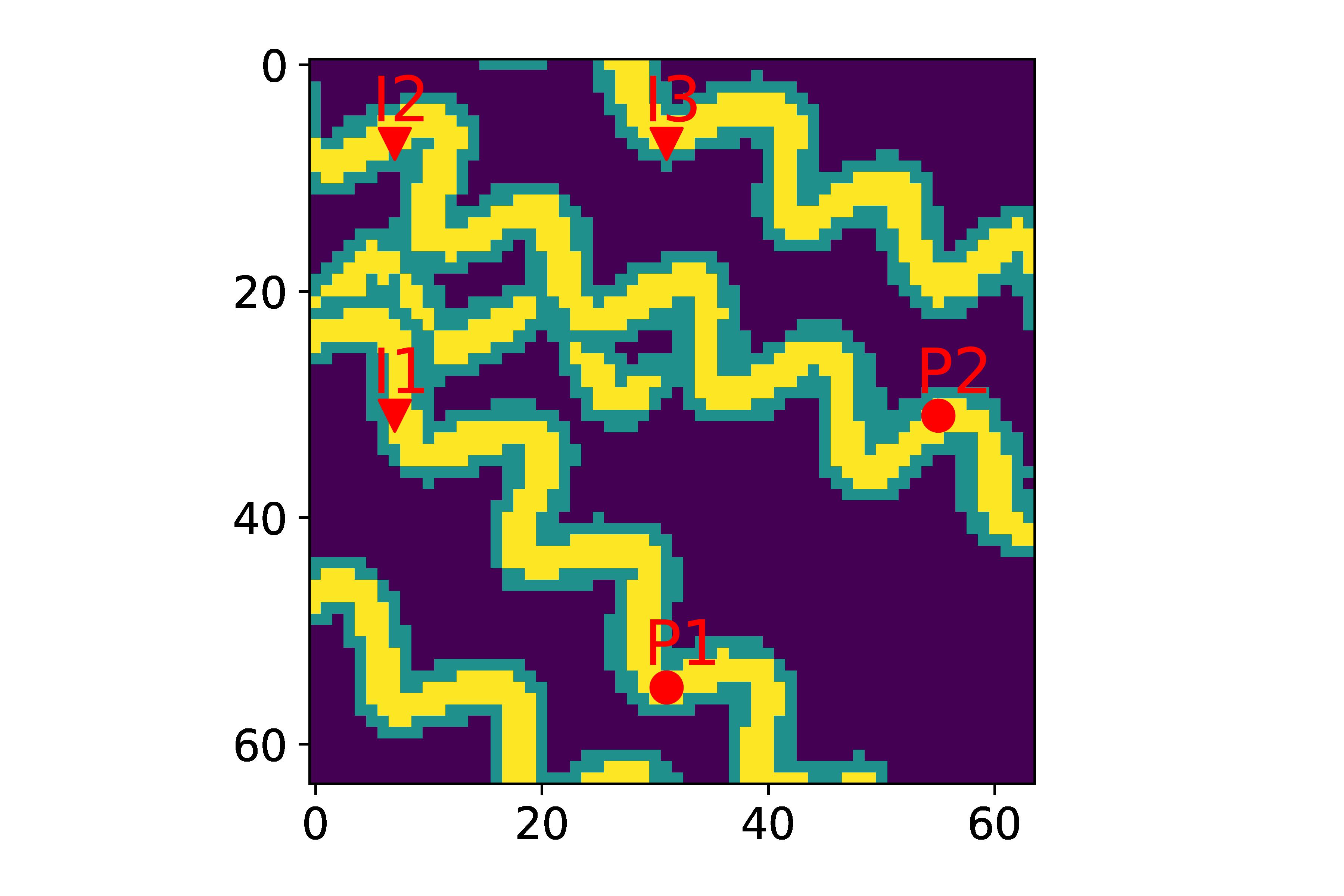}
        \caption{Post.~1}
        \label{fig:medoid_cluster_0_post_var}
    \end{subfigure}
    \begin{subfigure}[b]{0.18\textwidth}
        \centering
        \includegraphics[width=\textwidth, trim= 2cm 0 2cm 0, clip]{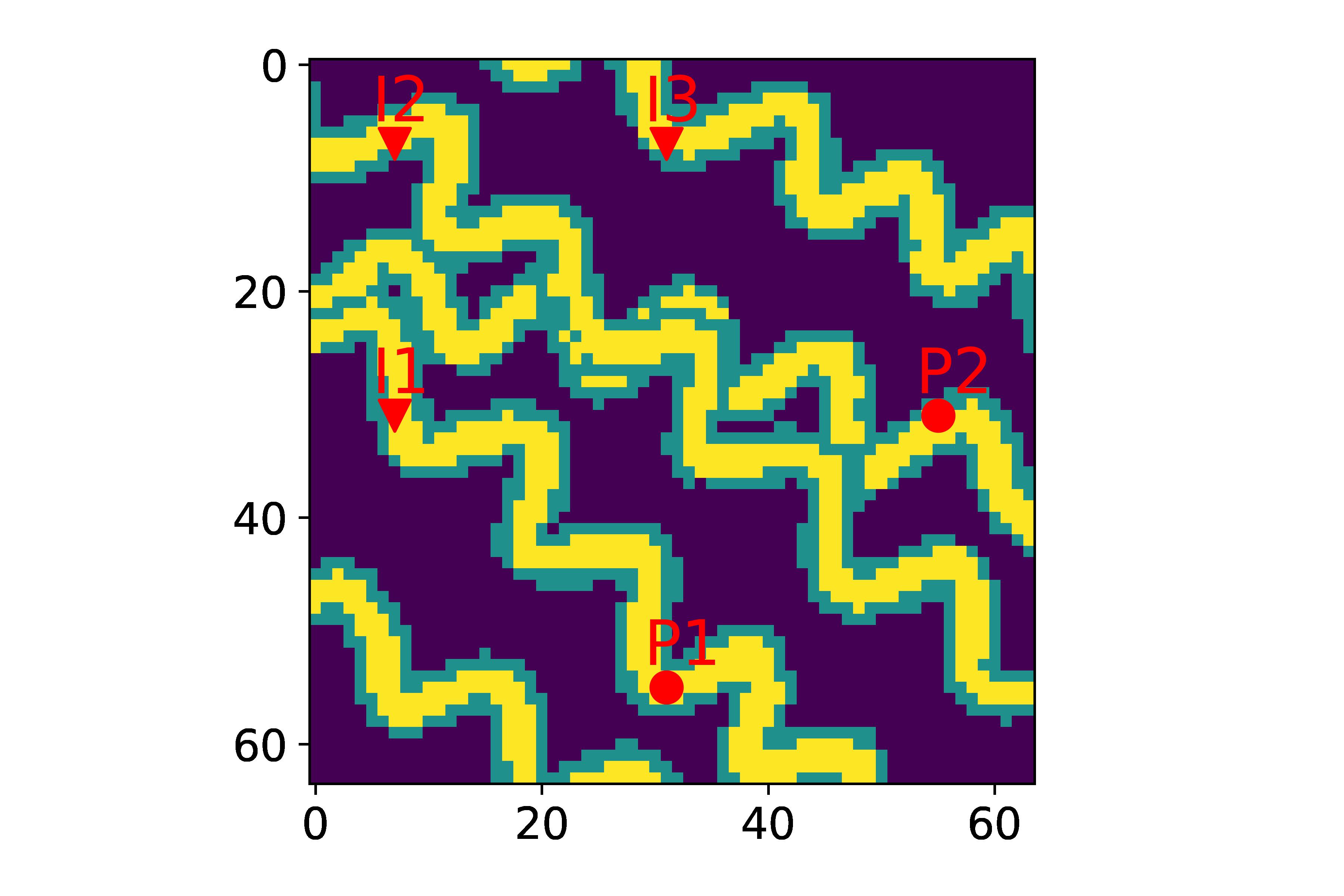}
        \caption{Post.~2}
        \label{fig:medoid_cluster_1_post_var}
    \end{subfigure}
    \begin{subfigure}[b]{0.18\textwidth}
        \centering
        \includegraphics[width=\textwidth, trim=2cm 0 2cm 0, clip]{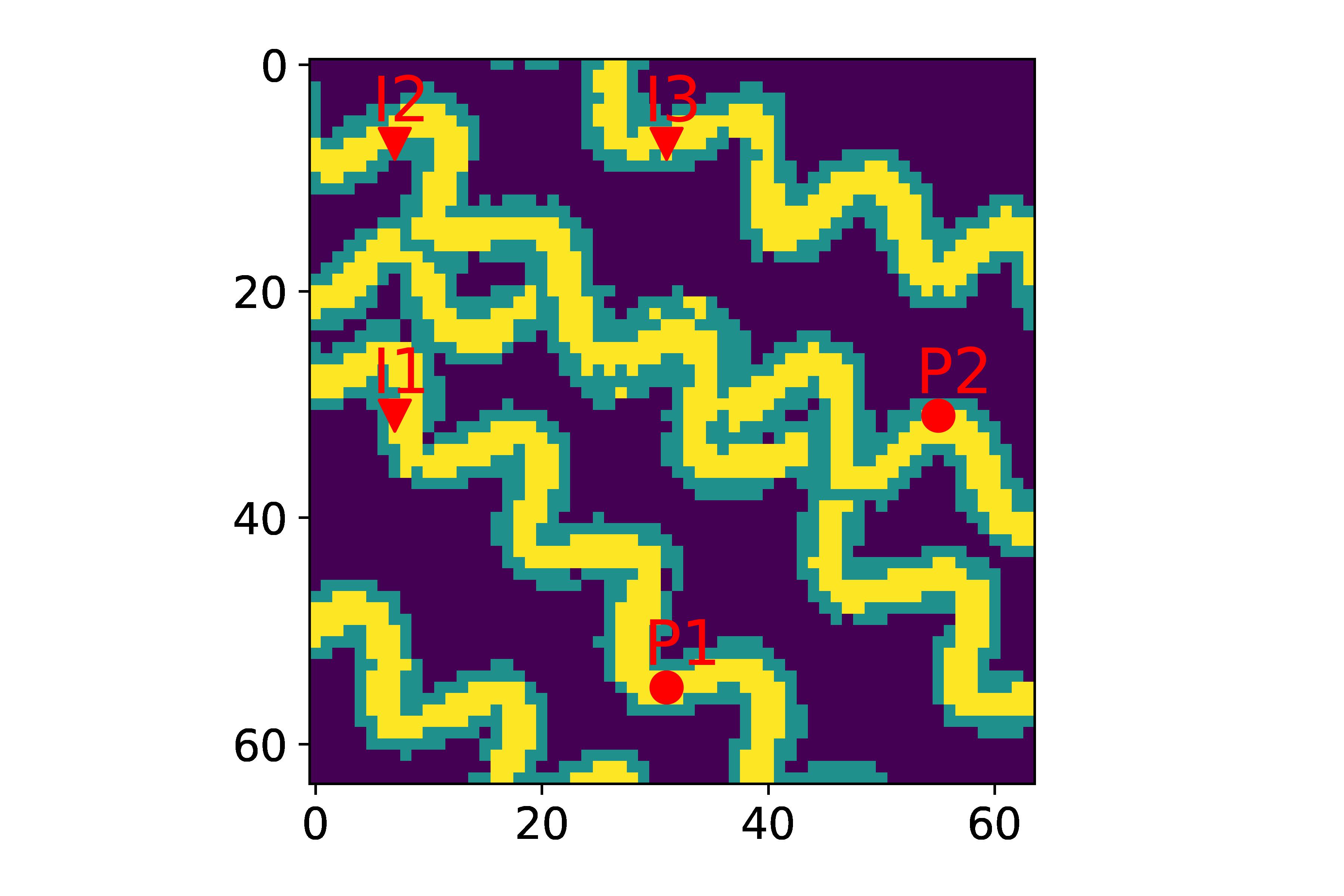}
        \caption{Post.~3}
        \label{fig:medoid_cluster_2_post_var}
    \end{subfigure}
    \begin{subfigure}[b]{0.18\textwidth}
        \centering
        \includegraphics[width=\textwidth, trim=2cm 0 2cm 0, clip]{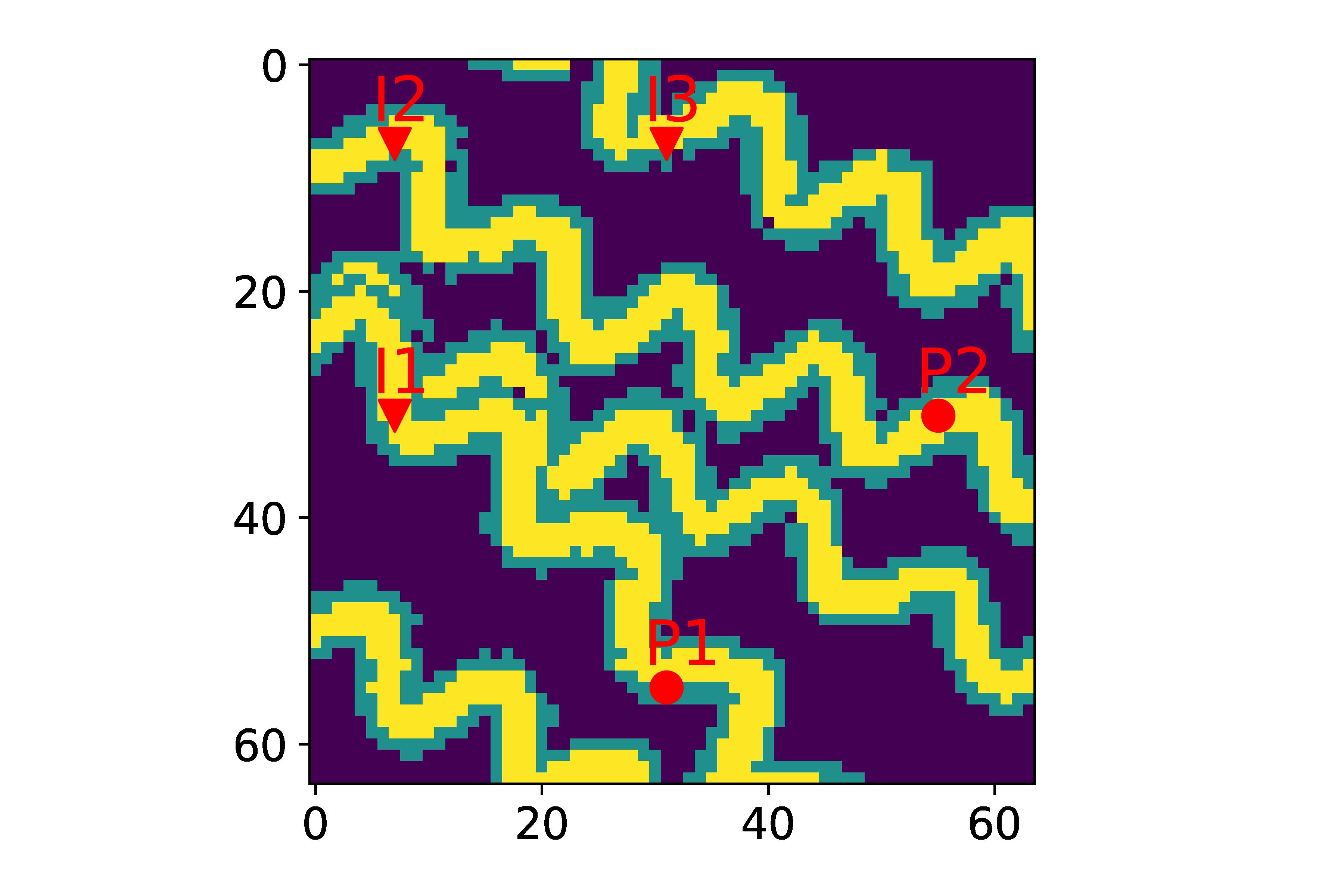}
        \caption{Post.~4}
        \label{fig:medoid_cluster_3_post_var}
    \end{subfigure}
    \begin{subfigure}[b]{0.18\textwidth}
        \centering
        \includegraphics[width=\textwidth, trim=2cm 0 2cm 0, clip]{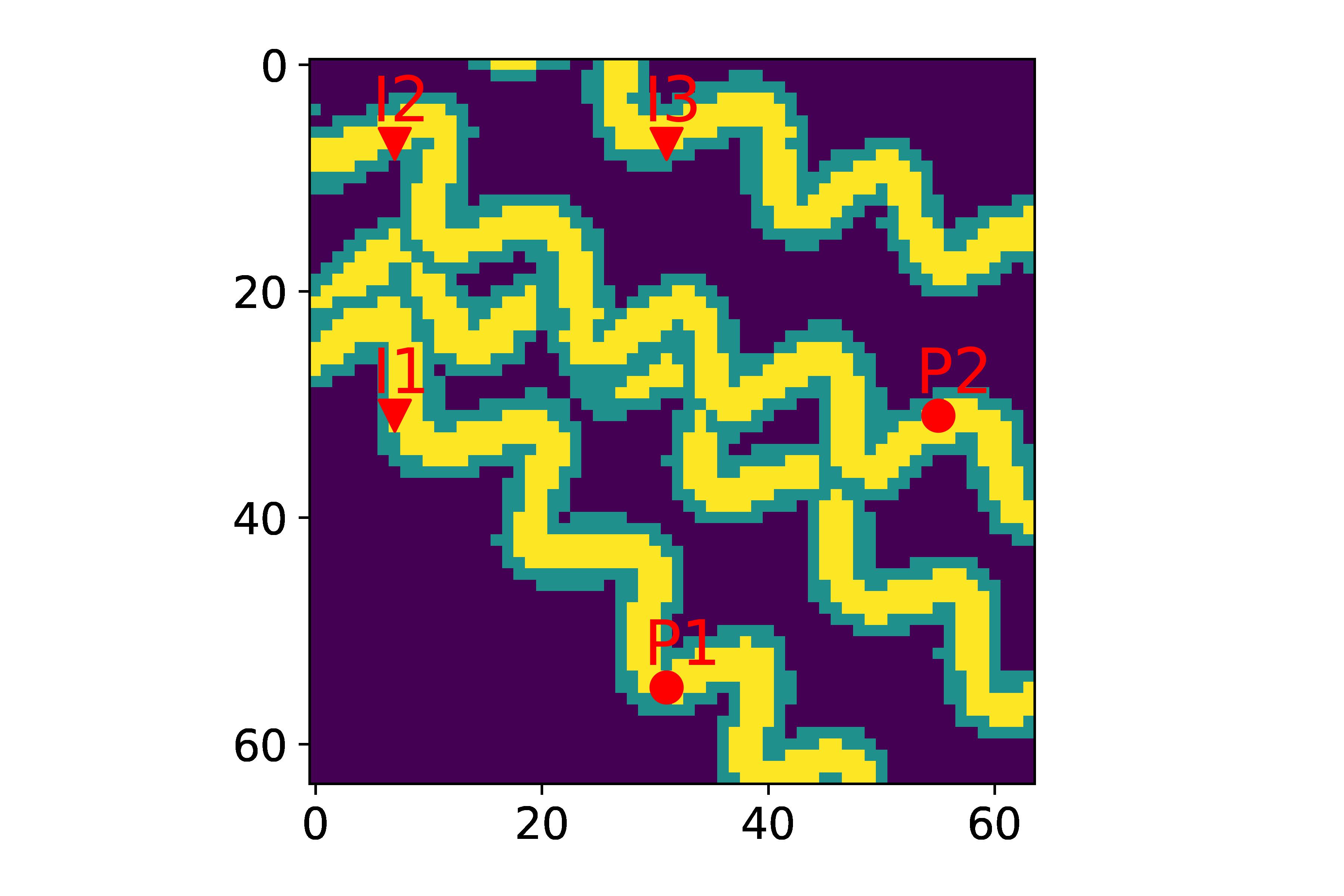}
        \caption{Post.~5}
        \label{fig:medoid_cluster_4_post_var}
    \end{subfigure}
    \caption{Posterior realizations (Case~2) determined by $k$-means clustering. Each model is the medoid of its cluster. True model shown in Figure~\ref{fig:trues}b.
    \label{fig:medoids_var}
}
\end{figure}

Representative posterior realizations, identified using the $k$-means clustering approach described previously, are shown in Figure~\ref{fig:medoids_var}. Note that the clustering is applied only to the facies models -- the porosity and permeability values do not enter this analysis. The representative prior realizations, which are the same as in Case~1, appear in Figure~\ref{fig:medoids}. All five posterior models exhibit three complete or partial channels connecting wells I1 and I2 to P1 and P2, along with an isolated channel where I3 is located. These features are consistent with the true model (Figure~\ref{fig:trues}b). There is still, however, a degree of variability in the posterior realizations. For example, the models in Figure~\ref{fig:medoids_var}a-d have an additional channel in the lower-left region, which is not present in the realization in Figure~\ref{fig:medoids_var}e. This channel does not exist in the true model, but it is not likely to have much effect on flow because no wells intersect it.

\begin{figure}
    \centering
    \begin{subfigure}[b]{0.45\textwidth}
        \centering
        \includegraphics[width=\textwidth]{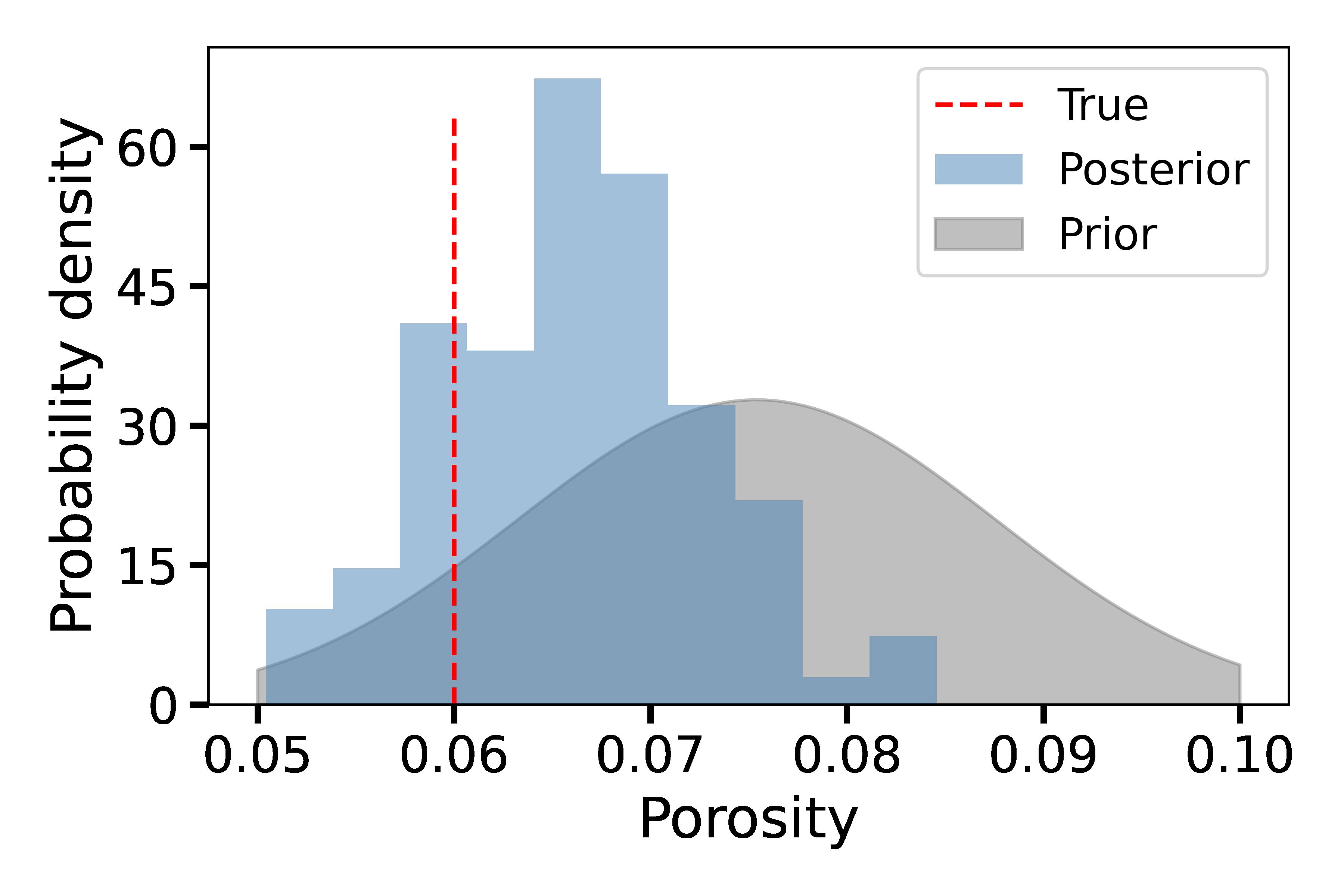}
        \caption{Mud porosity distribution}
        \label{fig:porosity_channel}
    \end{subfigure}
    \begin{subfigure}[b]{0.45\textwidth}
        \centering
        \includegraphics[width=\textwidth]{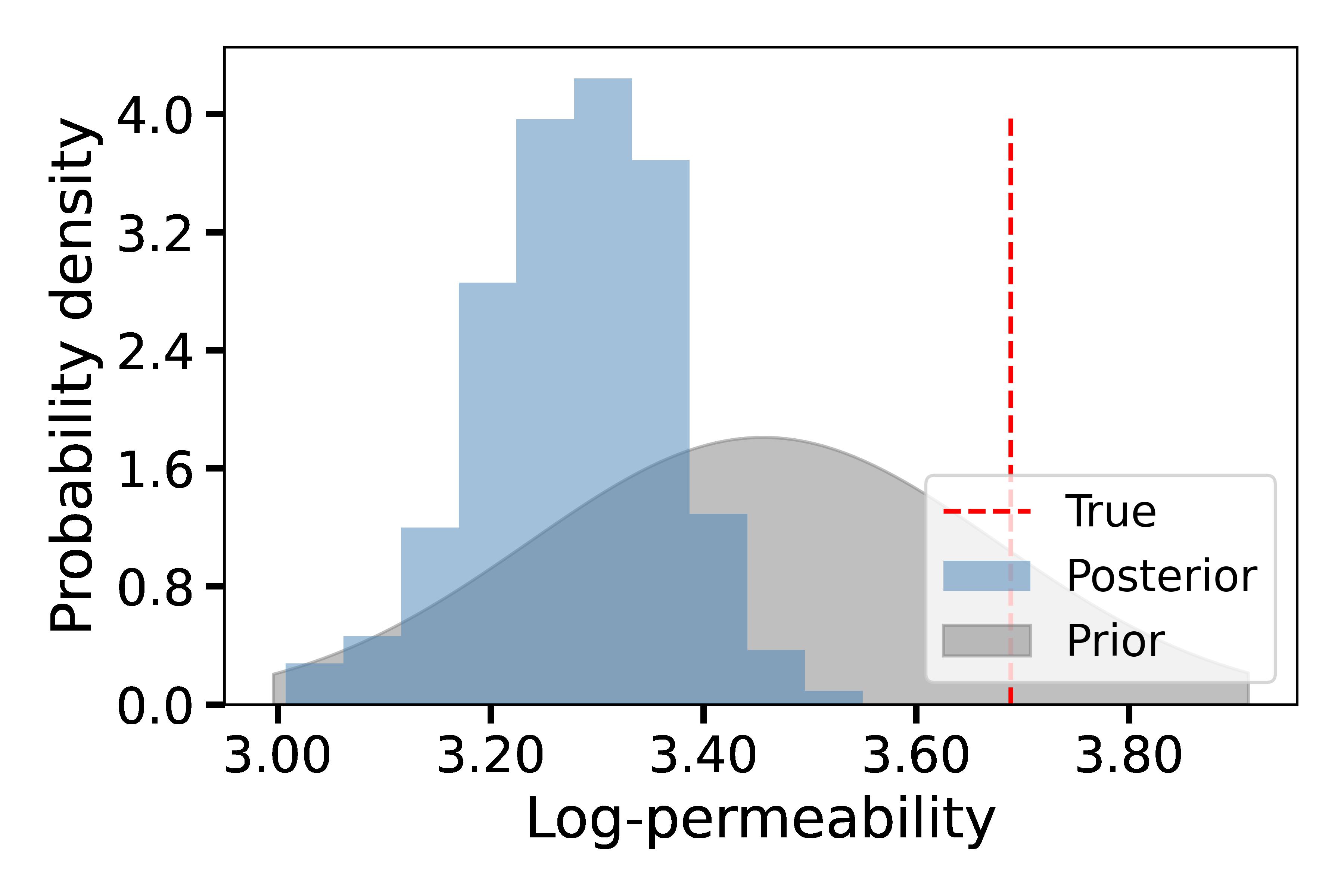}
        \caption{Mud permeability distribution}
        \label{fig:permeability_channel}
    \end{subfigure}
    \begin{subfigure}[b]{0.45\textwidth}
        \centering
        \includegraphics[width=\textwidth]{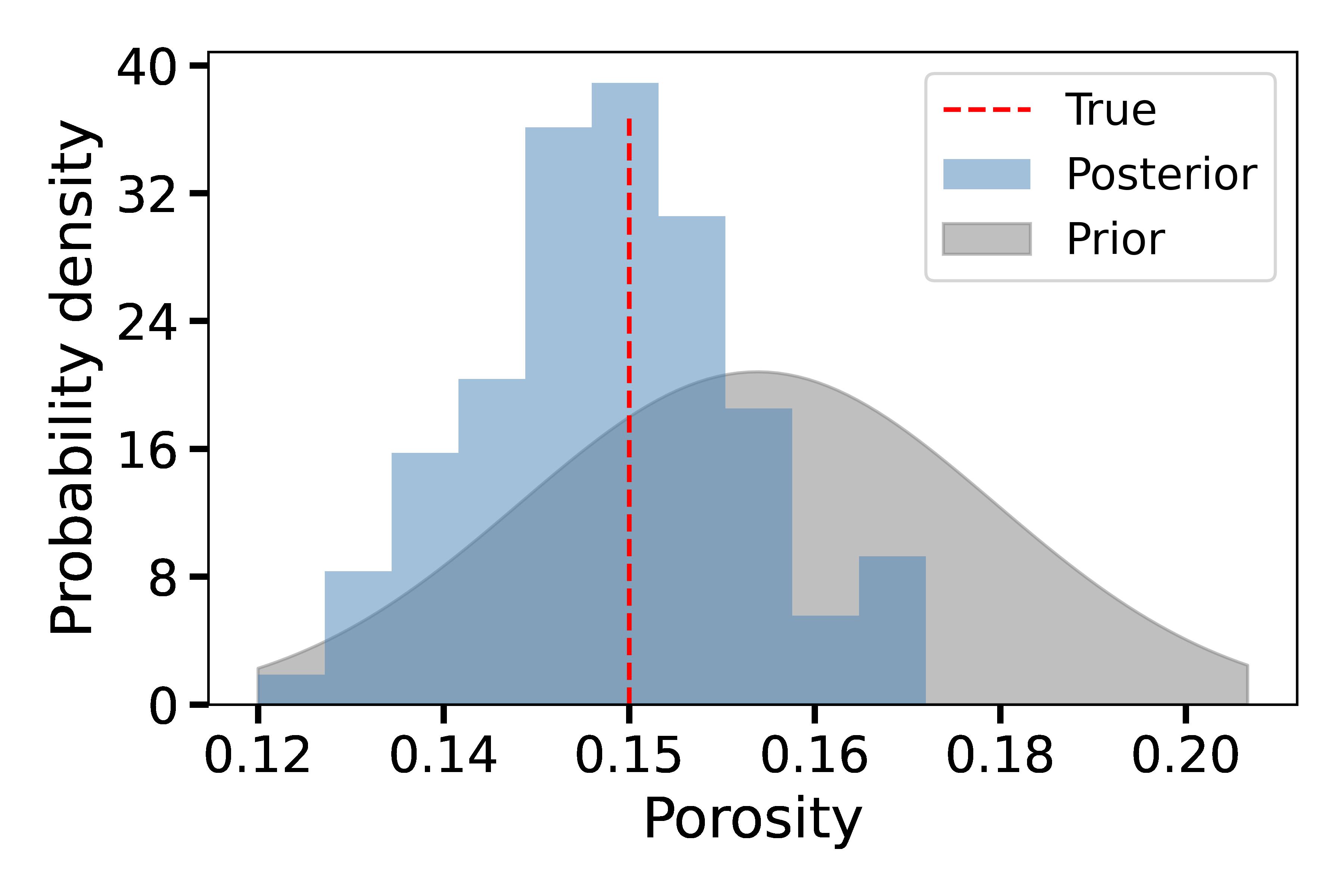}
        \caption{Levee porosity distribution}
        \label{fig:porosity_levee}
    \end{subfigure}
    \begin{subfigure}[b]{0.45\textwidth}
        \centering
        \includegraphics[width=\textwidth]{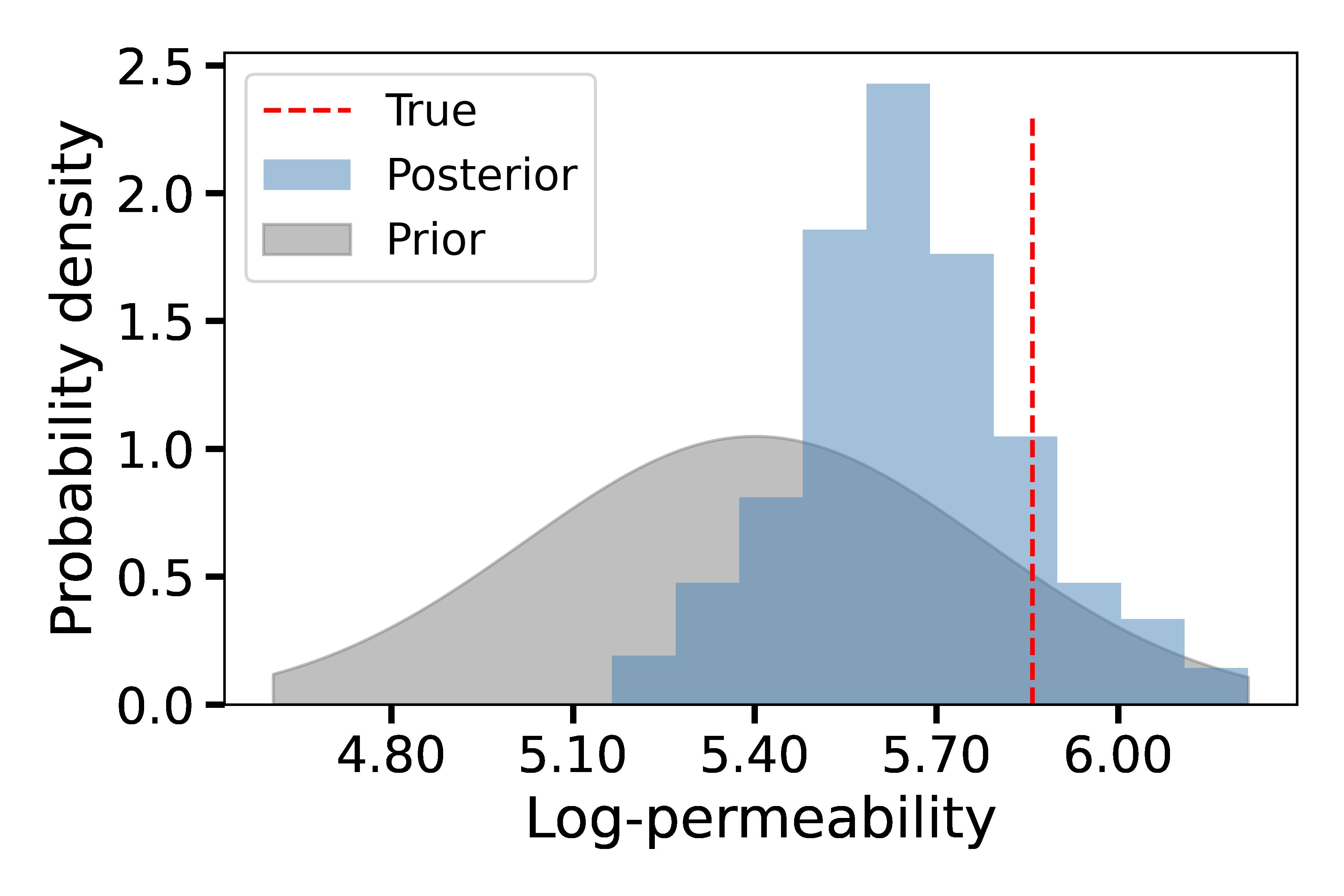}
        \caption{Levee permeability distribution}
        \label{fig:permeability_levee}
    \end{subfigure}
    \begin{subfigure}[b]{0.45\textwidth}
        \centering
        \includegraphics[width=\textwidth]{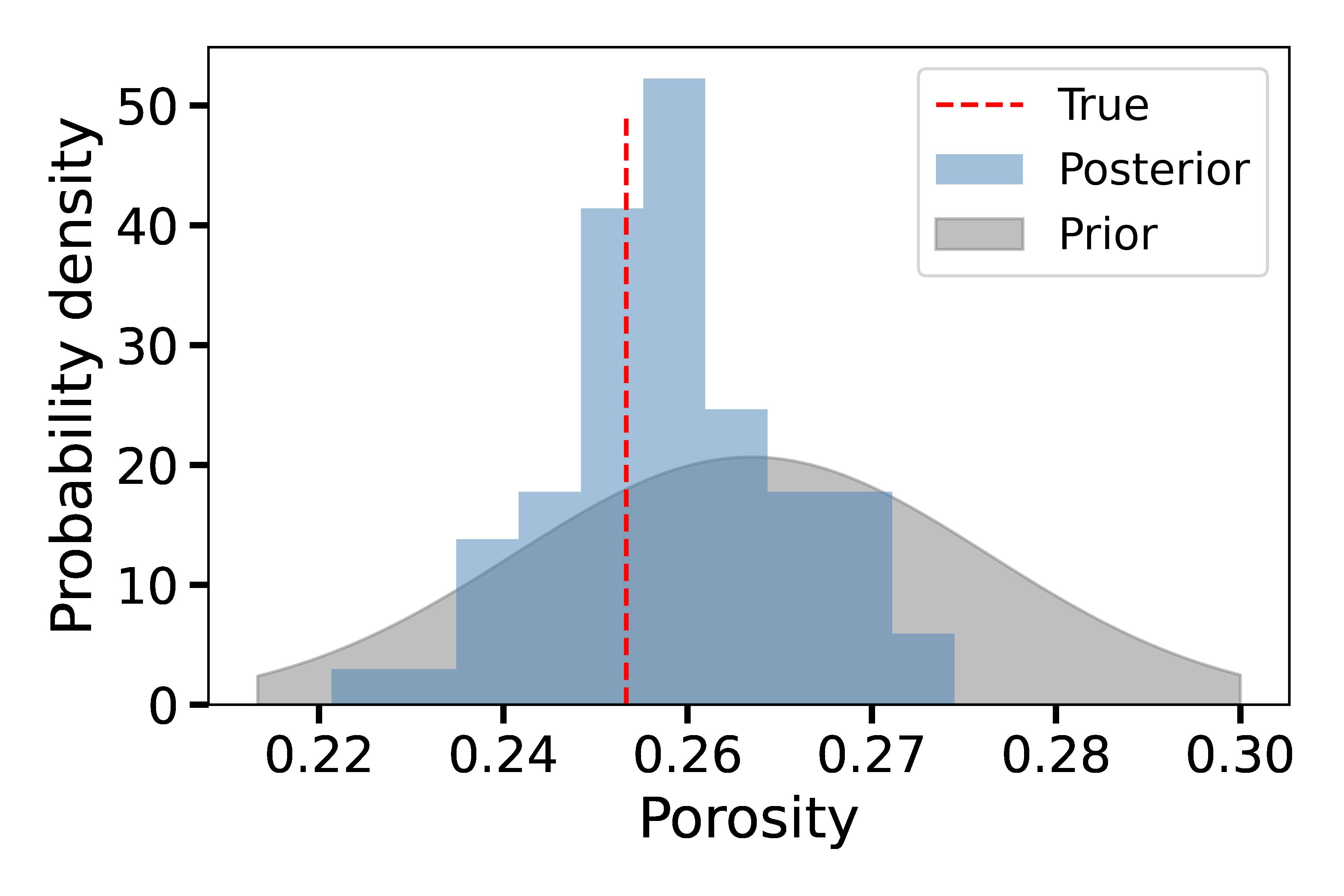}
        \caption{Channel porosity distribution}
        \label{fig:porosity_mud}
    \end{subfigure}
    \begin{subfigure}[b]{0.45\textwidth}
        \centering
        \includegraphics[width=\textwidth]{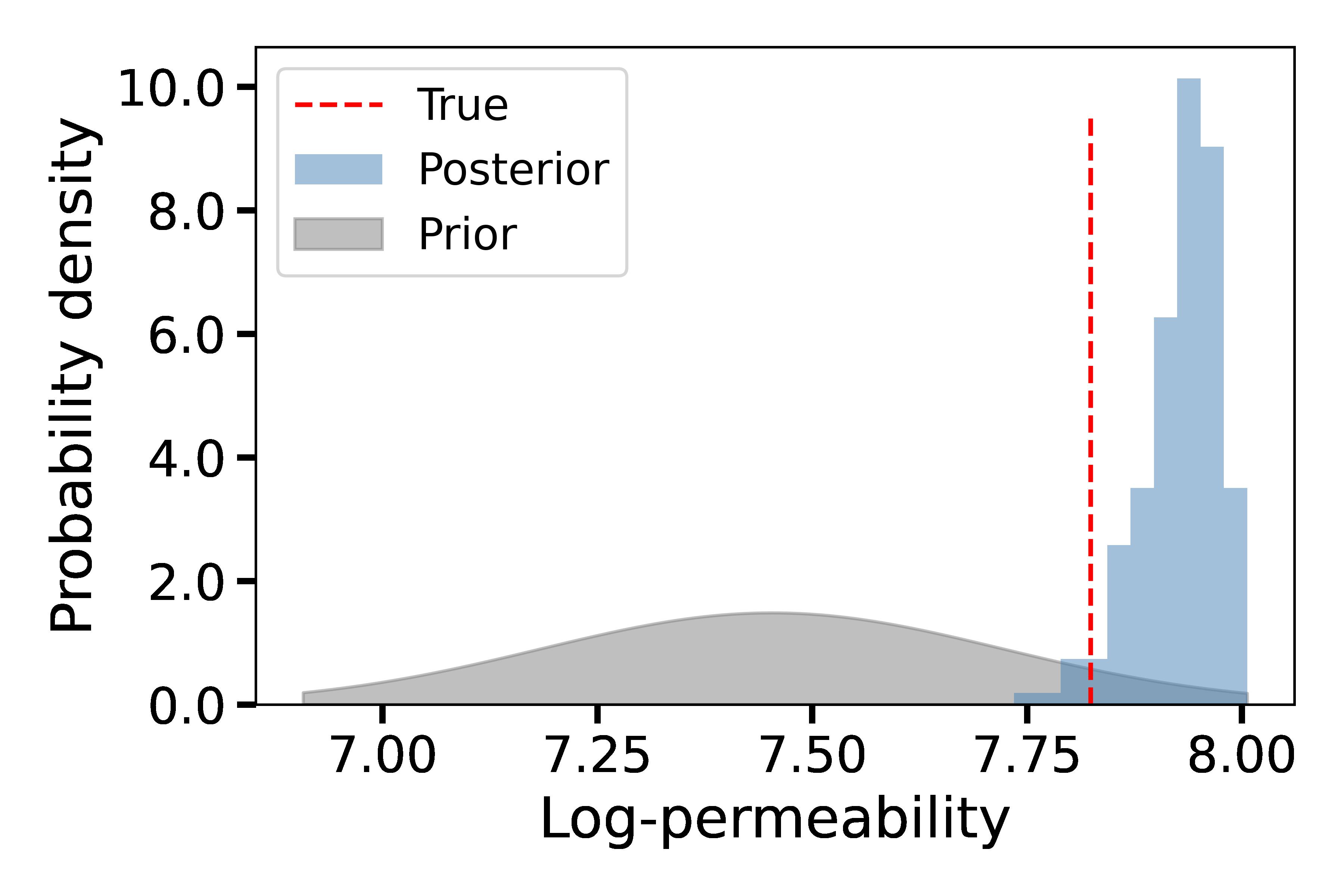}
        \caption{Channel permeability distribution}
        \label{fig:permeability_mud}
    \end{subfigure}
    \caption{History matching results (Case~2) for facies porosity and permeability. Gray regions show the prior distribution and blue histograms denote the posterior distribution. Vertical red dashed lines indicate true values.
    \label{fig:props_hm}
}
\end{figure}

Prior and posterior probability distributions for porosity and log-permeability of the three facies are presented in Figure~\ref{fig:props_hm}. Prior distributions are represented by the gray shaded regions, while posterior histograms are shown in blue. The vertical dashed lines indicate the ``true'' value for each property. In all but one case, the posterior distribution shifts towards (and includes) the true value. The one exception is for $\log k$ for mud (Figure~\ref{fig:props_hm}b), for which the true value lies outside the posterior histogram. This is presumably because the observed data (flow rates at wells) are not sensitive to this quantity. In any event, the mode of the posterior corresponds to $k_{mud}\approx 27$~md, while the true value is about 40~md.

\vfill\eject 
\section{Concluding remarks}
\label{sec:conclusion}

In this work, a geological parameterization based on generative latent diffusion models (LDMs) was introduced and applied for the representation of three-facies channelized systems in 2D. The generative model consists of a variational autoencoder to enable dimension reduction and a U-net to perform the reverse diffusion process. Our LDM implementation may have some advantages over previous methods for geomodel parameterization as it involves fast, deterministic DDIM sampling combined with reduced-dimensionality latent-variable representations. These features enable the use of the LDM for challenging applications such as data assimilation.

LDM training is accomplished in two steps. First, the VAE is trained to learn the latent space mapping, and then the U-net is trained to learn the denoising process. The training set in this work comprised 4000 conditional realizations generated using Petrel. A hard data loss term was included during VAE training to ensure the generated models honored the conditioning in the training set. The LDM-generated models were shown to accurately reproduce the geological and flow-based features of the training set. This was demonstrated through several comparisons, including spatial property distributions and two-phase flow simulation statistics. Furthermore, the LDM parameterization was shown to behave smoothly when the latent input variable is perturbed. Finally, history matching was applied using ESMDA in the LDM latent space. Both fixed and uncertain facies porosities and permeabilities were considered. Reduction in uncertainty for injection and production rates was observed, and posterior geomodels were shown to capture the key geological features appearing in the synthetic true model. 

In future work, it will be of interest to evaluate the performance of the LDM parameterization with other data assimilation methods, such as Markov chain Monte Carlo or randomized maximum likelihood methods. It is also of interest to enhance the parameterization to treat multiple geological scenarios involving, for example, different channel orientations, sinuosity, thickness, etc. This could enable history matching to be performed, in a unified manner, in situations involving scenario uncertainty. Finally, the LDM should be extended to handle large, realistic 3D models.

\section*{Declaration of competing interest}
The authors declare that they have no known competing financial interests or personal relationships that could have appeared to influence 
the work reported in this paper. 

\section*{Acknowledgements}
\label{sec:acknowledge}
We are grateful to the industrial affiliates of the Stanford Smart Fields Consortium for financial support. We also thank Tapan Mukerji, Suihong Song, and Su Jiang for useful discussions, Oleg Volkov for ADGPRS software support, and Dylan Crain and Wenchao Teng for assistance with the ESMDA code. We acknowledge the Stanford Doerr School of Sustainability (SDSS) Center for Computation for providing the computational resources used in this work.\\

\
\section*{Code availability}

The source codes and dataset will available for downloading upon publication. The codes are based on the implementations provided in the following repositories \url{https://github.com/huggingface/diffusers} and  \url{https://github.com/Project-abs/tutorials/tree/main/generative}, using the Python libraries \texttt{diffusers} and \texttt{monai}.

\newpage
\bibliographystyle{cas-model2-names.bst}
\bibliography{bibliography.bib}

\end{document}